%% file: arxiv.tex
\documentclass[letterpaper, 12pt]{article} 
\usepackage[preprint]{aaai2027}
\renewcommand{\twocolumn}[1][]{\onecolumn#1}
\usepackage[hyphens]{url}  
\usepackage{graphicx} 
\usepackage{natbib}  
\usepackage{caption} 
\usepackage{algorithm}
\usepackage{algpseudocode}

\usepackage{newfloat}
\usepackage{listings}
\DeclareCaptionStyle{ruled}{labelfont=normalfont,labelsep=colon,strut=off} 
\floatstyle{ruled}
\newfloat{listing}{tb}{lst}{}
\floatname{listing}{Listing}

\usepackage{booktabs}

\newcommand{\fitcol}{\ifdim\width>\columnwidth\columnwidth\else\width\fi}

\usepackage{amsmath,amssymb,amsthm}
\usepackage{enumitem}
\usepackage{xcolor}
\usepackage{tikz}
\usetikzlibrary{arrows.meta, positioning, backgrounds, fit}

\IfFileExists{microtype.sty}{\usepackage{microtype}}{} 
\newif\ifdraft
\draftfalse 

\newif\ifshowedits
\showeditstrue 
\ifshowedits\usepackage{soul}\soulregister\cite7\soulregister\ref7\fi 
\newcommand{\edittag}[1]{\if\relax\detokenize{#1}\relax\else\textsuperscript{\scriptsize\sffamily\bfseries\color{blue}[#1]}\fi}

\newcommand{\sys}{EG}                                
\newcommand{\syss}{EGs}                              
\newcommand{\sysname}{Entailment Graph}              
\newcommand{\sysnames}{Entailment Graphs}            
\newcommand{\sysfull}{Logical Entailment Graph}      
\newcommand{\sysfulls}{Logical Entailment Graphs}    
\newcommand{\sysnoun}{entailment graph}              
\newcommand{\sysnouns}{entailment graphs}            
\ifshowedits\soulregister\sys0\soulregister\syss0\soulregister\sysname0\soulregister\sysnames0\soulregister\sysfull0\soulregister\sysfulls0\soulregister\sysnoun0\soulregister\sysnouns0\fi

\newif\ifarxiv
\arxivtrue 

\newcommand{\cand}[1]{\mathbf{\mathsf{#1}}}
\newcommand{\candTJ}{\cand{TLS\_JWT}}
\newcommand{\candTI}{\cand{TLS\_IP}}
\newcommand{\candT}{\cand{TLS}}
\newcommand{\candCr}{\cand{CRED}}
\newcommand{\candCo}{\cand{COND}}

\lstdefinestyle{prompt}{basicstyle=\ttfamily\scriptsize,breaklines=true,
  columns=fullflexible,keepspaces=true,frame=single,extendedchars=true,
  literate={¬}{{$\neg$}}1 {∧}{{$\wedge$}}1 {∨}{{$\vee$}}1 {→}{{$\rightarrow$}}1
    {↔}{{$\leftrightarrow$}}1 {↛}{{$\not\rightarrow$}}1 {⊕}{{$\oplus$}}1 {—}{{---}}1}
\newtheorem{theorem}{Theorem}
\newtheorem{lemma}[theorem]{Lemma}
\newtheorem{definition}[theorem]{Definition}
\newtheorem{corollary}[theorem]{Corollary}
\newtheorem{proposition}[theorem]{Proposition}
\newtheorem{remark}[theorem]{Remark}

\title{Safe Autoformalization: Sufficiency and Necessity with \sysnames{}}
\author{Umut A. Acar\thanks{Umut A. Acar holds concurrent appointments as an
Amazon Scholar and as a Professor of Computer Science at Carnegie Mellon
University. This paper describes work performed at Amazon Web Services.}, Eli Goldweber}
\affiliations{Amazon Web Services}

\begin{document}

\maketitle

\begin{abstract}
Autoformalization translates a natural-language statement into a formula that
captures its \emph{intent}, which depends on the statement's \emph{context}: the
background that disambiguates it. LLMs make autoformalization practical, but their
probabilistic output carries no formal guarantees. We propose techniques for
\emph{safe} autoformalization that certify one-sided logical adequacy: a formula
\emph{sufficient} for the intent is safe to enforce, and a formula
\emph{necessary} for it is safe to rely on.

We use LLMs to generate candidate
formulas and order them by logical strength in an \emph{\sysname{}} (\sys{}), an
ordering a solver decides. Each edge carries a \emph{residual} and a
\emph{co-residual}, the two adjoints of the entailment order, all fixed by the
candidate pool alone. In a context, two searches return the
\emph{sufficient frontier}, the weakest candidates that entail the statement, and
the \emph{necessary frontier}, the strongest candidates the statement entails;
each step is one local check against an edge's residual or co-residual. The
frontiers \emph{bracket} the statement and yield the \emph{ambiguity residue},
which holds exactly of the situations the candidates certify neither as
satisfying nor as violating it; the statement is \emph{ambiguous} precisely when
the residue is satisfiable. For natural-language statements and contexts, an LLM
\emph{entailment oracle} decides the checks; it is \emph{sound} if it accepts
only checks that hold, and \emph{complete} if it accepts every check that holds.
We prove that a sound oracle makes the search \emph{safe} regardless of pool
quality, and that a complete one makes the frontiers \emph{tight}.

On security-critical AWS properties, \syss{} are safe in all $20$ cases of a
expert-adjudicated study, where majority vote over equivalence clusters
and Universal Self-Consistency are unsafe in $25\%$ and $10\%$. An ablation with
the same oracle but no graph is unsafe in $25\%$ and rejects every candidate in
another $30\%$. We machine-checked our proofs in Lean.
\end{abstract}

\section{Introduction}\label{sec:intro}

Autoformalization translates a natural-language (NL) statement into a formal representation; doing so correctly requires that the representation capture the \emph{intent} of the statement.
Over the past several years, autoformalization with Large Language Models (LLMs)
has drawn growing attention for a variety of domains including for formalizing mathematics~\cite{autoformalization-llm,
  autoformalization-subtasks, symbolic-consistency, autoformalization-eval}, for deriving temporal-logic specifications~\cite{nl2tl, nl2spec}, program contracts and
postconditions~\cite{nl2postcond, specgen, autospec}, and for security
policies~\cite{cedar-policy}. 
Across these domains, autoformalization allows requirements to be written in
natural language and enables a formal system such as a solver, a model checker, or a proof assistant to carry out the formally grounded work of verifying the desired task.
For example, autoformalization constitutes the first step in
end-to-end verification pipelines~\cite{loopy, clover, baldur, alphaverus,
  verifythisbench} that were previously infeasible.

The end-to-end promise in verification pipelines rests on the quality of the formalization. 
A specification that is too weak lets unsafe implementations pass; one
that is too strong rejects correct ones. 
Yet LLM-based autoformalization comes with no guarantees, especially given that 
natural language is ambiguous in ways LLMs resolve poorly~\cite{ambient}.
LLMs can formalize the same statement into multiple, inequivalent formulas, without any guarantee on logical adequacy.
Which of them is intended depends on the \emph{context}---the background that
disambiguates the statement.
There has been research on using statistical signals for autoformalization.
For example, semantic
entropy~\cite{semantic-entropy, semantic-entropy-nature} clusters a set of
candidates based on their equivalence and can quantify how much they disagree. The statistical selection rules it suggests, e.g., taking the largest
cluster, carry no guarantee on the logical adequacy of the selection with respect to the intent.  

Verification pipelines use formulas to \emph{verify a condition} by deciding whether an artifact such as an implementation or a configuration actually conforms to it.
The role played by the formula determines the meaning of logical adequacy.
A formula that specifies a
\emph{guarantee}---one to be enforced, rejecting artifacts that fail it---must be
\emph{sufficient} for the intent: every artifact it admits must be one the intent
admits, or enforcement lets a violation through.
A formula that specifies an \emph{assumption}---one relied on, rather than
checked, of artifacts someone else produces---must instead be \emph{necessary}:
it must claim no more than the intent guarantees.
Neither role
requires the formula to be equivalent to the intent; each requires one direction
of entailment.
The two roles are the two sides of a contract~\cite{contracts}, with the same
monotonicity: the role that enforces supplies a guarantee, which refinement may
strengthen, and the role that relies supplies an assumption, which refinement may
weaken. 
These one-sided guarantees align with the weakest preconditions and
strongest postconditions central to program
verification~\cite{dijkstra-wp,lin-snc-wsc}.

We therefore call an autoformalization \emph{safe} when it carries a certificate
in the direction its role requires. 
In this paper we propose techniques for safe autoformalization.
The proposed techniques revolve around a graph representation of candidate formulas and the logical strength relationships between them.
We present the techniques in three stages.
First, from a pool of candidates alone, we build the graph and define the two
formulas that separate adjacent candidates: an edge's \emph{co-residual} captures the
weakening along it and its \emph{residual} the strengthening. No context appears in
this stage---the graph and both formulas are fixed by the pool, and they are the two
adjoints of the entailment order (Section~\ref{sec:ig}).
Second, we fix a context and the intent, both as formulas, and define the search
algorithms that traverse the already-built graph, the frontiers they return, and the
ambiguity residue between them
(Sections~\ref{sec:search} and~\ref{sec:ambiguity}).
Third, we drop the assumption that the context and the intent are formulas, and
realize the searches by combining the symbolic machinery with LLMs
(Sections~\ref{sec:realizability} and~\ref{sec:impl}).

Given a pool of candidate formulas in first-order logic, we
``complete'' the candidates with universal ``True'' and ``False'' and
build a directed acyclic graph---the \emph{(Logical) \sysname{}} (\sys{})---whose
edges are strict entailments, an ordering a solver decides.
Each edge carries two formulas that separate its endpoints. The
\emph{co-residual}---the strongest formula whose disjunction with the stronger
candidate recovers the weaker---captures the weakening along the edge. The
\emph{residual}---the weakest formula whose conjunction with the weaker candidate
recovers the stronger---captures the strengthening.
Both are sharp for an algebraic reason
(Section~\ref{sec:ig}): they are the two \emph{adjoints} of the entailment order,
the residual right-adjoint to conjunction and the co-residual left-adjoint to
disjunction, in the Heyting algebra of formulas modulo equivalence.

Given a statement and a context, two searches traverse this graph, one in each
direction. The downward search returns the \emph{sufficient frontier}, the weakest
candidates that entail the statement; the upward search returns the \emph{necessary
frontier}, the strongest candidates the statement entails.
A single local check decides each edge traversal: weakening across an edge is decided
by its co-residual and strengthening by its residual, each checked against the
statement in the context. Because the two formulas are adjoints, each check is
exact: it holds precisely when the traversal is licensed. Relativizing the adjoints to
the context does not change either check (Lemma~\ref{lem:absolute}), so one graph
serves every context.
We prove both searches safe: every candidate in the sufficient frontier entails the
statement in context, and the statement entails every candidate in the necessary
frontier (Theorems~\ref{thm:weakest-correct} and~\ref{thm:strongest-correct}).

\input{fig-ig-bracket}

The two frontiers returned by the searches \emph{bracket} the statement, the
sufficient frontier from below and the necessary frontier from above. Combining them
yields a single formula---the \emph{ambiguity residue}---that holds exactly of the
situations the candidates certify neither as satisfying nor as
violating the statement (Figure~\ref{fig:ig-bracket}). 
The NL statement is \emph{ambiguous} precisely when the residue is
satisfiable.
A solver can decide this whenever the context is also a formula.
Unlike a scalar uncertainty score, the ambiguity residue is a formula
that identifies exactly what remains ambiguous, and so defines exactly what a clarifying
question would need to settle to resolve ambiguity.
These guarantees hold under ambiguity and for any pool of candidates; they assume
a formal statement and context whose entailments a solver decides.

The frontier searches take the intent and the context as formulas, and
autoformalization supplies neither: both arrive as natural language.
In Section~\ref{sec:realizability}, we eliminate these assumptions and
realize \syss{} and the frontier searches by
combining neural and symbolic reasoning, with an LLM as the entailment oracle.
The two roles stay separate: a solver decides the entailments \emph{between
candidates} that build the graph, and the oracle decides only the edge checks,
each an entailment against the natural-language statement and context.
We denote $(\mathit{ctx}, s)$ for the faithful formalization of the
natural-language context and statement. We define the oracle to be \emph{sound}
if every check it accepts holds of $(\mathit{ctx}, s)$---accepting the
co-residual check of an edge $(u,v)$ implies
$(v \land \lnot u) \land \mathit{ctx} \models s$, and accepting its residual check
implies $s \land \mathit{ctx} \models (v \Rightarrow u)$---and \emph{complete} if
it accepts every such check that holds. 
As usual, soundness permits no false positives, completeness no false negatives.
We prove that a sound oracle makes the selection \emph{safe}.
The sufficient frontier entails the intent in context and hence is safe to enforce;
the necessary frontier is entailed by the intent in context and hence safe to rely on
(Theorem~\ref{thm:realizability}).
We further prove that a complete oracle makes the frontiers \emph{tight}: the
sufficient frontier contains the weakest candidates sufficient for the intent,
and the necessary frontier the strongest candidates the intent entails
(Theorem~\ref{cor:accuracy}). 
Our safety guarantee thus admits a one-sided error:
the oracle may be incomplete without violating safety, because an incomplete
oracle only costs tightness.
These results rely crucially on \syss{}, which keep the search algorithms on the safe
side by ordering the candidates and localizing the oracle's decisions.
Neither result assumes anything about the
candidate pool: a poor pool makes the selection slack, but never unsafe.
We show that under a further \emph{coverage} premise
(Definition~\ref{def:coverage})---that the pool does contain the
intent up to algebra---it can be recovered exactly
(Theorem~\ref{thm:disjunctive-recovery}).

We implement \syss{} by combining LLMs with
\emph{satisfiability modulo theories} (SMT) solvers~\cite{smt-handbook}, reifying the
neurosymbolic split (Section~\ref{sec:impl}).
At a high level, the implementation grounds autoformalization in logic by
requiring an LLM to make local, inspectable decisions within a precise logical
structure.
Any autoformalization method must trust some judgment that a formula captures a natural-language
specification.
\syss{} split such a global judgment into multiple judgments, each tied
to a single edge between two candidates.
A consequence follows that does not depend on the oracle
answering well: the two error directions are not symmetric. A wrongly
rejected edge halts the traversal and returns a candidate that is slack but is still
certified to be stronger than the weakest sufficient candidate in the pool going down,
weaker than the strongest necessary one going up---so only a wrongly
\emph{accepted} edge can admit a violation, and that is a judgment about the gap
between two candidates rather than about a formula and a sentence. Our safety guarantee therefore does not require the oracle to be
\emph{complete}; it requires only that each local, per-edge judgment be
\emph{sound}. The structure turns a single global judgment into many local ones,
of which only one direction need be right.
%

We conduct an empirical evaluation by considering AWS cloud safety properties.
Experiments show that \sys{} selection is safe in all $20$ cases of an
expert-adjudicated study, where majority vote over equivalence clusters (MV) and
USC~\cite{usc} are unsafe in $25\%$ and $10\%$ of cases, respectively.
An \emph{ablation} tests the locality claim above directly, removing
the graph and keeping everything else: asked one global entailment question per
candidate instead of one local question per edge, the same oracle and model is
unsafe in $25\%$ of the cases and rejects every candidate in another $30\%$. The
entailment structure, not access to an entailment oracle, is what delivers the
guarantee.
To scale beyond expert adjudication, we judge a further $152$
statements two ways. We run a solver check against the oracle's sufficient
frontier, which \sys{} passes by construction and on which MV and USC fall short in
$22$--$24\%$ of statements. We then have an LLM judge flag suspect \sys{} selections
and inspect every flagged case by hand, which confirms $3$ \emph{unsafe}
selections, each too weak to entail the intent. Since we inspect only flagged
cases, $3$ is a lower bound (Section~\ref{sec:eval}).
We defer some technical detail,
including proofs, to the appendix, which also documents
machine-checked proofs of all our results in the Lean~4 proof
assistant~\cite{lean4}.

The contributions of the paper include the following.
\begin{itemize}[leftmargin=1.2em, itemsep=1pt, topsep=2pt]
  \item \textbf{Safe autoformalization} offers a
  definition of adequacy for verification-oriented applications, where a
  formalization decides whether an artifact conforms.

  \item \textbf{\sysnames{} and frontier searches} (Sections~\ref{sec:ig}
  and~\ref{sec:search}). An \sys{} orders a candidate pool by logical
  entailment. Two searches, both correct
  (Theorems~\ref{thm:weakest-correct} and~\ref{thm:strongest-correct}), find
  the weakest candidates sufficient for the statement and the strongest ones
  it makes necessary, which together \emph{bracket} it. Each search decides an
  edge with one formula, its co-residual or residual
  (Theorems~\ref{thm:main-ctx} and~\ref{thm:reverse}); both are fixed by the
  candidate pool alone, so one \sys{} serves every context
  (Lemma~\ref{lem:absolute}).

  \item \textbf{Ambiguity residue} (Definition~\ref{def:residue}) identifies what the
  candidate pool leaves undetermined by a formula. When the context is itself formalized, the
  solver decides the residue, and its unsatisfiability certifies that the context
  disambiguates the statement.

  \item \textbf{A machine-checked algebraic foundation}
  (Appendices~\ref{sec:heyting} and~\ref{sec:lean}) casts the two searches as the
  two adjoints of a bi-Heyting algebra, so each edge check is the
  \emph{tightest} question that decides its edge, and explains their asymmetry:
  the upward search is intuitionistically sound, the downward search needs Boolean
  structure. We machine-check the results in Lean over an arbitrary Boolean algebra.

  \item \textbf{A neurosymbolic realization and evaluation}
  (Theorems~\ref{thm:realizability} and~\ref{cor:accuracy};
  Section~\ref{sec:impl}). We use an LLM as an entailment
  oracle to drive the searches. A \emph{sound} oracle, one that accepts only
  checks that hold, makes the selection \emph{safe} regardless of the quality
  of the candidate pool; completeness is needed only for \emph{tightness}.
  Under coverage (Definition~\ref{def:coverage}), the search recovers the
  intent exactly (Theorem~\ref{thm:disjunctive-recovery}). On AWS cloud
  safety properties, \sys{} selection is safe in all $20$ expert-adjudicated
  cases, where majority vote and USC are unsafe in $25\%$ and $10\%$. An
  \emph{ablation} that keeps the oracle and model but removes the graph is
  unsafe in $25\%$ of the cases and rejects every candidate in another $30\%$: the
  safety comes from the entailment structure, not from access to an
  entailment oracle.
\end{itemize}

\section{Motivating Example}\label{sec:motivating}

We present an example by taking a safety statement given in natural language and autoformalizing it using \sysnames{}.
For clarity of exposition, we present here a simplified treatment of
the example, leaving more detailed versions to Appendix~\ref{sec:motivating-app} and
Appendix~\ref{app:generated}, where we also include more details
about the mechanics of search steps with mathematical
residuals and co-residuals.

Consider a security-policy statement $s$:
 \textbf{``All requests to the payment service must be authenticated.''}
The statement stays in natural language throughout this section. Formalizing it is the
problem, so we cannot exhibit it as a formula without naming the answer in advance.
Each verdict below is therefore supplied by an \emph{entailment oracle}: an LLM given
the statement and the context in natural language and asked one entailment question
per edge, about that edge's residual or co-residual. This is a role distinct from the
LLM that produces the candidates. Everything else---the graph, the residuals, and the
satisfiability checks---is the solver's, and
Section~\ref{sec:realizability} states what we assume of the oracle.
When asked to formalize this statement, an LLM may interpret it in multiple ways, based on its knowledge of the following authentication mechanisms:
\begin{itemize}
  \item \textbf{Mutual Transport Layer Security (mTLS):} A cryptographic protocol in which both the client and the server present certificates to each other, providing strong identity verification at the transport layer.
  \item \textbf{JSON Web Token (JWT):} A signed token that carries claims (e.g., ``this client is authorized for payment operations''). It provides application-level proof of authorization beyond transport-level identity.
  \item \textbf{IP allowlist:} A network-level control that restricts access to requests originating from pre-approved IP addresses.
\end{itemize}

For the example, suppose that an LLM produces the following pool of candidates,
when asked to formalize the statement. Without the structure that \syss{} provide, only probabilistic heuristics exist to distinguish between the candidate formulas.
\begin{itemize}
  \item $\candTJ$: All requests must use mutual TLS \textbf{and} include a signed JWT bearing a payment-scope claim. (Transport-level identity plus application-level authorization.)
  \item $\candTI$: All requests must use mutual TLS \textbf{and} originate from an allow-listed IP address. (Transport-level identity plus network-level restriction.)
  \item $\candT$: All requests must use mutual TLS. (Transport-level identity only.)
  \item $\candCr$: All requests must include valid credentials (API key, token, or certificate). (Any form of authentication.)
\end{itemize}

Each candidate is a closed first-order sentence, and we name its parts
for use below. Writing $P(r)$ for ``$r$ is a request to the payment service'' and
$\mathit{mtls}$, $\mathit{jwt}$, $\mathit{ipok}$, $\mathit{cred}$ for the
mechanism predicates, let
\[
\mathsf{TLS} \;\equiv\; \forall r.\; P(r) \Rightarrow \mathit{mtls}(r),
\]
and let $\mathsf{JWT}$, $\mathsf{IP}$, $\mathsf{CRED}$ abbreviate the same shape
over the remaining predicates. The pool is then
$\candTJ = \mathsf{TLS} \land \mathsf{JWT}$,
$\candTI = \mathsf{TLS} \land \mathsf{IP}$,
$\candT = \mathsf{TLS}$, and $\candCr = \mathsf{CRED}$.
Because $\forall$ distributes over $\land$, presenting the candidates over these
abbreviations is an \emph{equivalence}, not a propositional simplification: the
quantifiers are visible to the entailment checks, not to the graph, the searches,
or the residue, none of which inspect the structure of a candidate. These are the
formulas our pipeline emits; Appendix~\ref{app:generated} gives them verbatim.

\input{fig-ig-payment-short}

These candidates vary in their logical strength:
\[
\begin{array}{ll}
  \candTJ \Rightarrow \candT \Rightarrow \candCr \quad & \text{and}
\\  
\candTI \Rightarrow \candT
\end{array}
\]
We represent this ordering as an \sysname{} as shown in  Figure~\ref{fig:ig-example}, with vertices representing candidates and edges representing strict entailment (stronger $\to$ weaker).
In Figure~\ref{fig:ig-example}, we include, for simplicity of exposition, only the \emph{covering} edges---the Hasse diagram of the entailment order. The \sysname{} itself contains an edge for every strictly comparable pair (e.g., $\candTJ \to \candCr$, $\candTI \to \candCr$ are also edges).
%

Given the \sysname{}, we seek the
weakest sufficient conditions, which entail the statement, and the strongest
necessary conditions, which the statement entails.
Such conditions can be used in program verification, and also more
generally to determine how best to ensure a guarantee and what to
assume from a guarantee.
For example, a \emph{provider}, such as a payment service provider,
can \emph{guarantee} the statement by enforcing the weakest sufficient
condition, and a \emph{customer} who uses the service can safely
assume the \emph{strongest} necessary property.

We use a graph-search over the \sysname{} to find
weakest/strongest sufficient/necessary conditions.
Each search walks the entailment order while conducting a
\emph{local} entailment check on the minimal mathematical formula that
separates the two adjacent candidates as defined by the edge.

The \textbf{downward search} finds the \emph{sufficient frontier}
$F_\downarrow$ consisting of the weakest candidates that entail $s$ in
the context, i.e., the candidates that satisfy $s$.
It starts from
the roots that entail $s$ in the context and walks down each edge
(stronger $\to$ weaker), asking the entailment oracle at each step: \emph{do the
extra cases the weaker candidate now admits still entail $s$?}
We prove that those extra cases are exactly the edge's logical
\emph{co-residual} (Section~\ref{sec:ig}).
The search stops when it can no longer traverse an edge, yielding a
set of weakest candidates $F_\downarrow$ that ensure sufficiency.

The \textbf{upward search} finds the \emph{necessary frontier}
$F_\uparrow$ consisting of the strongest candidates that $s$ entails
in the context, i.e., the candidates entailed by $s$.
It starts from the sinks that $s$ entails in the context and walks up
each edge (weaker $\to$ stronger), asking the oracle at each step: \emph{does $s$
guarantee the extra content the stronger candidate adds?}
We prove that the extra content is exactly the edge's logical
\emph{residual} (Section~\ref{sec:ig}).
The search stops when it can no longer traverse an edge, yielding a
set of strongest candidates $F_\uparrow$ that ensure necessity.

The ambiguity bracket between these two frontiers contains the
excluded candidates, which either over-commit or under-commit.

\subsection{Application 1: PCI-DSS Compliant Gateway}

The statement to formalize is the one above. The context supplies three facts:
\begin{enumerate}
  \item the service processes credit card transactions, so payment traffic occurs:
    $\exists r.\; P(r)$;
  \item it runs on an open network, where no allowlist covers all legitimate traffic:
    $\exists r.\; P(r) \land \lnot \mathit{ipok}(r)$;
  \item it must comply with the Payment Card Industry Data Security Standard
    (PCI-DSS), which requires proof of payment-specific authorization.
\end{enumerate}
The first two constrain the situations, and a solver can use them as written. The
third constrains what the statement demands of a configuration, and formalizing it
would mean fixing what ``authenticated'' denotes---which is the formalization we are
trying to select. It stays in natural language, and the oracle reads it.

\paragraph{Downward search (sufficient frontier).} The search starts from the roots
the oracle accepts as entailing $s$.
Fact~(2) contradicts
$\candTI = \forall r.\, P(r) \Rightarrow (\mathit{mtls}(r) \land \mathit{ipok}(r))$,
so $\candTI \land \mathit{ctx}$ is unsatisfiable---a satisfiability question the
solver settles, with no appeal to the statement. It therefore entails $s$ only
vacuously, and the search skips it as redundant (Remark~\ref{rem:vacuous}). That
leaves $\candTJ$ as the only usable root.
From $\candTJ$ the only weakening drops the payment-scope JWT to reach $\candT$. The oracle rejects that edge check, because fact~(3) requires proof of payment-specific authorization, so the search cannot descend. It settles at $F_\downarrow = \{\candTJ\}$.

\paragraph{Upward search (necessary frontier).} The search starts from the sink $\candCr$, which the oracle accepts as entailed by $s$ (authentication requires at least valid credentials). The oracle accepts the next two residual checks as well, since the context forces strong transport-level identity and then payment-specific authorization, so the search climbs $\candCr \to \candT \to \candTJ$; it does not climb toward $\candTI$, since IP allowlisting cannot be assumed. It settles at $F_\uparrow = \{\candTJ\}$.

\paragraph{The ambiguity bracket.} Both frontiers coincide at $\candTJ$. The gap is zero---the PCI-DSS context fully disambiguates the statement. This is the ideal outcome and the statement has a single contextual meaning.

\subsection{Application 2: Defense-in-Depth Requirement}

The statement is again the one above. Here the context supplies a single fact: the
service operates in a regulated environment requiring defense-in-depth, so at least
two independent authentication layers must be present.
Unlike Application~1, this context constrains no situation---it rules out no
configuration as impossible---and so has no formal part. It constrains only what the
statement demands, and every verdict below rests on the oracle reading it.

\paragraph{Downward search (sufficient frontier).} The oracle accepts both roots $\candTJ$ and $\candTI$ as entailing $s$: each supplies two independent layers. Neither can weaken to $\candT$, because dropping either second mechanism leaves only mTLS, a single layer that violates defense-in-depth, so the oracle rejects both edge checks. The search settles at $F_\downarrow = \{\candTJ, \candTI\}$.

\paragraph{Upward search (necessary frontier).} The search starts from the sink $\candCr$, which the oracle accepts as entailed by $s$ (a regulated environment requires standing controls). It climbs to $\candT$, since defense-in-depth forces at least one mechanism like mTLS, but climbs no further: the oracle rejects both remaining checks, because the statement forces \emph{some} second layer without fixing which, so neither the JWT of $\candTJ$ nor the IP allowlist of $\candTI$ is entailed. It settles at $F_\uparrow = \{\candT\}$.

\paragraph{Defining ambiguity.} The ambiguity bracket is $[\{\candTJ, \candTI\}, \candT]$: either two-layer configuration suffices, and the statement guarantees at least $\candT$. With the abbreviations above,
$\varphi_\downarrow = \candTJ \lor \candTI$ and $\varphi_\uparrow = \candT$, so
the ambiguity residue is:
%
%
\[
\mathrm{Amb} \;=\; \varphi_\uparrow \land \lnot\varphi_\downarrow
\land \mathit{ctx} \;=\; \mathsf{TLS} \land \lnot\mathsf{JWT} \land
\lnot\mathsf{IP} \land \mathit{ctx}.
\]
The $\mathit{ctx}$ conjunct does not unfold here: this context has no formal part, so
it constrains the residue only through the oracle that read it, and the solver cannot
decide satisfiability on its own---the limitation Section~\ref{sec:realizability}
returns to.
Unfolding the remaining abbreviations shows the residue is first-order: $\lnot\mathsf{JWT}$ is $\exists r.\; P(r) \land \lnot\mathit{jwt}(r)$,
so the residue says that every request carries mTLS while \emph{some} request
lacks a JWT and \emph{some} request falls outside the allowlist, but not
necessarily the same request (a distinction no propositional reading of the
bracket can draw). The residue holds exactly for configurations that pair transport-level identity with a second layer that is \emph{neither} $\candTJ$'s JWT nor $\candTI$'s allowlist. The dispute is not which configuration to pick but whether another mechanism outside the pool would work, which leads to the clarifying question: must the second layer be JWT or IP allowlisting, or does any independent layer suffice?

\section{\sysnames{}}\label{sec:ig}

Everything in this section is fixed by the candidate pool alone; the context enters
only in Section~\ref{sec:search}.
Given a set or pool of candidates, which are closed first-order
formulas, we can order the candidates by their strength defined by the
logical entailment $\models$ operator.
For closed formulas, entailment and implication coincide ($u \models
v$ iff $u \Rightarrow v$), allowing us to use implication
$\Rightarrow$ as the computable counterpart for entailment.
We first cluster the candidates into equivalence classes---formulas that mutually
entail each other---and identify each class with any one of its members, since
equivalent formulas are interchangeable for our purposes. The classes, not the raw
candidates, are the vertices; clustering is what makes the resulting order
antisymmetric and the graph acyclic.

\begin{definition}[\sysfull{}]\label{def:ig}
A \textbf{\sysfull{}}, or \textbf{\sysname{}} (\sys{}) for
short, for a set of candidate formulas is
the DAG whose vertices are the equivalence classes of candidates (under logical
equivalence) and whose edges represent strict entailment: there is an edge
$(u, v)$ whenever $u \models v$ and $v \not\models u$. The roots are the
strongest candidates.
\end{definition}

The \sys{} is \emph{transitively closed}: an edge is present for \emph{every}
strictly comparable pair, not only covering pairs as in a Hasse diagram. The
search algorithms rely on this: their frontier updates skip successors already
visited, and transitive closure guarantees that a direct edge to a not-yet-visited
witness always exists.

\paragraph{Completion.} We always complete the \sys{} with two synthetic vertices:
$\bot$, the ``False'' candidate, which entails every candidate and is
thus the unique strongest vertex---a \emph{universal root}; and $\top$, the
trivial ``True'' candidate, which every candidate entails and is thus the unique
weakest vertex---a \emph{universal sink}. This also unifies disconnected DAG components.
We omit $\bot$ and $\top$ from
the figures for clarity.

\paragraph{Residuals and co-residuals.} An edge $(u,v)$ is a weakening step:
moving from $u$ to $v$ discards exactly the content that made $u$ stronger. Two
dual quantities name that content---the \emph{residual}, which recovers $u$ from
the weaker $v$ by conjunction, and the \emph{co-residual}, which recovers $v$ from
the stronger $u$ by disjunction.

\begin{definition}[Residual]\label{def:residual}
Given $u \models v$, the \textbf{residual} of $v$ w.r.t.\ $u$ is the weakest formula
$\rho$ whose conjunction with $v$ recovers $u$, i.e., $v \land \rho \iff u$;
classically $\rho = v \Rightarrow u$.
\end{definition}

\begin{lemma}[Residual Decomposition]\label{lem:residual}
Let $u$ and $v$ be formulas with $u \models v$. In other words there is an edge from $u$ to $v$ in the \sysnoun{}. Then:
\[
u \;\iff\; v \land (v \Rightarrow u)
\]
Moreover, $v \Rightarrow u$ is the weakest formula $r$ satisfying $v \land r \models u$ and thus  $v \Rightarrow u$ is the residual of $v$ with respect to $u$.
\end{lemma}

\begin{definition}[Co-residual]\label{def:coresidual}
Given $u \models v$, the \textbf{co-residual} of $u$ w.r.t.\ $v$ is the strongest
formula $\delta$ whose disjunction with $u$ recovers $v$, i.e., $u \lor \delta \iff v$;
classically $\delta = v \land \lnot u$.
\end{definition}

\begin{lemma}[Co-residual Decomposition]\label{lem:coresidual}
Let $u$ and $v$ be formulas with $u \models v$. In other words there is an edge from $u$ to $v$ in the \sysnoun{}. Then:
\[
v \;\iff\; u \lor (v \land \lnot u)
\]
Moreover, $v \land \lnot u$ is the strongest formula $d$ satisfying $v \models u \lor d$ and thus $v \land \lnot u$ is the co-residual of $u$ with respect to $v$.
\end{lemma}

The residual and co-residual are not arbitrary: both are \emph{adjoints}, and that is what grounds the
searches in logic. Formulas modulo equivalence, ordered by entailment, form a
Boolean---hence Heyting---algebra, and in it the residual is exactly the right
adjoint of conjunction, characterized by
\[
\rho \,\models\, (v \Rightarrow u) \quad\text{if and only if}\quad \rho \land v \,\models\, u,
\]
while the co-residual is the order-dual left adjoint of disjunction
(Appendix~\ref{sec:heyting}). Being adjoints, they are unique and extremal: the
residual is the \emph{weakest} content that strengthens $v$ back to $u$ and the
co-residual the \emph{strongest} content that weakens $u$ out to $v$, so each
edge check below asks the tightest question that decides its edge. No weaker
question suffices, and no stronger one is needed. The two sides are not quite
symmetric, and the algebra says why: the necessity direction is intuitionistically
sound, whereas sufficiency rests on Boolean complementation
(Appendix~\ref{sec:heyting}).
\section{\sysname{} Search}\label{sec:search}

A \emph{context} is background that disambiguates the statement, in whole or in part.
The statement is what must hold; the context is what may be assumed to hold already.
Every check conjoins the context on the left of the entailment, so a candidate need
only track the statement among the situations the context admits.
A formal context can carry facts about the situations themselves; facts about what the
statement demands are why a context is in general natural language, and why
formalizing one does not make the checks solver-decidable
(Section~\ref{sec:realizability}).

We now fix a context $\mathit{ctx}$ and a statement $s$, both as formulas, and define
the checks that drive the two searches below over the graph just built. The graph is
not rebuilt: it was constructed without reference to any context, and it is reused as
is.
Both adjoints likewise stay as they are. The co-residual $v \land \lnot u$ and the
residual $v \Rightarrow u$ are fixed by the edge and mention no context; the context
enters only through the entailment in which they sit. One might expect instead to need
the adjoints of the context-relativized order, which would make the graph depend on
the context and so have to be rebuilt for each one. Lemma~\ref{lem:absolute} shows
that this buys nothing: relativizing either adjoint leaves both checks unchanged.
One \sys{} therefore serves every
context---Section~\ref{sec:motivating} runs two contexts over the same pool.
Because $\bot \land \mathit{ctx} \models s$ and $s \land \mathit{ctx}
\models \top$ hold unconditionally, $\bot$ is always a sufficient candidate and
$\top$ always a necessary one. Completion is therefore a simplification rather than a
change of results: a search returns $\bot$ only when it is the sole sufficient vertex,
that is, when no real candidate is sufficient, and dually for $\top$. The mechanization
(Appendix~\ref{sec:lean}) records this as
\texttt{frontierDown\_nonempty}/\texttt{frontierUp\_nonempty}.
The order the context induces is another matter. It is not a harmless re-indexing of
the absolute one: it can collapse a candidate to $\bot$
(Remark~\ref{rem:vacuous}) and, as Section~\ref{sec:ambiguity} shows, identify
distinct candidates or reverse their absolute order. None of this touches the graph.

\begin{lemma}[Absolute adjoints suffice]\label{lem:absolute}
Let $(u,v)$ be an edge and $\mathit{ctx}$ a context. The residual of the edge in the
context-relativized order---the weakest $\rho$ with
$\rho \land v \land \mathit{ctx} \models u$---is $(v \land \mathit{ctx}) \Rightarrow u$,
and the relativized co-residual is $v \land \mathit{ctx} \land \lnot u$. Substituting
either for its absolute counterpart leaves the corresponding check unchanged:
\begin{align*}
s \land \mathit{ctx} \models (v \Rightarrow u)
  &\iff s \land \mathit{ctx} \models \bigl((v \land \mathit{ctx}) \Rightarrow u\bigr),\\
(v \land \lnot u) \land \mathit{ctx} \models s
  &\iff (v \land \mathit{ctx} \land \lnot u) \land \mathit{ctx} \models s.
\end{align*}
\end{lemma}

The following theorems reduce a global entailment question to local
checks on the single formula separating adjacent candidates.

\begin{theorem}[Edge Traversal for Sufficiency]\label{thm:main-ctx}
Let $(u,v)$ be an edge, $\mathit{cores} = v \land \lnot u$ its co-residual, and
$\mathit{ctx}$ a
context. If $u \land \mathit{ctx} \models s$, then
$v \land \mathit{ctx} \models s$ iff $\mathit{cores} \land \mathit{ctx} \models s$.
\end{theorem}

\begin{theorem}[Edge Traversal for Necessity]\label{thm:reverse}
Let $(u,v)$ be an edge, $\mathit{residual} = v \Rightarrow u$, and assume
$s \land \mathit{ctx} \models v$. Then
$s \land \mathit{ctx} \models u$ iff $s \land \mathit{ctx} \models \mathit{residual}$.
\end{theorem}

\begin{figure}[t]
\begin{algorithmic}[1]
\Procedure{SufficientFrontierSearch}{$G, s, \mathit{ctx}$}
  \State $\mathit{frontier} \gets \{\bot\}$ \Comment{universal root; always sufficient}
  \State $\mathit{visited} \gets \emptyset$
  \Repeat
    \State $\mathit{changed} \gets \mathbf{false}$
    \ForAll{$u \in \mathit{frontier}$}
      \State $\mathit{succ} \gets \{v \mid (u,v) \in \mathrm{out\text{-}edges}(u),\; v \notin \mathit{visited},\; (v \land \lnot u) \land \mathit{ctx} \models s\}$
      \If{$\mathit{succ} \neq \emptyset$}
        \State $\mathit{frontier} \gets (\mathit{frontier} \setminus \{u\}) \cup \mathit{succ}$
        \State $\mathit{visited} \gets \mathit{visited} \cup \{u\}$
        \State $\mathit{changed} \gets \mathbf{true}$
      \EndIf
    \EndFor
  \Until{$\neg\, \mathit{changed}$}
  \State \Return $(\mathit{frontier},\; \bigvee_{v \in \mathit{frontier}} v)$
\EndProcedure
\end{algorithmic}
\caption{Sufficient-Frontier Search. Given an \sysnoun{} $G$, a statement $s$, and a context $\mathit{ctx}$, returns the sufficient frontier---the weakest candidates that entail $s$ in context---and their disjunction.}
\label{fig:alg-weakest}
\end{figure}

\begin{figure}[t]
\begin{algorithmic}[1]
\Procedure{NecessaryFrontierSearch}{$G, s, \mathit{ctx}$}
  \State $\mathit{frontier} \gets \{\top\}$ \Comment{universal sink; always necessary}
  \State $\mathit{visited} \gets \emptyset$
  \Repeat
    \State $\mathit{changed} \gets \mathbf{false}$
    \ForAll{$v \in \mathit{frontier}$}
      \State $\mathit{pred} \gets \{u \in \mathrm{parents}(v) \mid u \notin \mathit{visited},\; s \land \mathit{ctx} \models (v \Rightarrow u)\}$
      \If{$\mathit{pred} \neq \emptyset$}
        \State $\mathit{frontier} \gets (\mathit{frontier} \setminus \{v\}) \cup \mathit{pred}$
        \State $\mathit{visited} \gets \mathit{visited} \cup \{v\}$
        \State $\mathit{changed} \gets \mathbf{true}$
      \EndIf
    \EndFor
  \Until{$\neg\, \mathit{changed}$}
  \State \Return $(\mathit{frontier},\; \bigwedge_{v \in \mathit{frontier}} v)$
\EndProcedure
\end{algorithmic}
\caption{Necessary-Frontier Search. Given an \sysnoun{} $G$, a statement $s$, and a context $\mathit{ctx}$, returns the necessary frontier---the strongest candidates that $s \land \mathit{ctx}$ entails---and their conjunction.}
\label{fig:alg-strongest}
\end{figure}

The traversal criteria drive two graph searches over the completed \sys{}, given a
statement $s$ and context $\mathit{ctx}$. The \textbf{sufficient-frontier search}
(downward, Figure~\ref{fig:alg-weakest}) weakens across edges while the
co-residual check passes, and returns the \emph{sufficient frontier}
$F_\downarrow$---the weakest candidates entailing $s$---together with its
disjunction $\varphi_\downarrow = \bigvee_{v \in F_\downarrow} v$. The
\textbf{necessary-frontier search} (upward, Figure~\ref{fig:alg-strongest}) is
dual: it strengthens across edges while the residual check passes, and returns
the \emph{necessary frontier} $F_\uparrow$ and its conjunction
$\varphi_\uparrow = \bigwedge_{u \in F_\uparrow} u$. We describe each in turn
below. Because $\bot$
and $\top$ are always present, both searches always have a start vertex and both
frontiers are nonempty; the sufficient frontier collapses to $\{\bot\}$ exactly
when no real candidate is sufficient, and dually $F_\uparrow =
\{\top\}$ when $s$ entails no real candidate. A search should report these cases as a ``no sufficient (necessary)
formalization'' status, not emit $\bot$ or $\top$ as a formula. $\bot$ is safe to
\emph{check}, since it rejects every candidate, but vacuous to
\emph{assume}, since every proof that assumes it passes; dually, $\top$ is safe to
assume but uninformative, and enforcing it admits every artifact.

\paragraph{Sufficient-frontier search.}

Suppose that we have a statement, expressed as a closed formula $s$, a completed \sys{}, and a context $\mathit{ctx}$. Our goal is to find a weakest vertex in the graph that entails $s$ within the provided context. The algorithm (Figure~\ref{fig:alg-weakest}) conducts a graph-search by maintaining a set of ``frontier'' vertices and a separate set of visited vertices. Initially, the frontier is the singleton $\{\bot\}$---the universal root, which is sufficient unconditionally---and the visited set starts out empty. (Completion is what lets the initialization be unconditional: the first weakening step out of $\bot$ has co-residual $v \land \lnot\bot \equiv v$, so it admits exactly the roots $r$ with $r \land \mathit{ctx} \models s$, recovering the usual root-based start.) At each step, the algorithm considers an edge $(u, v)$ such that $u$ is in the frontier and $v$ is not in the visited set. The algorithm computes the co-residual $v \land \lnot u$ and checks whether $(v \land \lnot u) \land \mathit{ctx} \models s$. If so, then the algorithm removes $u$ from the frontier and inserts it into the visited set. The algorithm then inserts $v$ into the frontier.

The search proceeds until the frontier stabilizes, meaning that after considering all out-edges of the vertices in the current frontier, the frontier remains the same. When the algorithm terminates, it returns the frontier.

\paragraph{Necessary-frontier search.}

The sufficient-frontier search finds the weakest candidate that entails $s$ (i.e., $v \models s$). But we may also want the reverse: the strongest candidate that $s$ entails (i.e., $s \models v$). Together, the two searches bracket the statement from both sides---one finds the ``floor'' and the other finds the ``ceiling.'' This search rests on the reverse traversal criterion established in Theorem~\ref{thm:reverse}.

The algorithm (Figure~\ref{fig:alg-strongest}) conducts a graph-search upward by maintaining a frontier and a visited set. Initially, the frontier is the singleton $\{\top\}$---the universal sink, which is necessary unconditionally---and the visited set starts out empty. (As in the downward case, the first strengthening step out of $\top$ has residual $\top \Rightarrow u \equiv u$, so it admits exactly the sinks $v$ with $s \land \mathit{ctx} \models v$, recovering the usual sink-based start.) At each step, the algorithm considers a reverse edge $(v, u)$---that is, an edge where $v$ is in the frontier and $u$ is a parent of $v$ not in the visited set. The algorithm checks if $s \land \mathit{ctx}$ entails the residual $v \Rightarrow u$. If so, by Theorem~\ref{thm:reverse}, $s \land \mathit{ctx} \models u$; the algorithm removes $v$ from the frontier, inserts it into the visited set, and inserts $u$ into the frontier.

The search proceeds until the frontier stabilizes. When it terminates, the frontier contains the strongest candidates that $s \land \mathit{ctx}$ entails.

\begin{theorem}[Sufficient-Frontier Search is Correct]\label{thm:weakest-correct}
The sufficient-frontier search returns weakest candidates each entailing $s$ in
context (no strict descendant does), and their disjunction $\varphi_\downarrow$ is
weaker than any individual returned candidate and still entails $s$ in context:
$\varphi_\downarrow \land \mathit{ctx} \models s$.
\end{theorem}

\begin{theorem}[Necessary-Frontier Search is Correct]\label{thm:strongest-correct}
The necessary-frontier search returns strongest candidates each entailed by
$s \land \mathit{ctx}$ (no strict ancestor is), and their conjunction
$\varphi_\uparrow$ is stronger than any individual returned candidate and is
still entailed: $s \land \mathit{ctx} \models \varphi_\uparrow$.
\end{theorem}

\begin{lemma}[Frontier Completeness]\label{lem:completeness}
$F_\downarrow$ is exactly the set of weakest elements of $\{w \in V(G) \mid w \land \mathit{ctx} \models s\}$: those that entail $s$ in context and have no strictly weaker member of the set below them. Dually, $F_\uparrow$ is exactly the set of strongest elements of $\{w \in V(G) \mid s \land \mathit{ctx} \models w\}$. Consequently the frontiers, and the aggregates $\varphi_\downarrow$ and $\varphi_\uparrow$, depend on $G$, $s$, and $\mathit{ctx}$ alone.
\end{lemma}

\begin{remark}[Contextually vacuous candidates]\label{rem:vacuous}
A candidate $c$ with $c \land \mathit{ctx}$ unsatisfiable is equivalent to $\bot$ in
context, so $c \land \mathit{ctx} \models s$ holds vacuously and $c$ can enter
$F_\downarrow$. Such candidates arise routinely, since a generator proposes
mechanisms the statement permits and the context may exclude some of them.
A vacuous $c$ does not corrupt the selected formula. Since $\varphi_\downarrow$ is a
disjunction, $(c \lor d) \land \mathit{ctx} \equiv d \land \mathit{ctx}$, so a vacuous
disjunct leaves it unchanged. Only a frontier whose every member is vacuous makes
$\varphi_\downarrow \land \mathit{ctx}$ unsatisfiable, and a search reports that as
``no sufficient formalization'' rather than emitting the formula, exactly as it does
for $F_\downarrow = \{\bot\}$.
A search may nonetheless skip such a candidate, and the frontiers as we define them
do. This loses nothing. Transitive closure gives every ancestor of $c$ a direct edge to each of
its descendants, so skipping $c$ leaves them reachable; and the checks out of $c$
coincide with the checks out of its ancestors, because $c \land \mathit{ctx}$
unsatisfiable gives $\mathit{ctx} \models \lnot c$ and hence
$(w \land \lnot c) \land \mathit{ctx} \equiv w \land \mathit{ctx}$ for every $w$. A
vacuous vertex is therefore \emph{redundant}: traversing $c$ reaches exactly the
candidates reached without it, and skipping it changes the frontier only by removing
$c$ itself. The solver decides the satisfiability when the context is
a formula; the oracle judges it when the context is natural language. Our
implementation omits this check, as it omits the analogous check on the roots and
sinks it seeds the searches with (Section~\ref{sec:realizability}).
The necessary side cannot be affected at all: if $s \land \mathit{ctx} \models c$ and
$c \land \mathit{ctx}$ is unsatisfiable, then $s \land \mathit{ctx}$ is unsatisfiable
too, so whenever the statement is satisfiable in context no vacuous candidate is
necessary. The asymmetry is what makes $\varphi_\uparrow$ safe as a conjunction,
which one vacuous member would otherwise destroy.
\end{remark}

\section{Bracketing and Defining Ambiguity}\label{sec:ambiguity}

Together the two frontiers \emph{bracket} $s$ in context:
$\varphi_\downarrow \land \mathit{ctx} \models s$ and
$s \land \mathit{ctx} \models \varphi_\uparrow$. They cannot cross---every
$v \in F_\downarrow$ satisfies $v \land \mathit{ctx} \models u$ for every
$u \in F_\uparrow$---and they are complete: $F_\downarrow$ and $F_\uparrow$
are exactly the weakest candidates sufficient for $s$ in context and the
strongest candidates $s$ entails in context, so the frontiers and their aggregates
$\varphi_\downarrow, \varphi_\uparrow$ are canonical invariants of
$(\text{\sys}, s, \mathit{ctx})$ (Appendix~\ref{sec:ambiguity-app}). The
region between them is the residual ambiguity, which we capture as a single
formula.

\begin{theorem}[Non-Crossing of Frontiers]\label{thm:non-crossing}
Let $F_\downarrow$ and $F_\uparrow$ be the frontiers returned by the downward and upward searches on the same \sys{}, statement $s$, and context $\mathit{ctx}$. Then for all $v \in F_\downarrow$ and $u \in F_\uparrow$:
\[
v \land \mathit{ctx} \;\models\; u.
\]
That is, in the context-relativized entailment order, every vertex of $F_\downarrow$ is at least as strong as every vertex of $F_\uparrow$, so the frontiers cannot cross. Moreover, if $u \models v$ (i.e., $u$ lies above $v$ in the \sys{}), then
\[
u \land \mathit{ctx} \;\iff\; v \land \mathit{ctx} \;\iff\; s \land \mathit{ctx},
\]
so any apparent inversion in the absolute order collapses: both candidates are contextually equivalent to the statement itself, and the bracket degenerates to a point.
\end{theorem}

\begin{remark}[Relativization can invert the absolute order]\label{rem:order-inversion}
Non-crossing holds in the context-relativized order, not the absolute one. Indeed, an $F_\uparrow$ vertex can be strictly stronger than an $F_\downarrow$ vertex in the absolute order: take candidates $u = p \land q$ and $v = p$ with $s = p$ and $\mathit{ctx} = q$. The downward search traverses the edge $(u, v)$ because the co-residual $p \land \lnot q$ is inconsistent with $\mathit{ctx}$ (so it vacuously entails $s$), giving $F_\downarrow = \{v\}$; the upward search climbs the same edge because $s \land \mathit{ctx} = p \land q$ entails the residual $\lnot p \lor q$, giving $F_\uparrow = \{u\}$. By Theorem~\ref{thm:non-crossing}, this situation is benign: $u$, $v$, and $s$ are all equivalent under $\mathit{ctx}$, so the context has fully disambiguated the statement and the bracket carries no residual ambiguity. This is a second way the relativized order departs from the absolute one (Section~\ref{sec:search}); as there, the \sys{} is untouched.
\end{remark}

\begin{definition}[Ambiguity Residue]\label{def:residue}
$\mathrm{Amb}(s; \sys{}, \mathit{ctx}) = \varphi_\uparrow \land \lnot \varphi_\downarrow
\land \mathit{ctx}$. The context \textbf{fully disambiguates} $s$ when the residue
is unsatisfiable; otherwise $s$ \textbf{remains ambiguous}. (Completion makes both
aggregates well defined without empty-set conventions: in the degenerate cases
$\varphi_\downarrow = \bot$ and $\varphi_\uparrow = \top$ are realized by the
universal root and sink.)
\end{definition}

The residue is exactly the region on which the pool is silent: the pool certifies every
model of $\mathit{ctx}$ as satisfying $s$ (it models $\varphi_\downarrow$) or as
violating $s$ (it falsifies $\varphi_\uparrow$), or leaves it in the residue.

\begin{proposition}[Trichotomy]\label{prop:trichotomy}
Every model of $\mathit{ctx}$ satisfies exactly one of:
\begin{enumerate}[label=(\roman*)]
  \item $\varphi_\downarrow$, and is then \emph{certified to satisfy} $s$;
  \item $\lnot \varphi_\uparrow$, and is then \emph{certified to violate} $s$;
  \item $\mathrm{Amb}(s; G, \mathit{ctx})$, and the pool then certifies it in neither direction.
\end{enumerate}
\end{proposition}

\begin{theorem}[Disambiguation Criterion]\label{thm:disambiguation}
The following are equivalent:
\begin{enumerate}[label=(\arabic*)]
  \item $\mathrm{Amb}(s; G, \mathit{ctx})$ is unsatisfiable;
  \item $\varphi_\uparrow \land \mathit{ctx} \models \varphi_\downarrow$;
  \item $\varphi_\downarrow \land \mathit{ctx} \;\iff\; s \land \mathit{ctx} \;\iff\; \varphi_\uparrow \land \mathit{ctx}$.
\end{enumerate}
In particular, when the residue is unsatisfiable the bracket determines the statement exactly: the output of either search expresses $s \land \mathit{ctx}$ up to equivalence.
\end{theorem}

\begin{corollary}[Inversions carry no ambiguity]\label{cor:inversion}
If some $u \in F_\uparrow$ lies above some $v \in F_\downarrow$ in the \sys{} (i.e., $u \models v$), then the residue is unsatisfiable.
\end{corollary}

\paragraph{When the criterion needs the context.} Deciding the residue
appears to require a formal $\mathit{ctx}$, but frequently it does not. If the two
frontiers share a candidate $c$, then $\varphi_\uparrow \models c$ (as
$c \in F_\uparrow$ and $\varphi_\uparrow$ is their conjunction) and
$c \models \varphi_\downarrow$ (as $c \in F_\downarrow$ and $\varphi_\downarrow$ is
their disjunction), so $\varphi_\uparrow \land \lnot\varphi_\downarrow$ is already
unsatisfiable and the residue is unsatisfiable whatever $\mathit{ctx}$ says. The
frontiers alone settle the criterion, and every disambiguated case in our
evaluation is of this kind (Section~\ref{sec:eval}). When the frontiers are
\emph{disjoint}, where the cases are in dispute,
$\varphi_\uparrow \land \lnot\varphi_\downarrow$ can be satisfiable, so whether the
residue is unsatisfiable depends on $\mathit{ctx}$. A formal context makes the
criterion decidable; with a natural-language context the solver returns no
verdict.
\begin{remark}[The residue localizes]\label{rem:single-edge}
When the gap is a single edge---$F_\downarrow = \{u\}$ and $F_\uparrow = \{v\}$ with $(u, v)$ an edge of the \sys{}---the residue is $v \land \lnot u \land \mathit{ctx}$: exactly that edge's co-residual, restricted to the context. In general the residue aggregates the region that neither search resolves, and decomposing it along the edges between the frontiers recovers the individual unresolved weakening steps---the natural targets for clarification questions.
\end{remark}

\begin{proposition}[Monotonicity]\label{prop:monotone}
Let $G$ and $G'$ be \sysnames{} over candidate pools $P \subseteq P'$, and let $\mathit{ctx}' \models \mathit{ctx}$. Then
\[
\mathrm{Amb}(s; G', \mathit{ctx}') \;\models\; \mathrm{Amb}(s; G, \mathit{ctx}).
\]
That is, adding candidates or strengthening the context can only shrink the residue: ambiguity is monotone non-increasing in both the pool and the context.
\end{proposition}

\begin{remark}[Relativity]\label{rem:relativity}
The definition is deliberately relative: to the candidate pool, which is a sampled approximation of the statement's interpretations, and---when the context is informal---to the entailment oracle. It is an epistemic, operational notion of ambiguity, in the same way that a version space~\cite{version-spaces} is relative to a hypothesis language. Proposition~\ref{prop:monotone} says the approximation errs in the safe direction: the residue computed from a finite pool over-approximates the undetermined region, and further sampling or further context only tightens it. Note also that the residue is computed by the solver alone from the two search outputs; no additional oracle calls are needed to decide whether a context fully disambiguates a statement.
\end{remark}

\section{Realizability}\label{sec:realizability}

\sysnames{} and graph searches let us select the
weakest sufficient and strongest necessary candidates from a pool of
candidate formulas, but they rely on a formal entailment check with respect
to a formal statement $s$.
In this section, we use \syss{} for autoformalization, by
using an \emph{entailment oracle}
implemented by an LLM that compares a formal candidate against
statement $\widetilde{s}$ in context $\widetilde{\mathit{ctx}}$, both
of which are given in natural language.
The oracle answers each check yes or no; Definition~\ref{def:oracle} states the
two conditions we may require of it, soundness and completeness. Throughout, a run is given a
natural-language statement $\widetilde{s}$ and context $\widetilde{\mathit{ctx}}$.
We write $s$ and $\mathit{ctx}$ for their \emph{faithful} formalizations, the
formulas that capture the statement's intent. The searches never see $s$ or
$\mathit{ctx}$; the oracle answers checks about them. No proof can show that a formula
captures an English sentence, so this faithful reading is the one
assumption we make about the formal side.

In this section we write $\widetilde{\mathit{ctx}}$ and $\widetilde{s}$ for the
context and statement as natural-language inputs, reserving $\mathit{ctx}$ and $s$
for a \emph{formal} context and statement in the logic, and we let $O$ be the
oracle, a function from the issued \emph{checks} to $\{\text{yes}, \text{no}\}$.
A \emph{check} is an entailment the oracle is asked to decide. A check
\emph{holds} or \emph{fails} of $(\mathit{ctx}, s)$; the oracle \emph{accepts} or
\emph{rejects} it. We call the act of issuing a check a \emph{query}.
A run of either search branches only on $O$'s boolean answers; the frontier reads
nothing else. Because completion makes both initializations unconditional
($\{\bot\}$ downward, $\{\top\}$ upward, requiring no oracle query), every check
the run issues is an \emph{edge check}---a check about a single edge---of one of
just two shapes: the
co-residual check that drives the downward search and the residual check that
drives the upward one---
\[
\begin{gathered}
(v \land \lnot u) \land \widetilde{\mathit{ctx}} \models \widetilde{s},
\quad
\widetilde{s} \land \widetilde{\mathit{ctx}} \models (v \Rightarrow u),
\end{gathered}
\]
in which the candidate parts ($u$, $v$) are formal but $\widetilde{s}$ and
$\widetilde{\mathit{ctx}}$ are the informal inputs the oracle must supply.
Throughout, an unqualified \emph{check} is an edge check; the one place a check of
another shape arises is a root-seeded initialization, which issues a
\emph{vertex}-level check instead (Appendix~\ref{sec:realizability-app}). (The
first edge steps out of $\bot$ and $\top$, whose co-residual and residual
reduce to $v$ and $u$ respectively, recover the usual initialization queries. Our current implementation does not issue these first-step queries: it seeds
the downward search with every root of the \sys{} and the upward search with every
sink, which assumes that every root entails $s$ in context, and dually for sinks.
When no root does, the returned frontier need not entail $s$. In our evaluation we
found these cases by manual review, in the $10$ scale-study pools with no sufficient
candidate, and counted them as abstentions; Appendix~\ref{sec:eval-judge} gives the
details.)
We show that when $O$ satisfies certain assumptions it drives the searches exactly
as a solver would on some formal pair $(\mathit{ctx}, s)$, and that when that pair
faithfully captures the natural-language inputs the resulting selection is safe.

That a formula faithfully renders an English sentence is the one assumption no
logical condition can discharge, and it is the only one we make about the formal
side; everything below is a property of the oracle alone. Neither soundness nor
completeness mentions the pool, and neither requires $s$ to be a Boolean
combination of the pool candidates, nor the oracle to \emph{know} $\mathit{ctx}$
or $s$---only to answer as though it did.

\begin{definition}[Sound and complete oracles]\label{def:oracle}
An oracle is \textbf{sound} if every check it \emph{accepts} holds of
$(\mathit{ctx}, s)$. It is \textbf{complete} if it accepts every check that holds
of $(\mathit{ctx}, s)$. Both are conditions on the checks a run actually issues.
\end{definition}

\begin{theorem}[Safety from a sound oracle]\label{thm:realizability}
If the oracle driving a run is sound, the selection is \textbf{safe}:
every sufficient-frontier candidate entails $s$ in $\mathit{ctx}$, hence is safe
to \emph{enforce}, and $s$ entails every necessary-frontier candidate, hence each
is safe to \emph{rely on}. Equivalently,
$\varphi_\downarrow \land \mathit{ctx} \models s$ and
$s \land \mathit{ctx} \models \varphi_\uparrow$. This assumes no property of the
pool, and does not require $s$ to be expressible from it. The frontiers need not
be the weakest sufficient, respectively the strongest necessary, candidates.
\end{theorem}

\begin{theorem}[Tightness from a complete oracle]\label{cor:accuracy}
If the oracle is moreover complete, the run returns exactly the
frontiers the oracle-free search returns on $(\mathit{ctx}, s)$, and by Frontier
Completeness (Lemma~\ref{lem:completeness}) they are \textbf{tight}: the
sufficient frontier is exactly the set of weakest candidates sufficient for $s$
in $\mathit{ctx}$, and the necessary frontier exactly the set of strongest
candidates $s$ entails in $\mathit{ctx}$.
\end{theorem}

Safety and tightness thus rest on opposite halves of the oracle's
behavior, and the two error directions trade one against the other rather than
both against safety. Accepting a check that fails can admit a violation; rejecting
one that holds merely halts a search early, returning a candidate stronger
(downward) or weaker (upward) than necessary---slack, but still safe. A sound
oracle therefore suffices for everything the safety claim needs, and completeness
buys only how tight the result is. Proofs are in
Appendix~\ref{sec:realizability-app}.

We emphasize that safety depends on \emph{no property of the candidate pool}. Theorems~\ref{thm:weakest-correct}
and~\ref{thm:strongest-correct} and Frontier Completeness
(Lemma~\ref{lem:completeness}) assume a formal $s$ and $\mathit{ctx}$ and no
oracle at all; Theorem~\ref{thm:realizability} adds a sound oracle and concludes
safety; Theorem~\ref{cor:accuracy} adds completeness and concludes tightness. Each
holds whether or not the pool \emph{covers} the intent, and indeed whether
or not any candidate is close to it. The pool's quality governs only how
\emph{tight} the frontiers are---a poor pool yields a slack frontier (in the
extreme, $\{\bot\}$), never an unsafe one. This is what separates \syss{} from
statistical selection, whose guarantees are only as good as the pool it ranks over.
Coverage is a separate premise that buys \emph{exact recovery}, not safety. The
oracle's soundness, by contrast, is an assumption: an edge check can pass vacuously,
which classical logic makes safe but uninformative, and which in a Heyting algebra
allows an unsafe traversal (Remark~\ref{rem:three-chain}).

\begin{remark}[What completeness buys]\label{rem:completeness-buys}
Weakness and strength are what require it. Both arguments
(Theorems~\ref{thm:weakest-correct} and~\ref{thm:strongest-correct})
derive their contradiction from an edge check that \emph{would} pass, which
presumes the oracle accepts it. Under a sound but incomplete oracle the frontiers
stay safe and may be slack: the sufficient frontier can sit strictly above the
weakest sufficient candidates, and the necessary frontier strictly below the
strongest necessary ones. Frontier Completeness
(Lemma~\ref{lem:completeness}) and exact recovery
(Theorem~\ref{thm:disjunctive-recovery}) consequently need completeness,
whereas Non-Crossing (Theorem~\ref{thm:non-crossing}), the residue trichotomy
(Proposition~\ref{prop:trichotomy}) and the Disambiguation Criterion
(Theorem~\ref{thm:disambiguation}) invoke only the two bracketing entailments
and so survive. The residue in particular \emph{over-approximates}: slack makes
$\varphi_\downarrow$ stronger and $\varphi_\uparrow$ weaker, so
$\varphi_\uparrow \land \lnot \varphi_\downarrow \land \mathit{ctx}$ can only
grow. An unsatisfiable residue therefore still certifies disambiguation, while a
satisfiable one may be spurious.
\end{remark}

\begin{remark}[Which oracle errors matter]\label{rem:error-direction}
Only accepting a check that fails can cost safety. A wrongly rejected downward
edge halts the descent and leaves a \emph{stronger} candidate---slack, but still
sufficient; a wrongly rejected upward edge leaves a \emph{weaker} one---less
informative, but still necessary. The one-sidedness of the guarantee is thus
matched by a one-sidedness in what the oracle must get right.
\end{remark}

\begin{remark}[Dropping faithfulness]\label{rem:some-reading}
If $O$ is sound for some formal pair other than the faithful one,
Theorem~\ref{thm:realizability} still applies to that pair: the frontiers
bracket the reading the oracle in fact answers for rather than the intended one,
so the guarantee is relative to that reading rather than to the intended one. It is the
faithfulness of $(\mathit{ctx}, s)$ that ties it to the statement the user wrote.
\end{remark}

We stress that Theorem~\ref{thm:realizability} uses no property of the candidate
pool: safety holds whether or not the pool \emph{covers} the intent (a premise
we introduce next). Coverage buys exact recovery---not
safety, which the frontiers already provide against the faithful intent whenever
the oracle is sound. A gap in the pool is therefore benign. With the safety objective, the downward search
brackets the intent from below, so a missing candidate can leave the frontier
\emph{stronger} than necessary, in the extreme collapsing to $\{\bot\}$. A
search should report this as ``no sufficient formalization'' rather than emit it as a
formula. In every case, the frontier never admits a violation the intent forbids, and by
monotonicity of the
residue (Proposition~\ref{prop:monotone}) enlarging the pool only tightens the
result. Oracle error is a separate failure mode, and the theory does not make it
benign: when the oracle wrongly admits a weakening, it is not sound for the
faithful pair, and Theorem~\ref{thm:realizability} no longer
applies. We observed this three times in the scale study
(Appendix~\ref{sec:eval-judge}).

The guarantee attaches to a \emph{class}, not a lucky guess: the same answer
sequence can realize distinct witness pairs, and the returned frontier brackets
the formal statement $s$ of \emph{every} pair consistent with the oracle's answers. The complementary
failure is answers realizable by no single reading at all, so the oracle is not
sound for any pair. This is partially checkable without the oracle:
re-running the four edge criteria on the returned frontier under solver semantics
tests whether the answers form a consistent fixpoint, flagging traversals no
co-residual licenses and climbs no residual supports. We have not run this check in our
evaluation.

\subsection{Exact Recovery Under Coverage}\label{subsec:exact-recovery}

Realizability gives safety while the oracle remains in the loop, answering the
informal queries; the frontiers \emph{bracket} the intent but need not
\emph{equal} it. We now add a premise about the pool under which the bracket
tightens to an exact equality---one frontier recovering the intent on the nose.

\paragraph{The coverage premise.} \sysnames{} never invent a
formalization; we are handed a pool of candidates and asked to \emph{choose}
among them. Implicit in this setup is a premise that
Definition~\ref{def:coverage} makes explicit: that the intended meaning of the statement, in
the given context, is one of the candidates the pool supplies, or, more
precisely, that it is bracketed by them.

We do not require the intended reading to be a single candidate, nor to coincide
with a frontier or to lie in the bracket the searches return---the bracket is an
\emph{output} of the search and cannot be a precondition on its input. Coverage is
not solver-checkable---it is a statement about the natural-language statement,
asserting that a witness \emph{exists}, not exhibiting one---and the searches never
construct $s^\star$: they run on the oracle's answers exactly as before. By
Realizability (Theorem~\ref{thm:realizability}), a sound oracle drives
the run precisely as the solver would on the formal statement $s$ realizing its
answers, and under coverage that $s$ is (equivalent to) a flat disjunction or
conjunction $s^\star$ of candidates. The witness thus plays a purely \emph{analytic}
role in what follows: it lets us reason about the frontier the oracle-driven run
returns, without ever being fed to the solver or formalized by hand.

\paragraph{From bracket to exact recovery.} Realizability certifies that the
frontiers \emph{bracket} $s^\star$; it does not say either frontier \emph{equals}
it. The flat shape of the witness is exactly what makes the bracket collapse onto
the intent---on the sufficient side for a disjunctive witness, and on the necessary
side for a conjunctive one.

\begin{definition}[Coverage]\label{def:coverage}
A pool covers the statement in context if the faithful $s$ is contextually
equivalent to a disjunction $\bigvee_i c_i$ (\textbf{disjunctive coverage}) or a
conjunction $\bigwedge_i c_i$ (\textbf{conjunctive coverage}) of pool candidates.
We call such a witnessing disjunction or conjunction a \emph{covering witness}
$s^\star$.
\end{definition}

\begin{theorem}[Exact recovery under coverage]\label{thm:disjunctive-recovery}
Under disjunctive coverage with witness $s^\star = \bigvee_i c_i$, the downward
search recovers the intent exactly: $\varphi_\downarrow \land \mathit{ctx} \iff
s^\star \land \mathit{ctx}$. Dually, conjunctive coverage yields
$\varphi_\uparrow \land \mathit{ctx} \iff s^\star \land \mathit{ctx}$ for the
upward search. By Theorem~\ref{thm:realizability} the oracle-driven run returns
the same frontier, so the conclusion holds for the pipeline even though
the solver never sees $s^\star$.
\end{theorem}

\paragraph{Why flat, and why not merely negation-free.} Definition~\ref{def:coverage}
restricts the covering witness to a flat disjunction or conjunction rather than an
arbitrary Boolean---or even an arbitrary \emph{monotone}---combination, and this
restriction is not incidental: it is exactly what exact recovery needs. The
frontiers are a \emph{flat} join of vertices ($\varphi_\downarrow$) and a
\emph{flat} meet of vertices ($\varphi_\uparrow$); neither captures a monotone
witness one level deeper, such as $(c_1 \land c_2) \lor c_3$ with $c_1 \land c_2$
equivalent to no vertex---$\varphi_\downarrow$ stays strictly below it,
$\varphi_\uparrow$ strictly above, and the residue is nonempty even though the
intent is fully expressible from the candidates and negation-free. So dropping
negation would not suffice; what recovers the intent is the flat disjunctive
(resp.\ conjunctive) shape itself. This is also the shape a safety intent naturally
takes---``any one of these sufficient configurations is acceptable'' is a
disjunction of candidates---so under disjunctive coverage the downward pipeline of
Section~\ref{sec:eval} is not merely safe but returns the intended reading exactly.

\begin{remark}[What coverage failure costs]\label{rem:no-coverage}
Coverage powers the exact-recovery results of this subsection, but---as the
paragraph following Theorem~\ref{thm:realizability} noted---it powers neither
safety nor the bracket:
when the intent is \emph{no} flat disjunction or conjunction of pool candidates,
the frontiers still bracket the intent for a sound oracle. What
degrades is twofold, and neither is safety. \emph{(i)} The bracket may be loose
or degenerate: no candidate need sit at the
intent, so the sufficient frontier can be strictly stronger than needed and, if
no real candidate entails the intent, collapse to $\{\bot\}$ ($\varphi_\downarrow = \bot$)---the
search's explicit report that the pool expresses no sufficient condition, not a
false one. \emph{(ii)} The residue over-approximates the true undetermined region
from outside, shrinking toward it as candidates are added
(Proposition~\ref{prop:monotone}).
\end{remark}

\section{Implementation and Evaluation}\label{sec:impl}

We implemented \sysnames{} and evaluated this approach by autoformalizing AWS cloud safety properties.
The implementation uses an LLM to generate candidate formulas in
SMT-LIB format~\cite{smtlib} from natural language (NL). It then uses an
SMT solver~\cite{z3,cvc5} to find the equivalence clusters, which
become the vertices, and entailment relations between vertices, which
become the edges in the \sys{}.
AWS publishes a JSON resource type schema for every resource CloudFormation
supports~\cite{cfn-schemas}. From these schemas, our tooling generates SMT-LIB
declarations for each resource type, the domain vocabulary in which every
candidate is written. Each
resource type's schema becomes an SMT-LIB datatype whose fields carry the
schema's base types (strings, integers, Booleans, nested objects, and lists).
Before generating candidates, an LLM agent builds each property's background context: it reads the
schemas and Terraform provider documentation for the resource types involved
and writes a natural-language \emph{context}, which the generator and the
oracle both receive.

Following Section~\ref{sec:realizability}, the implementation
supports the contextual entailment checks the graph-search
algorithms need, by combining LLMs with solver-computed residuals. We repeat each
check and decide it by strict majority; without a strict majority of ``yes'' votes
the search does not traverse, so oracle uncertainty makes the result more
conservative. Candidate generation, the oracle, and the audit judge of
Appendix~\ref{sec:eval-judge} all use the same model, Claude Opus~4.6, so they can
share a misreading of the statement. The implementation seeds the downward search
with the \sys{}'s roots and the upward search with its sinks, without an oracle check,
which assumes that every root entails $s$ in context
(Appendix~\ref{sec:realizability-app}).

\paragraph{What the oracle is asked.}
The solver computes the edge's residual and co-residual, and
checks with cvc5 that the residual recovers the stronger candidate. The oracle
receives both candidates and both formulas in SMT-LIB, with the statement and
its context in natural language (or formal expression if available). Figure~\ref{fig:oracle-interaction} shows one
complete interaction. The oracle first describes each candidate and their
difference in English, then decides the check by searching for a
counterexample: downward, a state that satisfies the co-residual but violates
the statement; upward, a state that satisfies the statement but falsifies the
residual. It accepts the check only if no disjunct of the formula yields one.

\begin{figure}[t]
\small
\textbf{Statement.} ``If \texttt{MetadataOptions.HttpProtocolIpv6} is set to
\texttt{enabled}, the instance must be a Nitro-based instance launched in a
subnet that supports IPv6.'' Edge $(A,B)$ of its candidate pool, with $A$ the
stronger; the downward search must decide whether to weaken to $B$.

\medskip
\begin{tabular}{@{}l@{\quad}p{0.78\textwidth}@{}}
\toprule
\textsc{solver} & Emits the edge's implication in SMT-LIB and discharges it with
\texttt{cvc5}:\newline
\texttt{(=> (forall ((j Int)) $\ldots$ HttpProtocolIpv6 $\ldots$) (forall ((j
Int)) $\ldots$ InstanceType $\ldots$))} \quad ($\approx$2.5\,kB elided)\newline
\textsc{valid}: $(B \land C) \Rightarrow A$. \\
\midrule
\textsc{oracle} & Receives both candidates in SMT-LIB and describes each in English.\newline
$A$: for every EC2 instance, if \texttt{HttpProtocolIpv6} is ``enabled'', then
(i) \texttt{InstanceType} is non-empty, (ii) \texttt{SubnetId} is non-empty, and
(iii) every subnet matching \texttt{SubnetId} has a non-empty
\texttt{Ipv6CidrBlock} or has \texttt{Ipv6Native} true.\newline
$B$: $\ldots$ then (iii) only. \\
\addlinespace
\textsc{oracle} & States the difference: ``$B$ would allow an instance with
\texttt{HttpProtocolIpv6} enabled but with an empty \texttt{InstanceType} and/or
an empty \texttt{SubnetId}.'' \\
\addlinespace
\textsc{oracle} & Judges it against: ``the statement says
\emph{the instance must be a Nitro-based instance}, so \texttt{InstanceType}
must be specified, and \emph{launched in a subnet that supports IPv6}, so
\texttt{SubnetId} must be specified. Both are required.'' \\
\addlinespace
\textsc{oracle} & Verdict \textsc{no}, unanimous over five samples. \\
\midrule
\textsc{search} & The search does not traverse the edge or weaken to $B$,
and $A$ remains on the sufficient frontier. \\
\bottomrule
\end{tabular}
\caption{One complete oracle interaction, logged. The left column is the
neurosymbolic boundary: the solver supplies the edge's implication and discharges
the symbolic part, and every judgment about meaning is the oracle's. The oracle
receives the candidates, residual, and co-residual as SMT-LIB formulas; for
readability we show its own English descriptions of them in place of the
formulas. Natural language abbreviated.}
\label{fig:oracle-interaction}
\end{figure}

Appendix~\ref{sec:eval-judge} reports the experimental configuration.

\subsection{Application: Cloud Security}
To evaluate \sysnames{}, we consider cloud security, specifically for Amazon Web Services (AWS), as an application domain.
The challenge is to verify that a piece of cloud infrastructure does not break a given security or safety property prior to deployment.

Engineers provision and maintain modern cloud infrastructure through Infrastructure-as-Code templates, or simply templates, which are declarative descriptions of the cloud infrastructure, and consist of \emph{resources} such as storage systems (e.g., S3), compute nodes (EC2), etc.
A single misconfigured resource in a template can cause the
entire infrastructure to fail or to become unsafe.
A \emph{safety property} is a condition on a template's resource
configuration that must hold for the deployment to succeed.
Because the cost of safety violations is high, cloud providers have
created large repositories of \emph{safety properties} written in
NL, e.g., AWS Well-Architected
Framework~\cite{aws-well-architected} prescribes thousands of such
properties.
Verifying that a template upholds a safety property involves
translating the safety property into a formal language such as
SMT-LIB, which our implementation supports, and then checking it against
the template.
In our evaluation, we focus on the formalization, the first step of this verification process.

Because the safety properties abound, evolve, and are written in NL,
the only viable solution for verification is to autoformalize them
into a formal language such as SMT.
In this setting, an unsafe autoformalization is unacceptable, because
it can compromise safety.
Our safe autoformalization framework requires that we select a sufficient autoformalization for the intent.
In our evaluation, we therefore select the result formula from the sufficient frontier, whose candidates entail
the safety property.
We compare this selection to majority vote over equivalence
clusters (MV)---the selection rule that the clustering behind semantic
entropy~\cite{semantic-entropy} suggests---and to Universal Self-Consistency~\cite{usc}
(Section~\ref{sec:eval}).

\subsection{Case study: API Gateway Logging}

In our experiments, we encountered many examples where \sysnames{} ensure safe
autoformalization whereas purely statistical approaches result in
unsafe selections.
Here we present one such example, which we chose due to its
simplicity.

Consider the deployment safety property from AWS's CFN Guard: \textit{``All methods in Amazon API Gateway stage has logging enabled. The rule is NON\_COMPLIANT if logging is not enabled.''} for the resource \texttt{AWS::ApiGateway::Stage}.

Our implementation derives 10 candidate formalizations $c_1, \ldots, c_{10}$ for
this property and clusters them based on equivalence, resulting in 4
clusters with a measured semantic entropy of 1.846 bits.
\[
\resizebox{\fitcol}{!}{$[0]=\{c_1,c_2,c_{10}\},\quad [1]=\{c_3\},\quad [2]=\{c_4,c_8\},\quad [3]=\{c_5,c_6,c_7,c_9\}.$}
\]

If we selected the candidate from the largest cluster, we would
choose Cluster 3 and ignore all structural relationships between clusters.
Using \syss{}, the choice is different. After building an
\sys{}, we observe that the candidates form a single connected
component.
%
%
The members are
$[0],[1],[2],[3]$ with entailment edges:
\[
[0]\to[1],[2],[3] \qquad [1]\to[2] \qquad [3]\to[1],[2].
\]

To compare the candidates, we abbreviate two
first-order sentences: $M$ (``the stage configures at least one method'') and
$L$ (``every configured method has logging enabled'').
The generated candidates are quantified formulas over the template's
resource lists, indexed by integers. Writing $\mathit{ms}$ for the stage's
method list, we have
\[
M \equiv \exists i.\; 0 \le i < |\mathit{ms}|
\quad\text{and}\quad
L \equiv \forall i.\; 0 \le i < |\mathit{ms}| \Rightarrow \mathit{logged}(\mathit{ms}[i]),
\]
with the index guards and list accessors that the pipeline emits
(Appendix~\ref{app:generated}). As in Section~\ref{sec:motivating}, the names
abbreviate sentences; cvc5 discharged the cluster and edge checks on the
quantified formulas.
The formulas for Clusters~[0]~and~[3] turn out to be:
\[
\varphi_{[0]} \;=\; M \land L
\qquad\text{and}\qquad
\varphi_{[3]} \;=\; M \Rightarrow L
\]
$[0]\to[3]$ is a valid edge since $M\land L \models M \Rightarrow L$.
The downward search cannot traverse $[0]\to[3]$ (nothing licenses reading a stage with no configured methods as compliant) and the upward search cannot climb $[3]\to[0]$ (the statement does not force any method to exist).
The frontiers are therefore $F_\downarrow=\{[0]\}$ and $F_\uparrow=\{[3]\}$, giving $\varphi_\downarrow = M\land L$ and $\varphi_\uparrow =  M \Rightarrow L$.

The difference between the two is exactly the ambiguity (Section~\ref{sec:ambiguity}), which stems
from the fact that we do not know whether API Gateway must be used:
\begin{itemize}
  \item \textbf{[0] enforces:} ``You MUST configure logging for your ApiGateway methods.''
  \item \textbf{[3] allows:} ``If you have ApiGateway methods configured, they must have proper logging, but having no method configuration is acceptable.''
\end{itemize}

The ambiguity residue, which is exactly this edge's co-residual
(Remark~\ref{rem:single-edge}), makes this precise:
\[
\begin{aligned}
\mathrm{Amb} &= \varphi_\uparrow \land \lnot\varphi_\downarrow \land \mathit{ctx}\\
&= (\lnot M \lor L) \land \lnot(M \land L) \land \mathit{ctx}\\
&= \lnot M \land \mathit{ctx}.
\end{aligned}
\]
Nothing in the context rules out a stage with no configured methods, so the residue is
satisfiable and the statement remains ambiguous, localized to ``no method''
($\lnot M$, i.e.\ $|\mathit{ms}| = 0$).
Rather than a scalar ($1.846$ bits), the residue is a formula that
names the ambiguity: \emph{should a stage that configures no methods
be compliant or not?}

\paragraph{Resolving ambiguity with intent.} On the AWS cloud, a stage with no logging configured is a \emph{violation}. The intended and \emph{safe} formalization is thus the strong reading $s^\star = \varphi_{[0]} = M \land L$, which coincides with the sufficient frontier $\varphi_\downarrow$.
Whereas the \sys{} returns exactly the correct formalization, MV diverges here because cluster $[3]$ (of
cardinality $4$) outweighs cluster $[0]$ (cardinality $3$), leading to
an unsafe formalization.
The residue $\lnot M$ is exactly where MV's choice, the largest cluster, disagrees with the intent.
The two largest clusters, $[3]$ and $[0]$, are exactly the disputed pair, so a
selector that compares only the two largest clusters could get this example right.
It would not certify that its pick suffices, nor produce the residue that names the
dispute.

\subsection{Experimental Evaluation}\label{sec:eval}
\input{new_eval.tex}


\section{Related Work}\label{sec:related-tight}\label{sec:related-app}

The approach presented in this paper builds upon and extends prior work in semantic entropy, autoformalization, natural language to formal logic translation, and candidate selection methods.
Before discussing these in more detail, we note that two established lines of
research use the names \emph{implication graph} and \emph{entailment graph} for
different notions.
In SAT solving, an implication graph records the literal
assignments forced by clauses during conflict-driven search and is analyzed to
derive learned clauses; its vertices are assignments, not formulas. In natural
language processing, entailment graphs order natural-language predicate
templates by entailment, with edges predicted from distributional evidence and
transitivity imposed as a global optimization. Our graphs order \emph{formulas},
their edges are decided by a solver rather than predicted, and transitive
closure therefore comes for free rather than as a constraint to enforce. We say
\emph{logical} entailment graph where the distinction matters.

\paragraph{Semantic Entropy and Uncertainty Estimation.}
Kuhn et al.~\cite{semantic-entropy} introduced semantic entropy for detecting confabulations (hallucinations) in language models by clustering outputs by logical equivalence and measuring Shannon entropy over the resulting distribution; the approach was subsequently demonstrated as a general hallucination detector~\cite{semantic-entropy-nature}. Semantic entropy is an uncertainty signal and does not select a candidate. Our
majority-vote baseline uses the clustering behind this signal to select the largest
cluster, a rule Kuhn et al.\ do not propose. Follow-up work has addressed the computational overhead of this approach: Kossen et al.~\cite{semantic-entropy-probes} proposed Semantic Entropy Probes that approximate semantic entropy from a single generation's hidden states, reducing the 5-to-10x cost of generating multiple samples. Qiu and Miikkulainen~\cite{semantic-density} introduced Semantic Density, which provides response-wise (rather than prompt-wise) uncertainty quantification by measuring confidence from a probability distribution perspective in semantic space. Nikitin et al.~\cite{kernel-language-entropy} relax hard equivalence clusters to graded semantic kernels, and Lin et al.~\cite{blackbox-uq} derive black-box uncertainty from similarity graphs over sampled generations. Closest to our use of an entailment order, Mohri and Hashimoto~\cite{conformal-factuality} obtain conformal factuality guarantees by walking a partial order of progressively weaker output statements until a coverage target is met. Our work goes beyond equivalence-based clustering: we order candidates by logical entailment, enabling finer-grained reasoning about the relationships between candidate formalizations.  This approach allows using the partial order to define the ambiguity contextually and precisely and to resolve it directionally.
Barnaby et al.~\cite{choosedontlabel} group candidate programs by their behavior on
Hoare-triple queries, synthesize multiple-choice questions that split the groups,
and evaluate convergence to the intended program in four domains. They thus
implement and measure the clarification loop that we leave to future work. The two
instruments differ: they choose their question to split the candidate set, whereas
our residue is a single formula that characterizes the region the pool leaves
undetermined. A residue could seed such a question.

\paragraph{Autoformalization.}
Wu et al.~\cite{autoformalization-llm} demonstrated that LLMs can correctly translate 25.3\% of mathematical competition problems to formal specifications in Isabelle/HOL and raised the success rate on MiniF2F from 29.6\% to 35.2\% using autoformalized data. Patel et al.~\cite{autoformalization-subtasks} proposed decomposing autoformalization into subtasks (unlinked formalization, entity linking, and type adjustment), making the overall task more tractable. Poiroux et al.~\cite{autoformalization-eval} provided benchmarks and evaluation metrics for statement autoformalization, reporting up to 45.1\% accuracy on undergraduate mathematics but noting significant difficulties with research-level content. Most closely related to our selection problem, Li et al.~\cite{symbolic-consistency} score candidate autoformalizations by clustering them under prover-checked symbolic equivalence combined with embedding-based semantic consistency; \sysnames{} refine exactly this picture by ordering the equivalence classes by entailment and selecting within the resulting DAG through contextual entailment. Our evaluation compares against both of Li et al.'s components separately: the MV baseline clusters candidates by prover-checked equivalence and selects the largest cluster (the symbolic-equivalence ingredient), while the USC baseline lets the LLM pick the most self-consistent candidate (the semantic-consistency ingredient). Unlike both, \syss{} provide a sufficiency guarantee.

\paragraph{Natural Language to Formal Specifications.}
Chen et al.~\cite{nl2tl} achieved over 95\% accuracy in translating natural language to temporal logic using fine-tuned T5 models with under 10\% of baseline training data, publishing a dataset of 28K NL-TL pairs. Cosler et al.~\cite{nl2spec} presented nl2spec, an interactive framework that uses LLMs to translate unstructured natural language into temporal logic specifications, with a sub-formula mapping mechanism that enables iterative disambiguation. Beyond temporal logic, LLMs have been used to produce formal software specifications: postconditions from natural-language intent, evaluated by their strength and bug-discriminating power~\cite{nl2postcond}; JML and ACSL specifications iteratively refined against verifiers~\cite{specgen,autospec}; loop invariants filtered by SMT-based checks~\cite{loopy}; and consistency-checked code--specification--documentation triples~\cite{clover}. Closest to our running example, LLMs have been applied to formalize natural-language security and access-control policies into analyzable policy languages~\cite{cedar-policy}. These systems validate or repair individual candidates against a verifier; unlike these approaches that produce a single formalization, our method generates multiple candidates and exploits the logical structure \emph{among} them, using entailment ordering to select based on strength and context.


\paragraph{Disambiguation and Entailment.}
Liu et al.~\cite{ambient} showed that LLMs perform poorly at disambiguating natural language---GPT-4's disambiguations are judged correct only 32\% of the time, compared to 90\% for the human-written disambiguations in their dataset---and characterized ambiguity through its effect on entailment relations, where multiple valid interpretations yield different entailment outcomes. This directly connects to our approach: we use logical entailment to order candidate interpretations and identify semantic differences between them, enabling systematic disambiguation through graph traversal.

\paragraph{Candidate Selection and Ranking.}
Bertsch et al.~\cite{mbr-decoding} showed that Minimum Bayes Risk (MBR) decoding, which selects outputs based on the lowest expected error rather than the highest probability, provides principled candidate selection across a variety of tasks. Self-consistency~\cite{self-consistency} selects the majority answer among sampled reasoning paths; Chen et al.~\cite{usc} proposed Universal Self-Consistency (USC), which extends this to free-form generation by leveraging LLMs themselves to select the most consistent answer among multiple candidates. Our approach differs from both by using logical entailment structure rather than statistical consistency or expected error to guide selection, enabling guarantees about the strength of the selected candidate relative to the statement and context. Relatedly, IronSpec~\cite{ironspec} tests the reliability of a trusted formal spec by mutation, organizing the surviving (``alive'') mutants into a strength-ordered \emph{hierarchy} (a DAG) to classify them---the same structural device as our \sysname{}, which we instead build over candidate formalizations of an ambiguous statement and search to bracket the intended reading.

\paragraph{Classical Antecedents: Boundary Sets and Extremal Conditions.}
\syss{} recast autoformalization as a search for extremal conditions over an
LLM-generated candidate pool with a natural-language statement and context; the
underlying bracket structure has deep roots in symbolic AI and formal methods. In autoformalization, prior
work evaluates the strength of a formalization after the fact~\cite{nl2postcond};
we select by strength.
Lin~\cite{lin-snc-wsc} defined the strongest necessary and weakest sufficient
conditions of a proposition under a background theory. Our frontiers are the
finite-pool analogues, with the context playing the role of the background
theory: the sufficient frontier is a weakest sufficient condition for the
statement and the necessary frontier a strongest necessary condition, which is
what their names record.
The correspondence is to those targets and not to the method.
Lin does not select from a pool of candidates: his conditions are constructed over
an \emph{entire} target sub-language by eliminating the remaining
vocabulary~\cite{doherty-snc-wsc}, so his setting has no pool, no order on one, and
no search.
His setting is also formal throughout---the theory and the proposition are given as
formulas---so no step of his construction rests on a judgment that can be wrong.
Ours replaces the statement and the context with natural language and the
entailment test with an oracle, and our guarantee is correspondingly about oracle
error: a sound oracle makes the selection safe (Theorem~\ref{thm:realizability})
and a complete one makes it tight (Theorem~\ref{cor:accuracy}).
The two-boundary representation itself echoes Mitchell's version spaces~\cite{version-spaces}, which delimit the hypotheses consistent with observed data by most-specific ($S$) and most-general ($G$) frontiers of a generality order---though the settings differ: version spaces bound a single consistency set around an unknown target concept and shrink incrementally as examples arrive, whereas our frontiers are one-shot extremal conditions of a known statement under a fixed context. What does carry over is the semantics of the gap: in both settings, the region between the frontiers is exactly what the available information fails to determine, and both suggest a targeted query---a discriminating example there, a clarification of a specific co-residual here (which we have not yet evaluated). The tradition of extremal formulas also includes Dijkstra's weakest precondition~\cite{dijkstra-wp}, and Craig interpolation~\cite{craig-interpolation} produces a formula lying between two entailment-ordered formulas---the gap that our residuals and co-residuals characterize locally. Ordering hypotheses by generality and traversing the resulting lattice with refinement operators is central to inductive logic programming~\cite{plotkin-lgg,ilp-survey}. Finally, our co-residuals are a logical analogue of the behavioral deltas computed by differential symbolic execution between program versions~\cite{dse}. Relative to these classical lines, \syss{} operate over LLM-generated candidate pools with contexts stated in natural language, delegating entailment checks to SMT solvers or LLMs.

\paragraph{Anti-unification.}
Anti-unification finds what two formulas have in common. It computes the most
specific pattern both match, with a variable wherever they differ. For example, $f(a, g(b))$ and
$f(c, g(b))$ share the pattern $f(x, g(b))$. Plotkin introduced it as the least
general generalization~\cite{plotkin-lgg}. Applied to two clusters of candidates, it
shows exactly where they differ, with no oracle and no negation, and it also works
for incomparable clusters, which have no \sys{} edge and about which our searches say
nothing. Anti-unification modulo theories matches terms up to associativity,
commutativity and idempotence. The state of the art is the
\texttt{yaspar-org/semi-persistent} library~\cite{semi-persistent}, which implements
exact anti-unification modulo these theories and marks the differing subterms. Run
after semantic entropy it flags a disagreement.

\paragraph{Contracts.}
The provider--consumer duality echoes contract-based design~\cite{contracts}, but
the scalar order \syss{} use does not order contracts. A contract $(P, Q)$ refines
another when it weakens the assumption $P$ and strengthens the guarantee $Q$
relative to $P$. Refinement is contravariant in $P$ and covariant in $Q$, and the
two components are not independent. Running our searches on $P$ and on $Q$
separately is therefore not sound. Extending \syss{} to contracts needs a graph over the
refinement order itself, with a per-edge criterion for that order. Realizability
adds a further obstruction: a strong $Q$ under a weak $P$ may admit no
implementation, so a nonempty frontier can still be useless. We leave this extension
to future work.

\paragraph{Summary.}
Our work differs from all these approaches in four ways: (1) ordering candidates by entailment rather than only equivalence, (2) computing crisp semantic differences between adjacent candidates, (3) supporting contextual reasoning through logical entailment, and (4) providing algorithms that guarantee selection of the weakest satisfying formula---in effect reviving the classical extremal-condition representation for statements given in natural language.


\section{Conclusion}

In verification-oriented applications a formalization is used to check a
condition on an artifact, so adequacy is one-sided: a formula that is enforced
must be sufficient for the intent, and one that is relied upon must be
necessary for it.
\sysnames{} (\syss{}) order an LLM's candidate formalizations by logical
strength, with no reference to a context. Given a context, two searches over this
order return the weakest candidates that entail the statement (safe to enforce)
and the strongest it entails (safe to rely on); together they expose whatever
ambiguity remains as a single residue, itself a formula.
A solver supplies the order and the residuals, and an LLM decides only the
local edge checks. We prove that a \emph{sound} oracle suffices for safety
(Theorem~\ref{thm:realizability}); completeness buys only tightness
(Theorem~\ref{cor:accuracy}), so a wrongly \emph{accepted} check is the only
way safety can fail.
On AWS safety properties, \sys{} selection is safe in all $20$
expert-adjudicated cases, whereas the baselines are unsafe in $10$--$25\%$.
The same oracle without the graph is unsafe in $25\%$ of the cases and rejects
every candidate in another $30\%$. A $152$-statement scale study
corroborates the baseline figure: both select a candidate weaker than the
oracle certified in $22$--$24\%$ of statements.

\sysnames{} require three ingredients: a pool of candidate formalizations, an
entailment relation between candidates that a solver can decide, and an oracle
for the edge checks. Wherever these are available the safety guarantee holds.
Neither the theorems nor the Lean development make assumptions  about the
domain: the theorems are stated for an arbitrary candidate pool and entailment
order, the Lean proofs over an arbitrary Boolean algebra.

\section*{Acknowledgments}
We thank Daniel Sokolowski for early experimentation and for implementing 
an agentic framework that generated NL context for AWS cloud safety properties.
We thank Remi Delmas for feedback on drafts of this paper.

\bibliography{references}

\ifarxiv 
\clearpage
\onecolumn
\appendix


\section{\sysnames{}: Proofs}\label{sec:ig-app}

This appendix proves the decomposition lemmas of Section~\ref{sec:ig}. The
edge-traversal theorems, which involve a context, are proved with the searches in
Appendix~\ref{sec:search-app}.

\begin{proof}[Proof of Lemma~\ref{lem:residual}]
($\Rightarrow$) Assume $u$. Since $u \models v$, we have $v$. We also have $u$, so $v \Rightarrow u$ holds. Hence $v \land (v \Rightarrow u)$.

($\Leftarrow$) Assume $v \land (v \Rightarrow u)$. By modus ponens, $u$ follows.

For weakness: suppose $v \land r \models u$ for some $r$. Then for any model $M$ with $M \models r$: if $M \models v$ then $M \models v \land r$, so $M \models u$, hence $M \models v \Rightarrow u$; if $M \not\models v$ then $M \models v \Rightarrow u$ vacuously. So $r \models (v \Rightarrow u)$.
\end{proof}

The dual notion is the \emph{co-residual}. Where the residual is the weakest formula that recovers the stronger $u$ from the weaker $v$ by \emph{conjunction}, the co-residual is the strongest formula that recovers the weaker $v$ from the stronger $u$ by \emph{disjunction}.

\begin{proof}[Proof of Lemma~\ref{lem:coresidual}]
($\Leftarrow$) Assume $u \lor (v \land \lnot u)$. If $u$, then since $u \models v$ we have $v$; if $v \land \lnot u$, then $v$ directly. Either way $v$.

($\Rightarrow$) Assume $v$. By excluded middle either $u$ or $\lnot u$. If $u$, then $u \lor (v \land \lnot u)$; if $\lnot u$, then $v \land \lnot u$, so again $u \lor (v \land \lnot u)$.

For strength: suppose $v \models u \lor d$ for some $d$. Then for any model $M$ with $M \models v \land \lnot u$: since $M \models v$ we have $M \models u \lor d$, and since $M \not\models u$ we get $M \models d$. So $(v \land \lnot u) \models d$.
\end{proof}

The two notions are lattice duals: the residual residuates meet and the co-residual residuates join, so the co-residual is to the join what the residual is to the meet. Classically they are also related by negation, $v \land \lnot u = \lnot(v \Rightarrow u)$, but it is the join-residuation that makes the naming symmetric; the reliance of this negation identity---and of the excluded-middle step in the proof above---on Boolean logic is exactly what separates the two searches intuitionistically (Appendix~\ref{sec:heyting}).

\section{\sysname{} Search: Proofs}\label{sec:search-app}

This appendix proves that the absolute adjoints suffice, the two edge-traversal
criteria, and the two frontier searches correct, together with frontier
completeness.

\begin{proof}[Proof of Lemma~\ref{lem:absolute}]
For the residual, apply the adjunction, absorb the repeated conjunct, and apply the
adjunction again:
$s \land \mathit{ctx} \models (v \land \mathit{ctx}) \Rightarrow u$ iff
$(s \land \mathit{ctx}) \land (v \land \mathit{ctx}) \models u$ iff
$s \land \mathit{ctx} \land v \models u$ iff
$s \land \mathit{ctx} \models (v \Rightarrow u)$. Only the adjunction and the
idempotence of $\land\,\mathit{ctx}$ are used, so this half holds in any Heyting
algebra.
For the co-residual the two antecedents are equal:
$(v \land \mathit{ctx} \land \lnot u) \land \mathit{ctx} = (v \land \lnot u) \land
\mathit{ctx}$, again by idempotence. The relativized co-residual differs from the
absolute one only by a $\mathit{ctx}$ conjunct, which the check already supplies.
\end{proof}

\begin{proof}[Proof of Theorem~\ref{thm:main-ctx}]
Any model $M$ such that $M \models v \land \mathit{ctx}$ falls into exactly one of two cases: either $M \models u$, or $M \models v \land \lnot u = \mathit{cores}$.

\textbf{($\Leftarrow$)} Suppose $\mathit{cores} \land \mathit{ctx} \models s$. Let $M$ be any model with $M \models v \land \mathit{ctx}$. If $M \models u$, then $M \models u \land \mathit{ctx}$, so $M \models s$ by hypothesis. If $M \not\models u$, then $M \models v \land \lnot u \land \mathit{ctx} = \mathit{cores} \land \mathit{ctx}$, so $M \models s$ by assumption. In either case $M \models s$, hence $v \land \mathit{ctx} \models s$.

\textbf{($\Rightarrow$)} Suppose $v \land \mathit{ctx} \models s$. Let $M$ be any model with $M \models \mathit{cores} \land \mathit{ctx}$. Since $\mathit{cores} = v \land \lnot u$, we have $M \models v$, so $M \models v \land \mathit{ctx}$, hence $M \models s$. Thus $\mathit{cores} \land \mathit{ctx} \models s$.
\end{proof}

\begin{proof}[Proof of Theorem~\ref{thm:reverse}]
By the Residual Decomposition (Lemma~\ref{lem:residual}), $u \iff v \land (v \Rightarrow u) = v \land \mathit{residual}$. Thus $s \land \mathit{ctx} \models u$ iff $s \land \mathit{ctx} \models v \land \mathit{residual}$. Since we already have $s \land \mathit{ctx} \models v$, this reduces to $s \land \mathit{ctx} \models \mathit{residual}$.
\end{proof}

Theorem~\ref{thm:main-ctx} and Theorem~\ref{thm:reverse} make the residual/co-residual duality explicit: the downward search checks co-residuals ($v \land \lnot u$) to determine whether a weaker candidate still entails $s$, while the upward search checks residuals ($v \Rightarrow u$) to determine whether $s$ still entails a stronger candidate.

\begin{proof}[Proof of Theorem~\ref{thm:weakest-correct}]
\emph{Termination.} Every iteration of the outer loop that sets $\mathit{changed}$ moves at least one vertex into $\mathit{visited}$, and $\mathit{visited}$ never shrinks. Since the vertex set is finite, the loop terminates.

\emph{Safety.} We prove by induction that every vertex that ever enters the frontier entails the statement in context. Initially, the frontier is $\{\bot\}$, and $\bot \land \mathit{ctx} = \bot \models s$. Inductively, a vertex $v$ enters the frontier only across an edge $(u, v)$ where $u$ is a frontier vertex and $(v \land \lnot u) \land \mathit{ctx} \models s$; since $u \land \mathit{ctx} \models s$ by the inductive hypothesis, Theorem~\ref{thm:main-ctx} yields $v \land \mathit{ctx} \models s$. In particular, every returned vertex entails the statement in context.

\emph{Weakness.} We must show that no returned vertex has a strict descendant that entails the statement in context. The soundness invariant alone does not give this: the frontier update retains a vertex $u$ whenever $\mathit{succ} = \emptyset$, which can happen either because the edge checks of its out-neighbors fail \emph{or} because the passing out-neighbors are already visited---so we must rule out the latter situation. Suppose for contradiction that at termination some frontier vertex $u$ has strict descendants entailing $s$ in context, and let $m$ be a weakest such descendant (one exists because the graph is finite and acyclic). By transitive closure, $(u, m)$ is an edge of $G$. We claim $m \notin \mathit{visited}$: a vertex enters $\mathit{visited}$ only when, as a frontier vertex, it has an out-neighbor $w \notin \mathit{visited}$ with $(w \land \lnot m) \land \mathit{ctx} \models s$; since $m \land \mathit{ctx} \models s$, Theorem~\ref{thm:main-ctx} would give $w \land \mathit{ctx} \models s$ with $w$ a strict descendant of $m$---contradicting the choice of $m$. Moreover, the edge check for $(u, m)$ passes: $(m \land \lnot u) \land \mathit{ctx} \models m \land \mathit{ctx} \models s$. Hence $m \in \mathit{succ}$ when $u$ is processed in the final pass, so the frontier changes---contradicting stabilization.

Finally, the disjunction of the frontier vertices $v_1 \lor \cdots \lor v_k$ entails $s$ in context: $(v_1 \lor \cdots \lor v_k) \land \mathit{ctx}$ is equivalent to $(v_1 \land \mathit{ctx}) \lor \cdots \lor (v_k \land \mathit{ctx})$ by distribution, and each disjunct entails $s$ by soundness. And since each $v_i \models v_1 \lor \cdots \lor v_k$, the disjunction is weaker than any individual frontier vertex.
\end{proof}

\begin{proof}[Proof of Theorem~\ref{thm:strongest-correct}]
\emph{Termination.} Every iteration of the outer loop that sets $\mathit{changed}$ moves at least one vertex into $\mathit{visited}$, and $\mathit{visited}$ never shrinks. Since the vertex set is finite, the loop terminates.

\emph{Safety.} We prove by induction that $s \land \mathit{ctx}$ entails every vertex that ever enters the frontier. Initially, the frontier is $\{\top\}$, and $s \land \mathit{ctx} \models \top$. Inductively, a vertex $u$ enters the frontier only across a reverse edge $(v, u)$ where $v$ is a frontier vertex and $s \land \mathit{ctx} \models (v \Rightarrow u)$; since $s \land \mathit{ctx} \models v$ by the inductive hypothesis, Theorem~\ref{thm:reverse} yields $s \land \mathit{ctx} \models u$. In particular, $s \land \mathit{ctx}$ entails every returned vertex.

\emph{Strength.} We must show that no returned vertex has a strict ancestor entailed by $s \land \mathit{ctx}$. As in the downward search, the soundness invariant alone does not give this: the frontier update retains a vertex $v$ whenever $\mathit{pred} = \emptyset$, which can happen either because the residual checks of its parents fail \emph{or} because the passing parents are already visited---so we must rule out the latter situation. Suppose for contradiction that at termination some frontier vertex $v$ has strict ancestors entailed by $s \land \mathit{ctx}$, and let $m$ be a strongest such ancestor (one exists because the graph is finite and acyclic). By transitive closure, $(m, v)$ is an edge of $G$, so $m \in \mathrm{parents}(v)$. We claim $m \notin \mathit{visited}$: a vertex enters $\mathit{visited}$ only when, as a frontier vertex, it has a parent $p \notin \mathit{visited}$ with $s \land \mathit{ctx} \models (m \Rightarrow p)$; since $s \land \mathit{ctx} \models m$, Theorem~\ref{thm:reverse} would give $s \land \mathit{ctx} \models p$ with $p$ a strict ancestor of $m$---contradicting the choice of $m$. Moreover, the residual check for the reverse edge $(v, m)$ passes: $s \land \mathit{ctx} \models m \models (v \Rightarrow m)$. Hence $m \in \mathit{pred}$ when $v$ is processed in the final pass, so the frontier changes---contradicting stabilization.

Finally, the conjunction of the frontier vertices $v_1 \land \cdots \land v_k$ is entailed by $s \land \mathit{ctx}$, because $s \land \mathit{ctx} \models v_i$ for each $i$ implies $s \land \mathit{ctx} \models v_1 \land \cdots \land v_k$. And since $(v_1 \land \cdots \land v_k) \models v_i$ for each $i$, the conjunction is stronger than any individual frontier vertex.
\end{proof}

\begin{proof}[Proof of Lemma~\ref{lem:completeness}]
One inclusion is given by Theorems~\ref{thm:weakest-correct} and~\ref{thm:strongest-correct}: every returned vertex entails $s$ in context (resp.\ is entailed by $s$ in context) and none of its strict descendants (resp.\ ancestors) does the same; by transitive closure, the strict descendants and ancestors of a vertex are exactly the strictly weaker and strictly stronger vertices, so every returned vertex is weakest (resp.\ strongest) in the stated set.

For the converse, let $m$ be a weakest element of $\{w \mid w \land \mathit{ctx} \models s\}$; we show $m$ is in the final frontier of the downward search. First, $m$ never enters $\mathit{visited}$: that would require an unvisited out-neighbor $w$ with $(w \land \lnot m) \land \mathit{ctx} \models s$, and since $m \land \mathit{ctx} \models s$, Theorem~\ref{thm:main-ctx} would give $w \land \mathit{ctx} \models s$ with $w$ strictly weaker than $m$, contradicting the choice of $m$. Second, $m$ enters the frontier: if $m = \bot$, it is in the initial frontier $\{\bot\}$ directly; otherwise $\bot < m$, and the universal root $\bot$ is in the initial frontier with $(\bot, m)$ an edge by transitive closure whose check passes, $(m \land \lnot \bot) \land \mathit{ctx} \equiv m \land \mathit{ctx} \models s$. Hence whenever $\bot$ is processed, $m \in \mathit{succ}$, so $\mathit{succ} \neq \emptyset$; the loop therefore cannot stabilize with $\bot$ in the frontier, and when $\bot$ is moved to $\mathit{visited}$, $m$ enters the frontier. Finally, $m$ never leaves the frontier, since leaving requires entering $\mathit{visited}$. The argument for the upward search is dual, with the universal sink $\top$ in place of $\bot$ and Theorem~\ref{thm:reverse} in place of Theorem~\ref{thm:main-ctx}.
\end{proof}

\section{Bracketing and Defining Ambiguity: Proofs}\label{sec:ambiguity-app}

This appendix proves the results of Section~\ref{sec:ambiguity}.

\begin{proof}[Proof of Theorem~\ref{thm:non-crossing}]
Let $v \in F_\downarrow$ and $u \in F_\uparrow$. By the safety of the two searches, $v \land \mathit{ctx} \models s$ and $s \land \mathit{ctx} \models u$. Since also $v \land \mathit{ctx} \models \mathit{ctx}$, we have $v \land \mathit{ctx} \models s \land \mathit{ctx} \models u$, which establishes the first claim.

For the second claim, suppose $u \models v$. Then $u \land \mathit{ctx} \models v \land \mathit{ctx}$; conversely, the first claim gives $v \land \mathit{ctx} \models u$, hence $v \land \mathit{ctx} \models u \land \mathit{ctx}$. So $u \land \mathit{ctx} \iff v \land \mathit{ctx}$. For equivalence with $s \land \mathit{ctx}$: from $s \land \mathit{ctx} \models u$ we get $s \land \mathit{ctx} \models u \land \mathit{ctx}$, and from $u \land \mathit{ctx} \iff v \land \mathit{ctx} \models s$ we get $u \land \mathit{ctx} \models s \land \mathit{ctx}$. Hence all three formulas are equivalent.
\end{proof}

\begin{proof}[Proof of Proposition~\ref{prop:trichotomy}]
Cases (i) and (iii) are disjoint because the residue contains $\lnot\varphi_\downarrow$; cases (ii) and (iii) because it contains $\varphi_\uparrow$. Cases (i) and (ii) are disjoint: a model of $\varphi_\downarrow \land \mathit{ctx}$ satisfies $s$, hence $s \land \mathit{ctx}$, hence $\varphi_\uparrow$. For coverage, a model of $\mathit{ctx}$ satisfying neither (i) nor (ii) satisfies $\varphi_\uparrow \land \lnot\varphi_\downarrow \land \mathit{ctx}$, which is the residue. The certificate in (i) is the soundness claim $\varphi_\downarrow \land \mathit{ctx} \models s$ of Theorem~\ref{thm:weakest-correct}; the certificate in (ii) is the contrapositive of $s \land \mathit{ctx} \models \varphi_\uparrow$ from Theorem~\ref{thm:strongest-correct}: a model of $\mathit{ctx}$ falsifying $\varphi_\uparrow$ must falsify $s$.
\end{proof}

\begin{proof}[Proof of Theorem~\ref{thm:disambiguation}]
(1)$\iff$(2) is immediate from the definition, and (3)$\Rightarrow$(2) is trivial. For (2)$\Rightarrow$(3): the entailments $\varphi_\downarrow \land \mathit{ctx} \models s \land \mathit{ctx}$ and $s \land \mathit{ctx} \models \varphi_\uparrow \land \mathit{ctx}$ hold unconditionally (Theorems~\ref{thm:weakest-correct} and~\ref{thm:strongest-correct}), and assumption (2) closes the cycle with $\varphi_\uparrow \land \mathit{ctx} \models \varphi_\downarrow \land \mathit{ctx}$, so all three formulas are equivalent.
\end{proof}

\begin{proof}[Proof of Corollary~\ref{cor:inversion}]
By Theorem~\ref{thm:non-crossing}, $u \land \mathit{ctx} \iff v \land \mathit{ctx}$. Since $\varphi_\uparrow \models u$ and $\lnot\varphi_\downarrow \models \lnot v$, the residue entails $u \land \lnot v \land \mathit{ctx}$; and $u \land \mathit{ctx} \iff v \land \mathit{ctx}$ makes this equivalent to $v \land \lnot v \land \mathit{ctx}$, which is unsatisfiable.
\end{proof}

\begin{proof}[Proof of Proposition~\ref{prop:monotone}]
Write $\varphi_\downarrow, \varphi_\uparrow$ for the aggregates on $(G, \mathit{ctx})$ and $\varphi'_\downarrow, \varphi'_\uparrow$ for those on $(G', \mathit{ctx}')$. We claim $\varphi_\downarrow \models \varphi'_\downarrow$ and $\varphi'_\uparrow \models \varphi_\uparrow$. For the first, let $v \in F_\downarrow$. Then $v \land \mathit{ctx}' \models v \land \mathit{ctx} \models s$, so $v$ belongs to $\{w \in P' \mid w \land \mathit{ctx}' \models s\}$; choose a weakest element $m$ of this set with $v \models m$ (one exists since $P'$ is finite). By Lemma~\ref{lem:completeness}, $m \in F'_\downarrow$, hence $v \models m \models \varphi'_\downarrow$; taking the disjunction over $F_\downarrow$ gives $\varphi_\downarrow \models \varphi'_\downarrow$. Dually, each $u \in F_\uparrow$ satisfies $s \land \mathit{ctx}' \models u$ and lies below some strongest element $m'$ of $\{w \in P' \mid s \land \mathit{ctx}' \models w\}$; by Lemma~\ref{lem:completeness}, $m' \in F'_\uparrow$, so $\varphi'_\uparrow \models m' \models u$ for every $u \in F_\uparrow$, giving $\varphi'_\uparrow \models \varphi_\uparrow$. Now any model of $\mathrm{Amb}(s; G', \mathit{ctx}')$ satisfies $\varphi'_\uparrow$ and hence $\varphi_\uparrow$; satisfies $\lnot\varphi'_\downarrow$ and hence $\lnot\varphi_\downarrow$ (contrapositive of the first claim); and satisfies $\mathit{ctx}'$ and hence $\mathit{ctx}$. It therefore satisfies $\mathrm{Amb}(s; G, \mathit{ctx})$.
\end{proof}

\section{Realizability: Proofs}\label{sec:realizability-app}

This appendix proves the results of Section~\ref{sec:realizability}.

\begin{proof}[Proof of Theorem~\ref{thm:realizability}]
Downward, by induction on entry into the frontier, following the soundness
argument of Theorem~\ref{thm:weakest-correct} with the oracle's accepted
checks in place of the criterion. Initially the frontier is $\{\bot\}$ and
$\bot \land \mathit{ctx} = \bot \models s$, with no query issued, since completion
makes both initializations unconditional. This relies on completion: an
implementation that instead seeds the frontier with its sufficient roots issues a
\emph{vertex}-level check $r \land \mathit{ctx} \models s$, which an oracle
answering only edge checks cannot be asked, so its initial state is not
oracle-realizable and its base case is not oracle-free. Inductively, a vertex $v$ enters the
frontier only across an edge $(u,v)$ with $u$ a frontier vertex and $O$ accepting
that edge's co-residual check; since $O$ is sound,
$(v \land \lnot u) \land \mathit{ctx} \models s$ in fact holds, and since
$u \land \mathit{ctx} \models s$ by the inductive hypothesis,
Theorem~\ref{thm:main-ctx} yields $v \land \mathit{ctx} \models s$.

Upward is dual. Initially the frontier is $\{\top\}$ and
$s \land \mathit{ctx} \models \top$. Inductively, $u$ enters only across a reverse
edge $(v,u)$ with $O$ accepting that edge's residual check, so
$s \land \mathit{ctx} \models (v \Rightarrow u)$ holds by soundness; with
$s \land \mathit{ctx} \models v$ from the inductive hypothesis,
Theorem~\ref{thm:reverse} yields $s \land \mathit{ctx} \models u$. The claims
for the aggregates follow, since a disjunction of formulas each entailing $s$ in
context entails it, and dually for the conjunction.

Neither induction consults a \emph{rejected} check: rejections only withhold
traversals. Completeness is therefore not needed.
\end{proof}

\begin{proof}[Proof of Theorem~\ref{cor:accuracy}]
Soundness and completeness together make $O$'s verdict coincide with the solver's
on every check the run issues, so the oracle run and the oracle-free run license
exactly the same traversals. We do \emph{not} argue that the two runs proceed in
lockstep: the step selects an \emph{arbitrary} frontier vertex with a licensed
successor, so the procedure is nondeterministic and the next query is not a
function of the state alone. Instead we appeal to the oracle-free analysis twice.
Because the licensed traversals agree, the oracle run's reachable states satisfy
the same five invariants as the solver run's
(Theorems~\ref{thm:weakest-correct} and~\ref{thm:strongest-correct}), so
by the stabilization argument there its terminal frontier equals the declarative
frontier---the weakest sufficient, respectively strongest necessary, vertices
of Lemma~\ref{lem:completeness}. The solver run's terminal frontier equals that
same declarative frontier. Hence the two coincide, and tightness is
Lemma~\ref{lem:completeness}.
\end{proof}

\begin{proof}[Proof of Theorem~\ref{thm:disjunctive-recovery}]
Downward. Safety (Theorem~\ref{thm:weakest-correct}) gives $\varphi_\downarrow
\land \mathit{ctx} \models s^\star$. Conversely, each disjunct satisfies
$c_i \models s^\star$, hence $c_i \land \mathit{ctx} \models s^\star$, so
$c_i$ belongs to $\{w \in V(G) \mid w \land \mathit{ctx} \models
s^\star\}$. By Frontier Completeness (Lemma~\ref{lem:completeness}) $F_\downarrow$
is the set of weakest elements of this set, and every element of the set entails
one of them; thus $c_i \models m \models \varphi_\downarrow$ for
some $m \in F_\downarrow$. Taking the disjunction, $s^\star = \bigvee_i c_i \models
\varphi_\downarrow$, so $s^\star \land \mathit{ctx} \models
\varphi_\downarrow \land \mathit{ctx}$; combined with soundness the two
are equivalent. The conjunctive case is dual: each conjunct satisfies $s^\star
\models c_i$, placing $c_i$ in $\{w \mid s^\star \land \mathit{ctx}
\models w\}$, whose strongest elements are $F_\uparrow$; hence
$\varphi_\uparrow \models c_i$ for every $i$, so $\varphi_\uparrow \models s^\star$,
and with soundness $\varphi_\uparrow \land \mathit{ctx} \iff s^\star
\land \mathit{ctx}$.
\end{proof}

\section{Algebraic Foundations: Residuation and the Boolean/Intuitionistic Boundary}\label{sec:heyting}

The algebraic structure underlying \sysnames{} is that of a \emph{Heyting algebra}, introduced by Heyting~\cite{heyting-original} in his formalization of intuitionistic logic. A Heyting algebra is a bounded lattice $(H, \leq, \land, \lor, \bot, \top)$ equipped with a binary operation $\to$ (the \emph{relative pseudo-complement}) satisfying:
\[
c \leq (a \to b) \quad\text{if and only if}\quad c \land a \leq b
\]

The set of classical logic formulas modulo equivalence, ordered by entailment, forms a Boolean algebra---the Lindenbaum--Tarski algebra~\cite{tarski-original,lindenbaum-tarski}---and every Boolean algebra is in particular a Heyting algebra:
\begin{itemize}
  \item The partial order $\leq$ is entailment ($\varphi \leq \psi$ iff $\varphi \models \psi$).
  \item Meet and join are conjunction and disjunction; $\bot$ and $\top$ are the contradictory and valid formulas.
  \item The Heyting implication $\to$ is realized by material implication ($\lnot\varphi \lor \psi$).
  \item The Heyting property holds: $c \models (\lnot a \lor b)$ iff $c \land a \models b$.
\end{itemize}

The residual (Definition~\ref{def:residual}) is a standard construction in Heyting algebras: the residual $v \Rightarrow u$ is precisely the relative pseudo-complement of $v$ relative to $u$, and the fact that it is the \emph{weakest} formula strengthening $v$ back to $u$ is exactly the defining adjunction. Consequently, the residual decomposition (Lemma~\ref{lem:residual}) and the upward search built on it (Theorem~\ref{thm:reverse}) are valid in \emph{any} Heyting algebra---they are intuitionistically sound.

The co-residual \emph{as we defined it}, by contrast, is specifically Boolean. The case split underlying Theorem~\ref{thm:main-ctx}, namely that every model of $v$ satisfies either $u$ or $v \land \lnot u$ (i.e., $v \iff u \lor (v \land \lnot u)$ for $u \models v$), relies on Boolean complementation, an instance of the law of excluded middle, and fails in non-Boolean Heyting algebras. The ambiguity residue of Section~\ref{sec:ambiguity}, which negates the sufficient-frontier disjunction, is likewise a Boolean construction.

\begin{remark}[A minimal counterexample]\label{rem:three-chain}
The failure is witnessed already by the smallest non-Boolean Heyting algebra: the three-element chain $\bot < a < \top$, with $x \to y = \top$ if $x \leq y$ and $x \to y = y$ otherwise (equivalently, the algebra of upward-closed sets of a two-world Kripke frame $w_0 \leq w_1$, where $a$ is the proposition true only at $w_1$). Take $u = a$ and $v = \top$ (a valid edge, since $u \leq v$ strictly), with $\mathit{ctx} = \top$ and $s = a$. The hypothesis of Theorem~\ref{thm:main-ctx} holds: $u \land \mathit{ctx} = a \leq s$. The co-residual is $\mathit{cores} = v \land \lnot u = \top \land (a \to \bot) = \bot$, so $\mathit{cores} \land \mathit{ctx} = \bot \leq s$ and the edge check passes vacuously---yet $v \land \mathit{ctx} = \top \not\leq a = s$. Traversing the edge would be unsound. The identity that breaks is exactly the case split: $u \lor (v \land \lnot u) = a \lor \bot = a < \top = v$, so $u$ and the co-residual together fail to exhaust $v$. By contrast, the residual decomposition is unharmed on the same data---with $\rho = v \to u = a$ we have $v \land \rho = \top \land a = a = u$---consistent with the upward search remaining sound intuitionistically.
\end{remark}

\begin{remark}[The co-residual beyond the Boolean case]\label{rem:coheyting}
The Boolean restriction above is an artifact of \emph{how we realized} the co-residual, not of the notion itself. We defined the co-residual concretely as $v \land \lnot u$ for simplicity, which is the right formula in our classical setting. But the abstract co-residual of Definition~\ref{def:coresidual}---the strongest $\delta$ with $u \lor \delta \iff v$---is the \emph{co-Heyting subtraction} $v \setminus u$, the exact order-dual of the residual: where the residual is the right adjoint of meet ($v \land (-) \dashv v \Rightarrow (-)$), the subtraction is the left adjoint of join. It exists in any \emph{co-Heyting} (Brouwerian) algebra, and there the decomposition $v \iff u \lor (v \setminus u)$ holds by the defining adjunction, with no appeal to excluded middle; Theorem~\ref{thm:main-ctx} then follows constructively by distributing $\mathit{ctx}$ over the join. What excluded middle actually buys us is only the \emph{identification} $v \setminus u = v \land \lnot u$: in the three-chain above the true subtraction is $\top \setminus a = \top$ (not $\bot$), which restores the theorem. Thus the residual/co-residual duality is a Heyting/co-Heyting duality, and the full bidirectional bracket lives in any \emph{bi-Heyting} algebra; classical logic is merely where the co-residual collapses to the familiar $v \land \lnot u$.
\end{remark}

This delineation has practical significance: if candidate formalizations live in a constructive logic (e.g., statements in a proof assistant based on intuitionistic type theory), the upward search remains sound as is, while the downward search's reduction to co-residuals requires the ambient logic to be classical.

In our setting---closed formulas in classical first-order logic checked by an SMT solver---both constructions apply, and the algebraic foundation guarantees that the quantities our algorithms compute are canonical: they are the unique, tightest characterizations of the semantic gaps between candidates.

\section{Machine-Checked Proofs}\label{sec:lean}

We have mechanized the results of this appendix in the Lean~4 proof
assistant~\cite{lean4} on top of Mathlib~\cite{mathlib}. The proof development
exploits the Lindenbaum--Tarski view of Section~\ref{sec:heyting} directly:
rather than model first-order syntax, it works \emph{abstractly} over an
arbitrary Boolean algebra (and, where the intuitionistic fragment is isolated, a
Heyting algebra), taking entailment as the lattice order $\leq$, conjunction and
disjunction as meet and join, the residual $v \Rightarrow u$ as the Heyting
implication, and the co-residual $v \land \lnot u$ as the Boolean difference.
Because vertices are algebra elements and edges are the strict order, the
transitive closure of the \sysname{} holds by construction and never
appears as a hypothesis.

One translation is worth spelling out, since it reverses a word. The paper speaks of
the \emph{weakest} candidates sufficient for $s$ and the \emph{strongest} candidates
$s$ entails. Taking entailment as $\leq$ makes a stronger formula the smaller one, so
those sets are respectively the $\leq$-\emph{maximal} and $\leq$-\emph{minimal}
elements, and that is how the development names them: \texttt{suffFrontier} filters
the sufficient set by \texttt{Maximal} and \texttt{necFrontier} filters the necessary
set by \texttt{Minimal}. Reading \emph{weakest} for \texttt{Maximal} and
\emph{strongest} for \texttt{Minimal} is the only renaming needed to match the two
statements of Lemma~\ref{lem:completeness}.

The mechanization covers the residual and co-residual decompositions
(Lemmas~\ref{lem:residual} and~\ref{lem:coresidual}), both edge-traversal
theorems (Theorems~\ref{thm:reverse} and~\ref{thm:main-ctx}), the
soundness and bracketing guarantees of the two searches
(Theorems~\ref{thm:weakest-correct} and~\ref{thm:strongest-correct}),
non-crossing (Theorem~\ref{thm:non-crossing}), frontier completeness
(Lemma~\ref{lem:completeness}), the trichotomy, disambiguation criterion, and
inversion corollary of Section~\ref{sec:ambiguity-app}, monotonicity of the
residue (Proposition~\ref{prop:monotone}), exact recovery under coverage
(Theorem~\ref{thm:disjunctive-recovery}), and safety from a sound oracle with
tightness from a complete one
(Theorems~\ref{thm:realizability} and~\ref{cor:accuracy}).
Lemma~\ref{lem:absolute} is the one result with no counterpart, because the
development has no need of it: every check is phrased through a single
context-relativized entailment, \texttt{EntailsCtx\ v\ ctx\ s} for
$v \sqcap \mathit{ctx} \leq s$, and both adjoints appear in it absolutely. The
necessity criterion does not mention the context at all---it is stated for an
arbitrary left-hand side, which the paper instantiates at $s \land \mathit{ctx}$---so
the relativized adjoints the lemma shows to be unnecessary are absent by construction.
The frontiers are formalized in two ways that we prove agree. First, we formalize them
\emph{declaratively}, as the extremal elements of the two context-relativized
entailment sets, with frontier completeness certifying that this characterization
coincides with the aggregate the search computes, independently of traversal
order. Second, we formalize them \emph{algorithmically}: both searches of
Figures~\ref{fig:alg-weakest} and~\ref{fig:alg-strongest} are modeled as
nondeterministic transition systems over $(\mathit{frontier}, \mathit{visited})$
states, and we prove that a termination measure strictly decreases on every step and
that every terminal state reachable from the initial state returns exactly the
declarative frontier---so the imperative loop meets its specification for any
processing order. The completion with $\bot$ and $\top$ is mechanized as well: we
prove that when the pool contains $\bot$ and $\top$ the two searches start from
nonempty frontiers and, at any terminal state, return nonempty frontiers
(\texttt{downInit\_nonempty}/\texttt{upInit\_nonempty} and
\texttt{frontierDown\_nonempty}/\texttt{frontierUp\_nonempty}), so the
empty-frontier edge case is discharged rather than assumed away; because the
searches are stated over an arbitrary finite vertex set, all guarantees above
apply verbatim to the completed pool.

Two aspects of the algebraic story surface cleanly in the mechanization. First,
the residual decomposition and the necessity search depend only on
propositional extensionality---no choice or Boolean complementation---confirming
their intuitionistic validity, whereas the sufficiency side is genuinely
classical. Second, the minimal counterexample of Remark~\ref{rem:three-chain} is
mechanized on the concrete three-element Heyting chain, where the co-residual
edge check passes vacuously yet the traversal it would license is unsafe. Every
theorem is proved without axioms beyond Lean's standard three
(\texttt{propext}, \texttt{Classical.choice}, \texttt{Quot.sound}) and with no
incomplete proofs (i.e., no \texttt{sorry}). 

\section{Full Worked Example}\label{sec:motivating-app}

This appendix reproduces the payment-service example in full, retaining the
conditional candidate $\candCo$ and the internal-microservice context that the
shortened body version (Section~\ref{sec:motivating}) omits. It exhibits the
full range of phenomena---incomparable roots, a chain of single-mechanism
weakenings, and a conditional sink---in a single example, and in particular
develops the chain-shaped conditionality ambiguity of Context~2.

We take a safety statement given in natural language and autoformalize it using the \sysnames{} based approach proposed in the paper. Consider a security policy statement $s$:
\begin{quote}
``All requests to the payment service must be authenticated.''
\end{quote}

When formalizing this statement, an LLM may interpret it in multiple ways. Before presenting the candidates, we briefly explain the authentication mechanisms that appear in them:
\begin{itemize}
  \item \textbf{Mutual Transport Layer Security (mTLS):} A cryptographic protocol in which both the client and the server present certificates to each other, providing strong identity verification at the transport layer.
  \item \textbf{JSON Web Token (JWT):} A signed token that carries claims (e.g., ``this client is authorized for payment operations''). It provides application-level proof of authorization beyond transport-level identity.
  \item \textbf{IP allowlist:} A network-level control that restricts access to requests originating from pre-approved IP addresses.
\end{itemize}

For the example, suppose that an LLM produces the following pool of candidates,
when asked to formalize the statement.  (The pool of candidates are representative of the result of our end-to-end pipeline;  Appendix~\ref{app:generated} reports an end-to-end example programmatically generated by the pipeline).
\begin{description}[align=left, leftmargin=6.5em, labelsep=0.5em, itemsep=2pt]
  \item[$\candTJ$:] All requests must use mutual TLS \textbf{and} include a signed JWT bearing a payment-scope claim. (Transport-level identity plus application-level authorization.)
  \item[$\candTI$:] All requests must use mutual TLS \textbf{and} originate from an allow-listed IP address. (Transport-level identity plus network-level restriction.)
  \item[$\candT$:] All requests must use mutual TLS. (Transport-level identity only.)
  \item[$\candCr$:] All requests must include valid credentials (API key, token, or certificate). (Any form of authentication.)
  \item[$\candCo$:] If a request reaches the payment service, it must carry valid credentials. (Conditional: only applies when requests actually arrive.)
\end{description}

As in Section~\ref{sec:motivating}, each candidate is a closed
first-order sentence and the names below abbreviate its parts. Over a sort of
requests, with $P(r)$ for ``$r$ is a request to the payment service'',
$\mathit{arrives}(r)$ for ``$r$ actually reaches the service'', and
$\mathit{mtls}$, $\mathit{jwt}$, $\mathit{ipok}$, $\mathit{cred}$ the mechanism
predicates:
\[
\begin{array}{r@{\;\equiv\;}l}
\mathsf{TLS}  & \forall r.\; P(r) \Rightarrow \mathit{mtls}(r) \\
\mathsf{CRED} & \forall r.\; P(r) \Rightarrow \mathit{cred}(r) \\
\mathsf{COND} & \forall r.\; \mathit{arrives}(r) \Rightarrow \mathit{cred}(r)
\end{array}
\]
and $\mathsf{JWT}$, $\mathsf{IP}$ follow $\mathsf{TLS}$'s shape, giving
$\candTJ = \mathsf{TLS} \land \mathsf{JWT}$,
$\candTI = \mathsf{TLS} \land \mathsf{IP}$, $\candT = \mathsf{TLS}$,
$\candCr = \mathsf{CRED}$, and $\candCo = \mathsf{COND}$. Note that $\candCr$ and
$\candCo$ differ only in the \emph{antecedent} of the quantified implication---a
distinction that exists only in the first-order reading, and the one Context~2
below turns out to hinge on. Appendix~\ref{app:generated} shows that our pipeline
emits exactly these two antecedents unprompted.

\input{fig-ig-payment-app}

These candidates vary in their logical strength:
\[
\resizebox{\fitcol}{!}{$\candTJ \models \candT \models \candCr \models \candCo \quad \text{and} \quad \candTI \models \candT$}
\]
We represent this ordering as an \sysname{} (Figure~\ref{fig:ig-example-app}), with vertices representing candidates and edges representing strict entailment (stronger $\to$ weaker). The graph has two roots $\candTJ$ and $\candTI$ (incomparable: $\candTJ$ requires JWT but not IP allowlisting, $\candTI$ requires IP allowlisting but not JWT) and one sink $\candCo$.
In Figure~\ref{fig:ig-example-app}, we include, for simplicity of exposition, only the \emph{covering} edges---the Hasse diagram of the entailment order. The \sysname{} itself contains an edge for every strictly comparable pair (e.g., $\candTJ \to \candCr$, $\candTJ \to \candCo$, $\candTI \to \candCo$, $\candT \to \candCo$ are also edges) and this is required for correctness of our approach.
The walkthroughs below traverse only covering edges, which suffices for exposition (the omitted edges do not change the results).

The two searches over this graph are driven by the two parties a
formalization must serve. A \emph{provider}, such as a payment service
provider, must \emph{enforce} the statement and asks: what is the
\emph{weakest} configuration I can build that still guarantees the
statement? A \emph{customer}, who uses the service, must \emph{rely on} the
statement and asks the dual question: what is the \emph{strongest}
property I can safely assume holds? Each view runs one search:
\begin{itemize}
  \item The provider's view drives a \textbf{downward search}, which finds the weakest candidates that still entail $s$ in context---the \emph{sufficient frontier} $F_\downarrow$ (each candidate is a sufficient condition for $s$, so meeting it certifies satisfaction, and taking the weakest such candidate minimizes implementation cost without over-promising).
  \item The customer's view drives an \textbf{upward search}, which finds the strongest candidates that $s$ entails in context---the \emph{necessary frontier} $F_\uparrow$ (each candidate is a necessary condition for $s$, so failing it certifies violation, and the customer relies on no more than the statement guarantees).
\end{itemize}
The ambiguity bracket between these two frontiers collects the candidates the context excludes from neither side: everything strictly stronger than $F_\downarrow$ over-commits (the provider would build more than the statement requires) and everything strictly weaker than $F_\uparrow$ under-commits (the customer would assume more than the statement guarantees), so the bracket is exactly what remains once both one-sided verdicts are applied. We now show how this bracket behaves under three different contexts.

Three conventions hold for all three contexts below. The statement stays in natural
language: formalizing it is the problem, so exhibiting it as a formula would name the
answer in advance. Each \emph{Yes} or \emph{No} below is therefore the answer the
entailment oracle gives to one edge check, not a derivation
(Section~\ref{sec:realizability}); the one exception is noted where it arises, a
satisfiability question the solver settles. And a context may constrain the situations,
in which case it has a formal part a solver can use, or it may constrain only what the
statement demands, in which case it has none. Context~1 does both; Contexts~2 and~3 do
only the latter.

\subsection{Context 1: PCI-DSS Compliant Gateway}

The context supplies three facts:
\begin{enumerate}
  \item the service processes credit card transactions, so payment traffic occurs:
    $\exists r.\; P(r)$;
  \item it runs on an open network, where no allowlist covers all legitimate traffic:
    $\exists r.\; P(r) \land \lnot \mathit{ipok}(r)$;
  \item it must comply with the Payment Card Industry Data Security Standard
    (PCI-DSS), which requires proof of payment-specific authorization.
\end{enumerate}
Facts~(1) and~(2) are the formal part; fact~(3) constrains what ``authenticated''
demands and stays in natural language. The statement remains
``all requests must be authenticated.''

\paragraph{Downward search (sufficient frontier: weakest that entail $s$).}

\begin{itemize}
  \item Root $\candTJ$: mTLS + payment-scope JWT satisfies PCI-DSS authentication. $\candTJ \land \mathit{ctx} \models s$. Yes.
  \item Root $\candTI$: mTLS + IP allowlisting. Fact~(2) says some payment request falls outside any allowlist, so $\candTI \land \mathit{ctx}$ is unsatisfiable and $\candTI$ cannot be implemented here. It entails $s$ only vacuously, so the search skips it (Remark~\ref{rem:vacuous}).
  \item Edge $(\candTJ, \candT)$: Does dropping the JWT violate the policy under PCI-DSS? Yes---PCI-DSS requires proof of payment-specific authorization. \textbf{Cannot traverse.}
\end{itemize}

\textbf{Sufficient frontier: $F_\downarrow = \{\candTJ\}$.}

\paragraph{Upward search (necessary frontier: strongest entailed by $s$).}

\begin{itemize}
  \item Sink $\candCo$: $s \land \mathit{ctx} \models \candCo$? Yes---the statement mandates authentication, which entails at least conditional credentials.
  \item Edge $\candCo \leftarrow \candCr$: PCI-DSS requires controls to be in place as a standing obligation, regardless of whether requests happen to arrive. Climb to $\candCr$.
  \item Edge $\candCr \leftarrow \candT$: PCI-DSS mandates strong cryptographic identity at the transport layer, implying mTLS. Climb to $\candT$.
  \item Edge $\candT \leftarrow \candTJ$: PCI-DSS requires proof of payment-specific authorization, implying a payment-scope JWT. Climb to $\candTJ$.
  \item Edge $\candT \leftarrow \candTI$: Context says IP restrictions are infeasible---cannot assume IP allowlisting. Do not climb to $\candTI$.
\end{itemize}

\textbf{Necessary frontier: $F_\uparrow = \{\candTJ\}$.}

\paragraph{The bracket.} Both frontiers coincide at $\candTJ$. The gap is zero---the PCI-DSS context fully disambiguates the statement. The residue $\candTJ \land \lnot \candTJ \land \mathit{ctx}$ is unsatisfiable, and the ambiguity bracket collapses to a single point. This is the simplest outcome the bracket can produce: the two searches meet, and the statement has a single contextual meaning.

\subsection{Context 2: Internal Microservice}

The context supplies one fact: the service runs in a trusted internal network with no
compliance requirements. It rules out no configuration, so it has no formal part; it
bears only on what the statement demands. The statement remains ``all requests must be
authenticated.''

\paragraph{Downward search (sufficient frontier).}

\begin{itemize}
  \item Both roots $\candTJ$ and $\candTI$ entail $s$ in context (mTLS + JWT and mTLS + IP allowlist both certainly constitute ``authenticated'').
  \item Edge $(\candTJ, \candT)$: Does a client with a valid certificate but no JWT satisfy ``must be authenticated'' in a trusted network? Yes---JWT is unnecessary internally. Traverse to $\candT$.
  \item Edge $(\candTI, \candT)$: Does a client with a valid certificate from any IP satisfy the policy internally? Yes---IP restrictions are unnecessary. Traverse to $\candT$.
  \item Edge $(\candT, \candCr)$: Does a client authenticated by an API key (without mTLS) satisfy the policy in a trusted network? Yes---any valid credential suffices internally. Traverse to $\candCr$.
  \item Edge $(\candCr, \candCo)$: The co-residual admits configurations that carry credentials only on requests that actually arrive---including one with no enforcement configured and, as it happens, no traffic. Does such vacuous satisfaction meet ``all requests must be authenticated''? Nothing in the context licenses reading the policy as conditional on traffic. \textbf{Cannot traverse.}
\end{itemize}

\textbf{Sufficient frontier: $F_\downarrow = \{\candCr\}$.}

\paragraph{Upward search (necessary frontier).}

\begin{itemize}
  \item Sink $\candCo$: Does $s \land \mathit{ctx} \models \candCo$? The statement says ``all requests must be authenticated,'' which certainly entails ``if a request arrives, it carries credentials.'' Yes.
  \item Edge $\candCo \leftarrow \candCr$ (residual: unconditional enforcement): Does the statement guarantee that the policy holds unconditionally, even in configurations where no requests arrive? ``All requests'' can be read as quantifying over the requests that actually occur (conditional) or as imposing a standing requirement on the service (unconditional), and nothing in a trusted internal network with no compliance requirements forces the stronger reading. No. \textbf{Stop.}
\end{itemize}

\textbf{Necessary frontier: $F_\uparrow = \{\candCo\}$.}

\paragraph{The bracket.} The ambiguity bracket is $[\candCr, \candCo]$: $\candCr$ certainly suffices to guarantee the statement, and the statement certainly guarantees at least $\candCo$. The gap between them is genuine residual ambiguity: the context neither licenses weakening $\candCr$ to $\candCo$ (the downward check fails) nor mandates strengthening $\candCo$ to $\candCr$ (the upward check fails). Both the downward and the upward check must fail for the gap to be genuine---had both succeeded, $\candCr$ and $\candCo$ would be contextually equivalent to the statement and to each other, and the bracket would collapse to a point (Theorem~\ref{thm:non-crossing}). Should the policy apply unconditionally ($\candCr$) or only to requests that actually arrive ($\candCo$)? The context does not resolve this, signaling that clarification may be needed on exactly this point. We capture this gap as a single formula, the \emph{ambiguity residue} (Section~\ref{sec:ambiguity}). To see what it says, recall the two readings from
the pool above and abbreviate them $\mathsf{std}$ and $\mathsf{auth}$: the
unconditional $\candCr = \mathsf{std} \equiv \forall r.\,P(r) \Rightarrow
\mathit{cred}(r)$ (``credential enforcement is a standing part of the
configuration, applied to every request whether or not any arrives'') and the
conditional $\candCo = \mathsf{auth} \equiv \forall r.\,\mathit{arrives}(r)
\Rightarrow \mathit{cred}(r)$ (``every request that actually arrives carries valid
credentials''). The residue is the necessary frontier minus the sufficient frontier, within the context:
\[
\begin{aligned}
\mathrm{Amb} \;&=\; \varphi_\uparrow \land \lnot\varphi_\downarrow \land \mathit{ctx}
\;=\; \candCo \land \lnot \candCr \land \mathit{ctx}\\
\;&=\; \mathsf{auth} \land \lnot\mathsf{std} \land \mathit{ctx}.
\end{aligned}
\]
The context---a trusted internal network with no compliance requirements---asserts nothing that forces enforcement to be standing, so it does not entail $\mathsf{std}$ and the $\lnot\mathsf{std}$ conjunct survives; the residue is therefore satisfiable and the statement remains ambiguous.

It is worth dwelling on \emph{which} configurations the residue picks out, because the ambiguity region is more delicate than it first appears. A natural mistake is to equate the ambiguous configurations with the weak reading $\candCo$ itself---after all, $\candCo$ is the frontier the context could not push past. But that is wrong, and the residue makes the correction precise. Since $\candCr \models \candCo$, the weak reading decomposes as
\[
\candCo \;=\; \candCr \;\lor\; (\candCo \land \lnot \candCr),
\]
and only the second disjunct is in dispute. The first disjunct, $\candCr$, is the standing-guarantee region; it lies at or below the sufficient frontier and is \emph{certified to satisfy} $s$, so it carries no ambiguity at all. Subtracting it is exactly what $\lnot\varphi_\downarrow$ does. The residue is the remainder, $\candCo \land \lnot \candCr$: configurations in which every request that actually arrived was authenticated, yet nothing in the configuration \emph{guarantees} this. These are \emph{authenticated in fact, but not by mandate}---compliant as observed, not compliant by construction.

The distinction is not academic, and its sharpest witness is the degenerate model already surfaced by the downward search: a service that configures no credential enforcement whatsoever but, as it happens, receives no traffic. Concretely, this is a freshly deployed service that has not yet been attached to the network---no requests have reached it, and none can until it is connected. Every arriving request is (vacuously) authenticated, so it satisfies $\mathsf{auth}$; yet it plainly fails $\mathsf{std}$, since the moment it is wired up, the first request would go unauthenticated.
This witness is where the first-order reading earns its keep: the model
satisfies $\forall r.\,\lnot\mathit{arrives}(r)$, which makes $\mathsf{auth}$ true
by an empty quantifier range while leaving $\mathsf{std}$ false. Treating
$\mathsf{std}$ and $\mathsf{auth}$ as unrelated propositional atoms would admit
the same verdict by fiat; only the quantified forms \emph{explain} it, and only
they record that the two differ solely in the antecedent. This configuration satisfies $\candCo$ but not $\candCr$, sits squarely in the residue, and is precisely the kind of case a reader who conflated ambiguity with $\candCo$ would overlook. It is also the case that separates the two readings operationally: under the conditional reading the unattached service is already compliant---a clean bill of health that evaporates the instant it goes live---whereas under the unconditional reading it is a violation waiting to happen, flagged before it can do any harm. This is the conditionality question, isolated: should the policy bind unconditionally, or only on the requests that actually arrive?

\subsection{Context 3: Defense-in-Depth Requirement}

The context supplies one fact: the service operates in a regulated environment
requiring defense-in-depth, so at least two independent authentication layers must be
present. As in Context~2 it rules out no configuration and so has no formal part. The
statement remains ``all requests must be authenticated.''

\paragraph{Downward search (sufficient frontier).}

\begin{itemize}
  \item Root $\candTJ$: mTLS + JWT---two independent layers. $\candTJ \land \mathit{ctx} \models s$. Yes.
  \item Root $\candTI$: mTLS + IP allowlist---two independent layers. $\candTI \land \mathit{ctx} \models s$. Yes.
  \item Edge $(\candTJ, \candT)$: Dropping JWT leaves only mTLS---one layer. Violates defense-in-depth. \textbf{Cannot traverse.}
  \item Edge $(\candTI, \candT)$: Dropping IP allowlist leaves only mTLS---one layer. Violates defense-in-depth. \textbf{Cannot traverse.}
\end{itemize}

\textbf{Sufficient frontier: $F_\downarrow = \{\candTJ, \candTI\}$.}

\paragraph{Upward search (necessary frontier).}

\begin{itemize}
  \item Sink $\candCo$: $s \land \mathit{ctx} \models \candCo$? Yes.
  \item Edge $\candCo \leftarrow \candCr$: A regulated environment requires its authentication layers as standing controls, regardless of traffic. Climb to $\candCr$.
  \item Edge $\candCr \leftarrow \candT$: Defense-in-depth requires multiple layers, which implies at least one strong mechanism like mTLS. Climb to $\candT$.
  \item Edge $\candT \leftarrow \candTJ$: Does the statement guarantee a JWT specifically? Defense-in-depth requires multiple layers but does not prescribe which ones. No.
  \item Edge $\candT \leftarrow \candTI$: Does the statement guarantee an IP allowlist specifically? Same reasoning. No. \textbf{Stop.}
\end{itemize}

\textbf{Necessary frontier: $F_\uparrow = \{\candT\}$.}

\paragraph{The bracket.} The ambiguity bracket is $[\{\candTJ, \candTI\}, \candT]$: either two-layer configuration, $\candTJ$ or $\candTI$, suffices to guarantee the statement, and the statement guarantees at least transport-level identity $\candT$. To read the residue, recall from the pool above that
$\candTJ = \mathsf{TLS} \land \mathsf{JWT}$, $\candTI = \mathsf{TLS} \land
\mathsf{IP}$, and $\candT = \mathsf{TLS}$, where each abbreviation is a quantified
sentence requiring the corresponding mechanism of every request. Here the sufficient frontier has \emph{two} incomparable elements, so its aggregate is a disjunction, $\varphi_\downarrow = \candTJ \lor \candTI = \mathsf{TLS} \land (\mathsf{JWT} \lor \mathsf{IP})$, while $\varphi_\uparrow = \candT = \mathsf{TLS}$. The residue is
\[
\begin{aligned}
\mathrm{Amb} \;&=\; \varphi_\uparrow \land \lnot\varphi_\downarrow \land \mathit{ctx}
\;=\; \mathsf{TLS} \land \lnot(\mathsf{JWT} \lor \mathsf{IP}) \land \mathit{ctx}\\
\;&=\; \mathsf{TLS} \land \lnot\mathsf{JWT} \land \lnot\mathsf{IP} \land \mathit{ctx}.
\end{aligned}
\]
The residue holds exactly of configurations that pair transport-level identity with a second independent layer that is \emph{neither} of the two the pool names.
Unfolded, the negated conjuncts become existentials---$\lnot\mathsf{JWT}$
is $\exists r.\,P(r) \land \lnot\mathit{jwt}(r)$ and $\lnot\mathsf{IP}$ is
$\exists r.\,P(r) \land \lnot\mathit{ipok}(r)$---so the residue is satisfied by a
configuration in which the request missing a JWT and the request outside the
allowlist are \emph{different} requests. Read propositionally the residue looks
like a single configuration lacking both layers; read first-order, as the solver
reads it, it is weaker than that. This is subtler than a choice between $\candTJ$ and $\candTI$: both named two-layer configurations are already \emph{certified to satisfy} $s$ (they lie in $\varphi_\downarrow$), so choosing between them is not in dispute. What the context leaves undetermined is whether defense-in-depth may be met by some second mechanism outside the enumerated options---the residue is satisfiable exactly when such a mechanism exists, and it names the clarification to ask: must the second layer be specifically JWT or IP allowlisting, or does any independent layer suffice?

Where the internal-microservice residue $\mathsf{auth} \land \lnot\mathsf{std}$ ran \emph{along a chain}---isolating a question of strength, whether enforcement is unconditional or merely conditional---this residue sits \emph{between incomparable candidates}, isolating a question of kind: which layer. The two shapes are the operational counterpart of the linguistic distinction between generality and ambiguity proper.

\subsection{Summary}

\begin{center}
\footnotesize
\setlength{\tabcolsep}{4pt}
\resizebox{\fitcol}{!}{%
\begin{tabular}{llll}
\hline
\textbf{Context} & $F_\downarrow$ & $F_\uparrow$ & \textbf{Gap} \\
\hline
PCI-DSS & $\candTJ$ & $\candTJ$ & None (disambiguated) \\
Internal & $\candCr$ & $\candCo$ & conditionality \\
Defense-in-depth & $\{\candTJ, \candTI\}$ & $\candT$ & which 2nd layer \\
\hline
\end{tabular}}
\end{center}

The same \sysname{}, applied to the same statement, yields different ambiguity brackets under different contexts. When the bracket collapses to a point (PCI-DSS), the context fully determines the meaning. When a gap remains, the co-residuals within it identify exactly what is ambiguous.
Even in the presence of ambiguity, the two frontiers let each party act
soundly: the provider selects from $F_\downarrow$ and the customer from
$F_\uparrow$, exactly as motivated above.
In the internal context, for instance, the provider implements unconditional credential checks ($\candCr$)---free to use any credential mechanism---while the customer assumes only that arriving requests carry valid credentials ($\candCo$).
What the provider guarantees is strictly stronger than what the customer relies on: the residual ambiguity is absorbed as slack between the two roles, and both selections are sound.

\section{A Machine-Generated Overview Example}\label{app:generated}

The full worked example of Section~\ref{sec:motivating-app} uses a hand-curated
candidate pool (the five candidates $\candTJ$, $\candTI$, $\candT$, $\candCr$, $\candCo$) chosen to exhibit the full range of
phenomena---incomparable roots, a chain of single-mechanism weakenings, and a
conditional sink---in a single tidy picture. It is natural to ask what an LLM
actually produces for the same statement. This appendix reports a pool generated
directly by the pipeline of Section~\ref{sec:impl} and works the bracket through
on it. The example is not a substitute for Section~\ref{sec:motivating-app}; it is a
companion showing that the framework recovers essentially the same
structure---two incomparable roots and the curated $\candTJ \to \candT \to \candCr \to \candCo$ chain as a
sub-order---from machine-generated input.

\paragraph{Generation.} We prompted Claude Opus~4.6\footnote{Bedrock model ID \texttt{global.anthropic.claude-opus-4-6-v1}.} on AWS Bedrock at
temperature~$1.0$ (the setup of Section~\ref{sec:eval}) to formalize the same
statement---``All requests to the payment service must be authenticated''---in
SMT-LIB, supplying the payment-service domain vocabulary (mutual TLS, a
payment-scope JWT, an IP allowlist, and a generic valid credential). A single
formalization prompt tends to sample readings along one axis only: fixing a
deployment yields incomparable readings but no weakening chain, while asking for
graded strengths yields a chain but---because the strongest rung, ``require
every mechanism,'' is the same across deployments and dominates all others---only
a single root. To obtain both incomparable roots \emph{and} chains we gave the
model, for each deployment, a strength ladder \emph{local to that deployment's
own mechanisms}: for a trusted internal network the mechanisms are an IP
allowlist ($I$) and a generic credential ($C$); for an open PCI-DSS network they
are mutual TLS ($M$) and a payment-scope JWT ($J$). Within each we asked for the
strongest reading using only those mechanisms, then progressively weaker ones.
Because the two mechanism sets are disjoint, the strongest readings stay
incomparable. Clustering the resulting formulas by solver-checked equivalence,
and
abbreviating each mechanism predicate as in Section~\ref{sec:motivating}---a
universally quantified implication over requests to the payment service, with
$\text{arrive}\Rightarrow\!\cdot$ marking the variant whose antecedent is instead
that the request actually reaches the service---the readings are
\[
\setlength{\arraycolsep}{3pt}
\begin{array}{lll}
  V_1 = I \land C, & V_2 = M \land J, & V_3 = M,\\[2pt]
  V_4 = C, & V_5 = (\text{arrive}\Rightarrow M), & V_6 = (\text{arrive}\Rightarrow C).
\end{array}
\]
Every candidate the model returned was a quantified sentence over an
uninterpreted sort of requests---$34$ of $34$ across the three prompt variants.
The AWS pools of Section~\ref{sec:eval} are heavier still: in the properties we
ship as examples, every candidate quantifies over integer-indexed resource lists,
with bodies in the string and datatype theories. The abbreviations above are therefore names for
the parts of formulas like $V_2$, emitted verbatim as:
\begin{lstlisting}[basicstyle=\ttfamily\scriptsize]
(declare-sort Request 0)
(declare-fun isPaymentServiceRequest (Request) Bool)
(declare-fun hasMutualTLS (Request) Bool)
(declare-fun hasSignedPaymentScopeJWT (Request) Bool)
(assert
  (forall ((r Request))
    (=> (isPaymentServiceRequest r)
        (and (hasMutualTLS r)
             (hasSignedPaymentScopeJWT r)))))
\end{lstlisting}
Since $\forall$ distributes over $\land$, this is equivalent to
$M \land J$ with $M$ and $J$ as named above, so no strength is lost in the
display. The conditional readings $V_5$ and $V_6$ differ from $V_3$ and $V_4$
only in replacing the antecedent \texttt{isPaymentServiceRequest} with
\texttt{reachesService}---the distinction Appendix~\ref{sec:motivating-app}'s
Context~2 turns on, produced here without being asked for.

Four coincide with curated candidates: $V_2 = M\land J$ is $\candTJ$, $V_3 = M$ is $\candT$,
$V_4 = C$ is $\candCr$, and $V_6$ is the conditional sink $\candCo$. The internal ladder
supplies the second root $V_1 = I\land C$. Distinct cells collapsed as expected:
the ``credential-only'' rungs of both ladders gave $V_4 = C$ regardless of
deployment, since ``any credential'' does not depend on the network. The
remaining reading $V_5 = (\text{arrive}\Rightarrow M)$ is the traffic-conditional
analogue of $V_3 = M$ (since $M \models V_5 \models V_6$), lying on a parallel branch $V_3 \to V_5 \to V_6$; it is not on the curated chain and none
of the frontiers below select it, so we omit it from the walkthroughs.

\paragraph{The \sysname{}.} We order the readings by entailment, using
the domain's stated reading that an mTLS certificate and a payment-scope JWT are
each instances of a valid credential---so $M \models C$ and $J \models C$. This
gives the DAG of Figure~\ref{fig:ig-generated}: two incomparable roots $V_1 =
I\land C$ and $V_2 = M\land J$ (disjoint mechanism sets, so neither entails the
other) over a shared conditional sink $V_6$. The right branch $V_2 \to V_3 \to
V_4 \to V_6$ reproduces the curated chain $\candTJ \to \candT \to \candCr \to \candCo$ exactly, while
$V_1$ is a genuine second root feeding into the common weakening $V_4 = C$. The
per-deployment ladders thus recover both structural features a single-axis pool
misses: the chain (from the strength axis) and the incomparable second root
(from the deployment axis).

\input{fig-ig-generated}

\paragraph{Bracketing under each context.} We run both searches
(Section~\ref{sec:search-app}) with $s$ the natural-language statement.

\emph{PCI-DSS.} Let $\mathit{ctx}$ require payment-specific authorization on an
open network (IP allowlisting infeasible). Sufficient-frontier (downward): the internal root $V_1$ is
excluded (no allowlist available), and $V_2 = M\land J$ entails $s$ but cannot
weaken to $V_3 = M$ (dropping the JWT loses the required payment authorization),
so $F_\downarrow = \{V_2\}$. Necessary-frontier (upward): $s \land \mathit{ctx}$ climbs the chain to
$V_2$ because the context forces both $M$ and $J$, and no higher. Thus
$F_\uparrow = \{V_2\}$: the frontiers coincide, the residue is unsatisfiable, and
the context fully disambiguates, selecting $V_2 = \candTJ$
(Theorem~\ref{thm:disambiguation}).

\emph{Network perimeter.} Let $\mathit{ctx}$ describe a trusted network whose
access is controlled at the perimeter by source-IP allowlisting, with callers
additionally presenting a credential; TLS client certificates and JWTs are not
deployed. This context selects the \emph{other} root. Sufficient-frontier (downward): $V_2 = M\land J$
is excluded (its mechanisms are undeployed), $V_1 = I\land C$ entails $s$, and it
cannot weaken to $V_4 = C$ (dropping the allowlist removes the required perimeter
control), so $F_\downarrow = \{V_1\}$. Necessary-frontier (upward): $s \land \mathit{ctx}$ climbs
$V_6 \to V_4 \to V_1$ (the context forces a standing credential and the
allowlist), giving $F_\uparrow = \{V_1\}$. The frontiers coincide at $V_1$: the
second root is exactly the reading this context pins down---something the
single-root pool of the naive cross product could not express.

\emph{Internal, no compliance.} Let $\mathit{ctx}$ describe a trusted internal
network with no compliance requirement. Sufficient-frontier (downward), both roots weaken: mTLS, the
JWT, and the allowlist are all unnecessary, so the searches descend $V_2 \to V_3
\to V_4$ and $V_1 \to V_4$ to ``any credential'' ($V_4$), but cannot cross $V_4
\to V_6$, since nothing licenses reading the policy as conditional on traffic
arriving. Thus $F_\downarrow = \{V_4\}$. Necessary-frontier (upward), $s \land \mathit{ctx}$ entails
the sink $V_6$ but not the standing obligation $V_4$, so $F_\uparrow = \{V_6\}$.
The residue is $V_6 \land \lnot V_4 \land \mathit{ctx}$
(Remark~\ref{rem:single-edge}): exactly the configurations
that authenticate every arriving request without an unconditional guarantee.
This is the same conditionality gap the curated internal walkthrough isolates as
$\candCr \to \candCo$---here recovered as $V_4 \to V_6$.

\paragraph{Takeaway.} On machine-generated input the framework behaves exactly
as the theory predicts. The solver recovers the entailment order and merges
equivalent surface forms; the PCI-DSS context collapses the bracket to the root
$V_2 = \candTJ$; the perimeter context collapses it to the \emph{other} root $V_1$; and
the internal context reproduces the conditionality gap $V_4 \to V_6 = \candCr \to \candCo$.
The generated pool differs from the curated one only in which specific second
root it contains---$V_1 = I\land C$ rather than the curated $\candTI = M\land I$, since
no cell named that particular pair. By the monotonicity of the residue
(Proposition~\ref{prop:monotone}) such differences are gaps in the pool, not the
framework---adding candidates can only refine the picture---the empirical
counterpart of the relativity discussed in Remark~\ref{rem:relativity}: the
residue is always relative to the pool at hand, and it errs in the safe
direction.

\section{\sys{} Selection Quality at Scale}\label{sec:eval-judge}

To scale the evaluation beyond the expert-adjudicated study of
Section~\ref{sec:eval}, we enlarge the corpus with
systematically generated \emph{mutants} and check each selection against the
oracle's sufficient frontier, and audit \sys{}'s selections with an LLM and by hand. Starting from the $64$ deployment properties we synthesize two mutants
of each---small semantic perturbations of the requirement, some deliberately
weakened---for $192$ statements. Of the $192$, $176$ resulted in valid autoformalized candidates, and $152$ are
\emph{non-zero-entropy} (the pool splits into more than one equivalence class), 
the case where safe selection matters. 

We generate candidates with a single prompt eliciting ten \emph{distinct}
formalizations, rather than ten independent samples as in Section~\ref{sec:eval}.
The statements carry no briefing, so the oracle, USC, and the audit see only the
one-line statement. Asking for distinct candidates manufactures diversity.
Of the $176$ statements with valid candidates, $152$ ($86.4\%$) have non-zero entropy, against $20$ of $64$ ($31.2\%$) in
Section~\ref{sec:eval}. It also handicaps MV, which counts frequency, so the
baseline numbers here are not comparable with those of Section~\ref{sec:eval}. 

Our primary metric is \emph{agreement with the
oracle's sufficient frontier}; the LLM judge serves only as an audit that flags \sys{}
selections for manual inspection.

\paragraph{Compute and configuration.} All experiments were run on a single cloud
virtual machine---$16$ AMD EPYC~9R14 (Zen~4) vCPUs, $30$\,GiB RAM, Amazon
Linux~2023 (Linux~6.12)---with no GPU, as LLM inference is served by a hosted API
(AWS Bedrock). The symbolic layer uses the cvc5~1.3.1 SMT solver~\cite{cvc5}
driven from Python~3.10; candidate generation, the entailment oracle, and the
audit judge all use Claude Opus~4.6\footnote{Bedrock model ID \texttt{global.anthropic.claude-opus-4-6-v1}.} at temperature~$1.0$. We draw $10$ candidate
formalizations per statement and take a $5$-sample majority quorum on every edge
check. The machine-checked proofs (Appendix~\ref{sec:lean}) are developed in
Lean~4~\cite{lean4} with Mathlib~\cite{mathlib}.

\paragraph{Results.} We compare the \sys{} sufficient frontier $F_\downarrow$ against
majority vote over equivalence clusters (MV) and Universal Self-Consistency (USC)
on the $152$ non-zero-entropy statements (Table~\ref{tab:eval-judge}). The metric is
\emph{agreement with the oracle's sufficient frontier}: given $F_\downarrow$, the
solver-computed entailment order decides whether a selection lies at or above
$F_\downarrow$. \sys{} passes by construction, so the metric measures how often a
baseline selects a candidate weaker than the oracle certified. It is not
independent of the oracle; an independent number needs external ground truth.
We do not use the LLM judge to score selections, because it is too lenient: on the
20-case study, where the expert also graded every selection, the judge accepted
majority vote's pick in $75\%$ of cases and the expert in only $44\%$.

Under this metric MV agrees with the frontier on $76.3\%$ of statements and USC on
$77.6\%$; both select a candidate weaker than the oracle certified on roughly a
quarter. \sys{} agrees on all $152$ by construction. We also run the \emph{flat filter} ablation of Section~\ref{sec:eval} at this scale, which asks the
same oracle one global question per candidate instead of one check per edge. Measured against \sys{}'s own frontier, which treats its local decisions as correct, and counting abstentions as safe, the
flat filter lies at or above that frontier on $125/152$ ($82.2\%$), against $76.3\%$ for MV and $77.6\%$ for USC.
What removing the graph costs is structural: the flat filter rejects every candidate in
$12$ statements and contradicts the entailment order in $8$, and \sys{} does neither. Two facts bear on \sys{}'s actual
safety.
First, manual review found $10$ pools that contain no sufficient candidate. \sys{}'s
search emitted a candidate in these pools too, because it seeds its frontier with
the \sys{}'s roots without checking them (Appendix~\ref{sec:realizability-app}), and
that candidate is too weak. No selector could pick a correct candidate there; a
search should instead report ``no sufficient formalization''.
Second, because the metric shares the oracle that builds $F_\downarrow$, an oracle
error passes it. We therefore audited \sys{}'s selections with an LLM judge and
inspected every flagged case by hand. The judge flagged $13$: the $10$ pools
above and $3$ genuine too-weak selections, each an oracle error (we analyze them below).
The audit covered \sys{} only, used the same model as the oracle, and did not inspect
unflagged selections; each of these choices favors \sys{}.
Counting the $10$ pools as abstentions, \sys{} is safe on at least $149/152$
($98.0\%$); scoring the formula it emitted, on $139/152$ ($91.4\%$). On the $142$
pools that contain a sufficient candidate, \sys{} is safe on $139$ ($97.9\%$), MV agrees
with the frontier on $108$ ($76.1\%$), and USC on $109$ ($76.8\%$).

\paragraph{Leaving the majority.} \sys{}'s selection lies outside the two largest
clusters in $20$ of the $152$ statements ($13.2\%$); no rule that ranks clusters by
size can make these selections.

\paragraph{Abstention and trivial selectors.} Abstention is available to \sys{} and not
to the baselines. If MV abstains on its $33$ ties and abstention counts as safe, it
scores $124/152$ ($81.6\%$); on the $119$ statements with a unique largest cluster
it agrees with the frontier on $91$ ($76.5\%$). One-sided safety is also easy to
game: conjoining every candidate agrees with the frontier on $152/152$, and picking
the strongest candidate on $126$ ($82.9\%$, if every root must pass) to $138$
($90.8\%$, if one passing root suffices).

\begin{table}[ht]
\centering
\small
\setlength{\tabcolsep}{4pt}
\begin{tabular}{@{}p{0.60\columnwidth}cc@{}}
\hline
\textbf{Quantity} & \textbf{Count} & \textbf{Fraction} \\
\hline
Statements ($64$ base $+$ $128$ mutants) & 192 & --- \\
\quad Successfully autoformalized & 176 & 91.7\% \\
\quad Zero entropy (single class; agree) & 24 & 13.6\% \\
\quad \textbf{Non-zero entropy (the benchmark)} & \textbf{152} & \textbf{86.4\%} \\
\hline
\multicolumn{3}{l}{\emph{The 152 informative cases, by agreement}} \\
\quad Agree on same candidate & 85 & 55.9\% \\
\quad Diverge & 67 & 44.1\% \\
\hline
\multicolumn{3}{l}{\emph{Agrees with the oracle's sufficient frontier}} \\
\quad Majority vote over clusters (MV) & 116 & 76.3\% \\
\quad Universal Self-Consistency (USC) & 118 & 77.6\% \\
\quad Conjoin all candidates & 152 & 100\% \\
\quad Strongest candidate (every / one root) & 126--138 & 82.9--90.8\% \\
\quad \sys{} sufficient frontier (by construction) & 152 & 100\% \\
\hline
\multicolumn{3}{l}{\emph{Ablation: \sys{} with the graph removed (same oracle, same model)}} \\
\quad Flat filter (one global query per candidate) \\
\quad\quad \emph{--- agrees with the frontier} & 113 & 74.3\% \\
\quad\quad \emph{--- below the frontier} & 27 & 17.8\% \\
\quad\quad \emph{--- rejects every candidate} & 12 & 7.9\% \\
\quad\quad \emph{--- contradicts the entailment order} & 8 & 5.3\% \\
\hline
\multicolumn{3}{l}{\emph{\sys{} after audit}} \\
\quad Safe, $10$ empty pools as abstentions & 149 & 98.0\% \\
\hline
\multicolumn{3}{l}{\emph{On the $142$ pools with a sufficient candidate}} \\
\quad \sys{} (safe after audit) & 139 & 97.9\% \\
\quad MV / USC (agree with frontier) & 108 / 109 & 76.1 / 76.8\% \\
\quad MV, abstaining on ties (of $152$) & 124 & 81.6\% \\
\hline
\end{tabular}
\caption{Selection quality on the $152$ non-zero-entropy statements of the scale
study ($64$ properties $+$ $128$ mutants). The first block checks whether a selection
lies at or above the oracle's sufficient frontier in the solver-computed \sys{} order;
\sys{} passes by construction, and the check cannot see oracle errors. The second block
scores \sys{} after an LLM audit with manual inspection, which found $3$ oracle errors;
$10$ pools contain no sufficient candidate, and \sys{} emitted a too-weak candidate in
each. The trivial selectors show that this one-sided check does not penalize
over-strong picks. Percentages: the autoformalization rate is of the $192$
synthesized statements; the entropy split is of the $176$ autoformalized; all other
rows are of the $152$ non-zero-entropy statements unless stated.}
\label{tab:eval-judge}
\end{table}

\paragraph{USC.} USC~\cite{usc} shows one LLM call the statement and all
candidates and returns the most self-consistent one; its prompt is neutral as to
logical strength and, in this study, carries no briefing. It agrees with the
frontier on $77.6\%$, close to MV's $76.3\%$. USC tests whether self-consistency
tracks the intended reading, which is a different question from sufficiency.

\paragraph{Analysis of the three \sys{} misses.} We performed a manual audit of three
too-weak \sys{} selections. The failure mode was the same in all cases; the entailment oracle relaxes past
an external LLM audit's safety threshold. Two are the \emph{same} EFS property---the base
and one of its mutants---requiring that an \texttt{EFSVolumeConfiguration}
set \texttt{TransitEncryption} to
\texttt{"ENABLED"}. The selected frontier additionally admits the lowercase literal
\texttt{"enabled"}; the oracle reasoned that casing is unspecified, but the
CloudFormation schema fixes the enum to $\{\texttt{ENABLED},\texttt{DISABLED}\}$, so
\texttt{"enabled"} does not turn transit encryption on and the selected candidates are too
weak. A template satisfying the selected candidate would still deploy, but unsafely allows for an
unencrypted resource. 

The third case is a mutated AWS Lambda Function property specifying that inline
\texttt{ZipFile} code requires \texttt{Handler}$=$\texttt{"index.handler"} ``(Node.js
and Python only).'' The selected candidate guards the requirement with an added
runtime condition, enforcing the \texttt{index.handler} constraint only when the
runtime is empty or begins with \texttt{nodejs}/\texttt{python}. Consequently a
template that sets \texttt{ZipFile} together with any other runtime---e.g.\
\texttt{java17}---\emph{satisfies} the selected candidate, because its guard is never
triggered and the check simply passes, even though such a template is not
deployable: inline \texttt{ZipFile} is valid only for Node.js and Python. 

The entailment oracle admitted this weakening by reading the parenthetical ``(Node.js
and Python only)'' as a \emph{scope restriction} rather than an incidental remark: it
reasoned that a non-Node.js/Python runtime lies outside the property's scope,
so leaving those cases unconstrained still ``satisfies'' the requirement. Under that
reading the extra runtime guard looks like a faithful tightening of scope rather than
a loss of coverage. The reading is textually defensible. We expect that giving the oracle the resource
schema's enum and realizability constraints would prevent all three misses, but we
have not tested this.

\paragraph{On the divergent cases.} There are $67$
\emph{divergent} cases where \sys{} and MV select different candidates; these isolate selection quality, and there the gap is stark. MV is incorrect---too weak or too strong---in $47.8\%$ of
them.

\paragraph{No dominant reading.} The \sys{}'s advantage concentrates where MV has no signal. In $33$ of the $152$ cases ($21.7\%$) there is
\emph{no unique largest cluster}---two or more equivalence classes tie for the most
frequent. So MV must break the tie arbitrarily. In $14$ of these
($9.2\%$) \emph{all} classes are equal in cardinality, making its choice
effectively random. The \sys{} selection does not use frequency and is unaffected, so
MV degrades most in these cases
and the \sys{}'s margin is largest.

\section{Entailment Oracle Quality}\label{sec:oracle-quality}

\subsection{Self-Consistency}\label{sec:oracle-consistency}

\begin{table}[ht]
\centering
\small
\setlength{\tabcolsep}{4pt}
\begin{tabular}{@{}p{0.56\columnwidth}cc@{}}
\hline
\textbf{Metric} & \textbf{Suff.} & \textbf{Nec.} \\
\hline
Edge checks ($\times 5$ samples) & 54 & 57 \\
Unanimous ($5/5$ identical) & 85\% & 86\% \\
Mean agreement with quorum & 0.96 & 0.97 \\
Verdict flips, $k{=}3$ vs $k{=}5$ & 4\% & 0\% \\
Verdict flips, $k{=}1$ vs $k{=}5$ & 4\% & 7\% \\
\hline
\end{tabular}
\caption{Oracle quorum stability in the 20-case study of
Section~\ref{sec:eval} ($111$ edge checks, $5$ votes each, $555$ samples). This measures the oracle's \emph{self-consistency}, not its \emph{soundness} against labeled ground truth. $86\%$ of checks are unanimous. Shrinking the quorum from $5$ to $3$ changes at most $4\%$ of verdicts, so the quorum size is not a sensitive parameter. }
\label{tab:oracle-consistency}
\end{table}

Every edge traversal decision reduces to a local edge check that the LLM entailment oracle answers. Since safety is conditional on the oracle being sound (Theorem~\ref{thm:realizability}), in the 20-case study we decide each check by a
$5$-vote quorum and present how stable the LLM entailment oracle is.

Over $555$ individual judgments the oracle is highly self-consistent and the $5$-quorum size is not a sensitive hyperparameter (Table~\ref{tab:oracle-consistency}). Roughly $86\%$ of edge checks are unanimous, mean agreement with the quorum verdict is $0.96$--$0.97$, and reducing the quorum to $k{=}3$ changes at most $4\%$ of verdicts.

\subsection{Soundness}\label{sec:oracle-soundness}

 To accurately measure oracle soundness, each check must be decided with a formal statement and a formal context. We take the $20$ cases of Section~\ref{sec:eval} and use each case's annotated candidate as the formal statement $s$. The context is natural language in our corpus, so we formalize with a best-effort approach and only in part: for four statements we add schema facts as formulas (an enum, a numeric range, IAM's rule that every policy statement names an action, and the form and uniqueness of IAM role ARNs). cvc5 then decides every check the searches issued, and we classify each accepted check three ways. A check \textbf{holds} if it holds under this partial context, in which case it also holds under the full one, since added context can only make a check hold. A check is \textbf{wrong} if it fails and cvc5's counterexample is a valid configuration that violates $s$. A check is \textbf{undecided} if it fails and the counterexample is excluded by a fact that remains informal: the oracle read that fact in the natural-language context, the formalized fragment lacks it, and so the fragment settles the check in neither direction. An undecided check is evidence of neither soundness nor unsoundness, so the rate we report is over decided checks. The results are in Table~\ref{tab:oracle-soundness}.

\begin{table}[ht]
\centering
\small
\setlength{\tabcolsep}{4pt}
\begin{tabular}{@{}p{0.46\columnwidth}ccc@{}}
\hline
\textbf{Metric} & \textbf{Total} & \textbf{Suff.} & \textbf{Nec.} \\
\hline
Checks the oracle accepts & 30 & 22 & 8 \\
\quad and the check holds & 22 & 15 & 7 \\
\quad and the check is wrong & 1 & 1 & 0 \\
\quad and the check is undecided & 7 & 6 & 1 \\
Rate holding, decided checks & $96\%$ & $94\%$ & $100\%$ \\
\hline
\end{tabular}
\caption{Oracle soundness on the 20 cases of Section~\ref{sec:eval}, with each case's annotated candidate as the formal statement and a partly formalized context. An \emph{undecided} check fails under the formalized fragment, but its counterexample is excluded by a fact that remains informal, so the fragment settles it in neither direction. The rate is over decided checks, those that hold plus the one that is wrong; undecided checks are excluded from it.}
\label{tab:oracle-soundness}
\end{table}

The one wrongly accepted check has a valid counterexample with two functions that use KMS keys: one role grants every action, and the other role has no policies. The oracle judged the edge one function at a time and missed that the second function violates $s$. The seven undecided checks need facts the oracle cites but the formal context lacks: which EC2 instance families are Nitro-based, that a launch template names either an instance type or instance requirements but not both, that a Docker image's registry is the part before the first ``/'', and whether CloudFormation accepts a task definition named by family alone. Formalizing these facts would decide each check. Two of the seven accepted necessity checks that hold leave a necessary-frontier candidate above a sufficient-frontier one in the \sys{}; under the added schema facts the two candidates are equivalent, which is the case of Corollary~\ref{cor:inversion}, so the residue is correctly unsatisfiable. The one wrongly accepted check did not produce \sys{}'s pick, so Table~\ref{tab:eval} is unaffected.

\section{Use of Generative AI}\label{app:ai-use}

LLMs are part of the evaluation: they generate the candidate formalizations, serve as the entailment oracle, and act as the audit judge (Section~\ref{sec:impl}). Separately, we used AI assistants to edit the prose and to write evaluation scripts. The authors checked every result and take responsibility for the content.

\fi 

\end{document}

%% file: fig-ig-bracket.tex
\begin{figure}[t]
\setlength{\abovecaptionskip}{4pt}   
\centering
\resizebox{\fitcol}{!}{%
\begin{tikzpicture}[
  node distance=0.9cm and 1.8cm,
  vertex/.style={circle, draw, thick, minimum size=1.8em, inner sep=1pt, font=\small},
  dimvertex/.style={vertex, draw=gray!50, text=gray!50},
  uppervertex/.style={vertex, fill=blue!15, draw=blue!70, thick},
  lowervertex/.style={vertex, fill=red!15, draw=red!70, thick},
  ambigvertex/.style={vertex, fill=yellow!25, draw=orange!70},
  edge/.style={-{Stealth[length=5pt]}, thick},
  dimedge/.style={edge, draw=gray!40},
  ambigregion/.style={rounded corners=8pt, draw=orange!70, dashed, thick, fill=yellow!8}
]
  \node[dimvertex] (r1) {$r_1$};
  \node[dimvertex, right=of r1] (r2) {$r_2$};

  \node[lowervertex, below left=0.5cm and 0.2cm of r1] (u1) {$u_1$};
  \node[lowervertex, below right=0.5cm and 0.2cm of r2] (u2) {$u_2$};

  \node[ambigvertex, below right=0.5cm and -0.3cm of u1] (m1) {$m_1$};
  \node[ambigvertex, below left=0.5cm and -0.3cm of u2] (m2) {$m_2$};

  \node[uppervertex, below right=0.5cm and -0.3cm of m1] (v1) {$v_1$};
  \node[uppervertex, below left=0.5cm and -0.3cm of m2] (v2) {$v_2$};

  \node[dimvertex, below right=0.5cm and 0.2cm of v1] (s1) {$s_1$};
  \node[dimvertex, below left=0.5cm and 0.2cm of v2] (s2) {$s_2$};

  \begin{scope}[on background layer]
    \node[ambigregion, fit=(m1)(m2), inner sep=4pt] (ambigrect) {};
  \end{scope}

  \draw[dimedge] (r1) -- (u1);
  \draw[dimedge] (r1) -- (u2);
  \draw[dimedge] (r2) -- (u2);
  \draw[dimedge] (r2) -- (u1);

  \draw[edge] (u1) -- (m1);
  \draw[edge] (u1) -- (m2);
  \draw[edge] (u2) -- (m2);
  \draw[edge] (u2) -- (m1);

  \draw[edge] (m1) -- (v1);
  \draw[edge] (m1) -- (v2);
  \draw[edge] (m2) -- (v1);
  \draw[edge] (m2) -- (v2);

  \draw[dimedge] (v1) -- (s1);
  \draw[dimedge] (v1) -- (s2);
  \draw[dimedge] (v2) -- (s1);
  \draw[dimedge] (v2) -- (s2);

  \draw[red!70, very thick, dashed] ([xshift=-1.2cm]u1.west) -- (u1.west);
  \draw[red!70, very thick, dashed] (u1.east) -- (u2.west);
  \draw[red!70, very thick, dashed] (u2.east) -- ([xshift=1.2cm]u2.east)
    node[right, font=\small, text=red!70] {lower frontier $F_\downarrow$};
  \draw[blue!70, very thick, dashed] ([xshift=-1.2cm]v1.west) -- (v1.west);
  \draw[blue!70, very thick, dashed] (v1.east) -- (v2.west);
  \draw[blue!70, very thick, dashed] (v2.east) -- ([xshift=1.2cm]v2.east)
    node[right, font=\small, text=blue!70] {upper frontier $F_\uparrow$};

  \node[font=\scriptsize, text=gray!70, right=2.2cm of r2] {stronger};
  \node[font=\scriptsize, text=gray!70, right=2.2cm of s2] {weaker};
  \draw[gray!50, -{Stealth[length=4pt]}] ([xshift=2.5cm]r2.south) -- ([xshift=2.5cm]s2.north);

  \node[font=\small, text=orange!80!black, right=0.4cm of ambigrect] {ambiguity gap};

\end{tikzpicture}}
\caption{Bracketing a statement with an \sysname{}. The lower
  frontier $F_\downarrow$ (red) marks the weakest candidates that
  entail the statement.
%
  The upper frontier $F_\uparrow$ (blue) marks the strongest
  candidates the statement entails.
  The shaded region remains ambiguous, and the ambiguity residue, a
  logical formula, describes exactly the situations it covers.}
\label{fig:ig-bracket}
\end{figure}
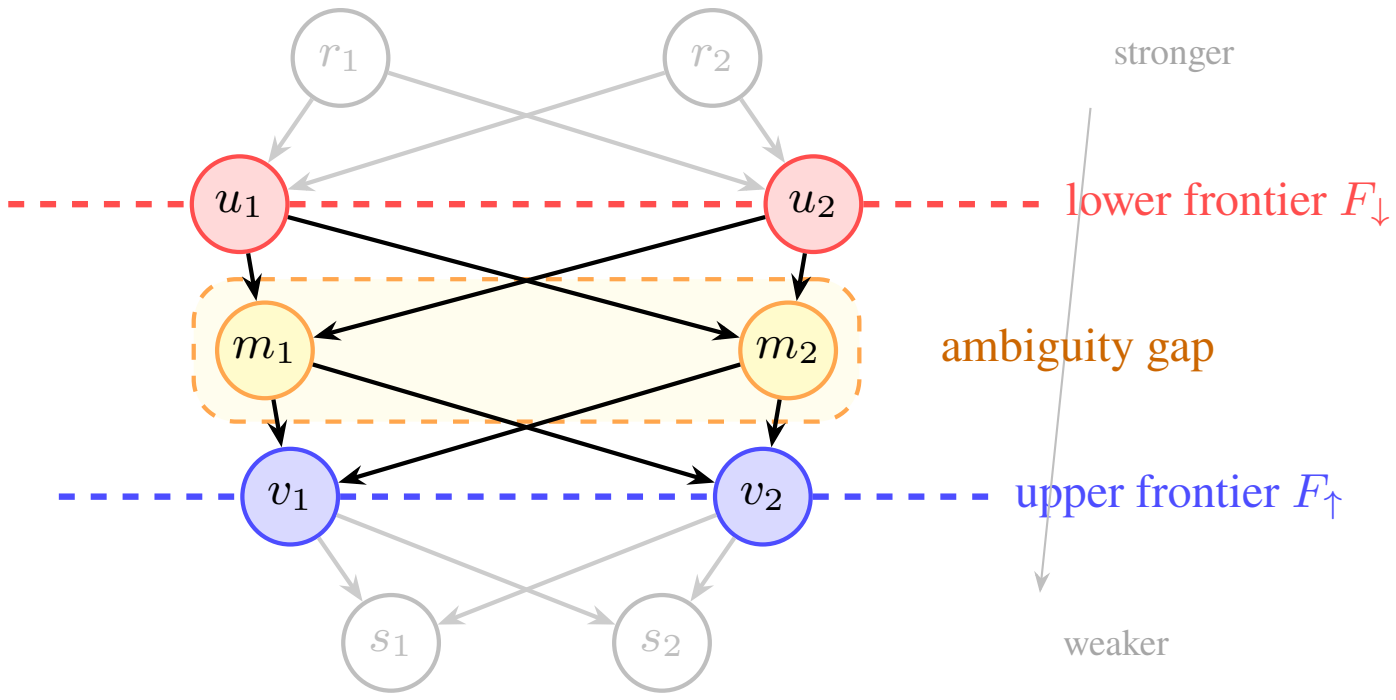

%% file: fig-ig-payment-short.tex
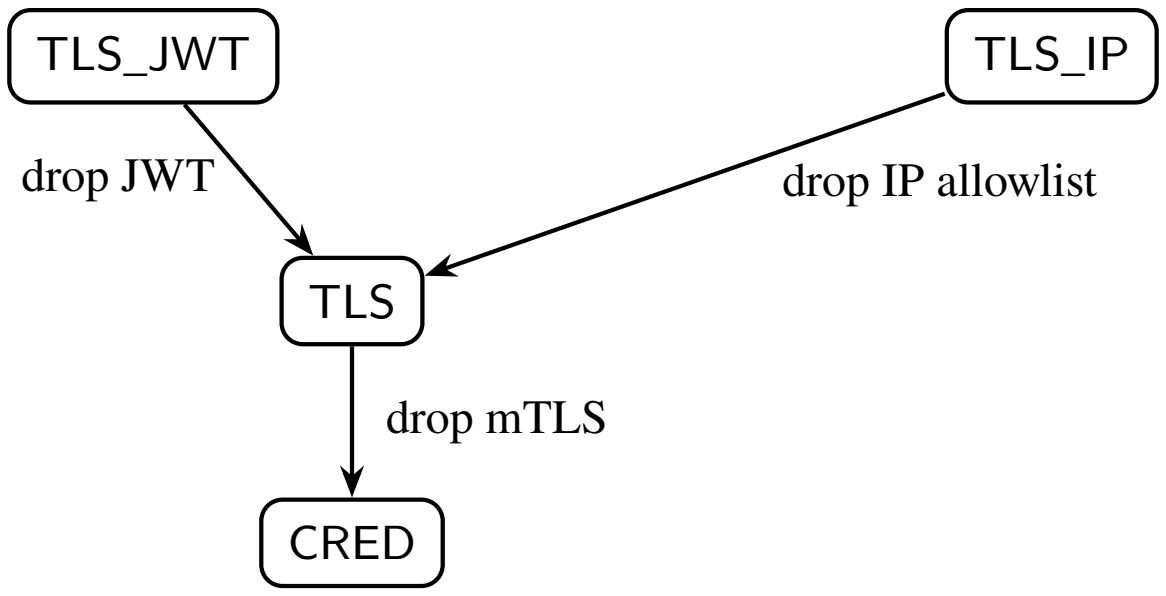
\begin{figure}[t]
\centering
\resizebox{\ifdim\width>0.8\columnwidth 0.8\columnwidth\else\width\fi}{!}{%
\begin{tikzpicture}[
  node distance=1.0cm and 1.2cm,
  vertex/.style={rectangle, rounded corners, draw, thick, font=\small, inner sep=5pt},
  edge/.style={-{Stealth[length=6pt]}, thick},
  edgelabel/.style={font=\footnotesize, align=center, inner sep=2pt}
]
  \node[vertex] (A) {$\candTJ$};
  \node[vertex, right=4.4cm of A] (B) {$\candTI$};
  \node[vertex, below right=1.0cm and 0cm of A] (C) {$\candT$};
  \node[vertex, below=of C] (D) {$\candCr$};

  \draw[edge] (A) -- node[edgelabel, left=4pt] {drop JWT} (C);
  \draw[edge] (B) -- node[edgelabel, right=16pt] {drop IP allowlist} (C);
  \draw[edge] (C) -- node[edgelabel, right=4pt] {drop mTLS} (D);
\end{tikzpicture}}
\caption{\sysname{} for the payment service authentication
  example. Edges point from stronger (more restrictive) to weaker
  (more permissive) candidates and edge labels show the corresponding
  weakening. Roots $\candTJ$ and $\candTI$ are incomparable; sink
  $\candCr$ is the weakest. For readability, we draw only covering
  edges; the \sys{} also contains the transitive edges (e.g., $\candTJ \to
  \candCr$).}
\label{fig:ig-example}
\end{figure}

%% file: new_eval.tex
We compare \sys{} selection against two baselines. The first is \textbf{majority vote
over equivalence clusters (MV)}: it clusters candidates by solver-checked
equivalence and selects from the largest cluster. The clustering behind semantic
entropy~\cite{semantic-entropy} suggests this selection rule;
semantic entropy itself measures uncertainty and does not select. MV sees no
context. The second is Universal Self-Consistency (\textbf{USC})~\cite{usc}: one
LLM call sees the statement, its briefing, and all candidates, and returns the most
self-consistent one. USC's prompt is neutral as to logical strength, so it measures
self-consistency rather than sufficiency.
Together these two baselines instantiate the two ingredients of the
closest prior autoformalization selector, Li et
al.~\cite{symbolic-consistency}, which scores candidates by
prover-checked symbolic equivalence (the MV component) combined with
semantic consistency (the USC component).
Because our goal in this paper is safety, we first evaluate on $20$ cases that a
human expert adjudicated, then on $152$ statements, where we check each selection against the oracle's sufficient
frontier and audit \sys{}'s selections with an LLM and by hand.

\begin{table}[ht]
\centering
\small
\setlength{\tabcolsep}{4pt}
\begin{tabular}{@{}p{0.60\columnwidth}cc@{}}
\hline
\textbf{Quantity} & \textbf{Count} & \textbf{Fraction} \\
\hline
Properties & 64 & --- \\
\quad Zero entropy (single eq class) & 44 & 68.8\% \\
\quad \textbf{Non-zero entropy (the benchmark)} & \textbf{20} & \textbf{31.2\%} \\
\hline
\multicolumn{3}{l}{\emph{Accuracy on the 20 (expert manual audit)}} \\
\quad Majority vote over clusters (MV) \\
\quad\quad \emph{--- safe, matches reference}  & 8 & 40\% \\
\quad\quad \emph{--- safe but conservative} & 7 & 35\% \\
\quad\quad \emph{--- unsafe (admits a violation)} & 5 & 25\% \\

\quad Universal Self-Consistency (USC) \\
\quad\quad \emph{--- safe, matches reference} & 10 & 50\% \\
\quad\quad \emph{--- safe but conservative} & 8 & 40\% \\
\quad\quad \emph{--- unsafe (admits a violation)} & 2 & 10\% \\

\quad \textbf{\sys{} selection (ours)} \\
\quad\quad \emph{--- safe, matches reference} & \textbf{18} & \textbf{90\%} \\
\quad\quad \emph{--- safe but conservative} & \textbf{2} & \textbf{10\%} \\
\quad\quad \emph{--- unsafe} & \textbf{0} & \textbf{0\%} \\

\hline
\multicolumn{3}{l}{\emph{Ablation: \sys{} with the graph removed (same oracle, same model)}} \\
\quad Flat filter (one global query per candidate) \\
\quad\quad \emph{--- safe, matches reference} & 6 & 30\% \\
\quad\quad \emph{--- safe but conservative} & 3 & 15\% \\
\quad\quad \emph{--- unsafe (admits a violation)} & 5 & 25\% \\
\quad\quad \emph{--- rejects every candidate} & 6 & 30\% \\

\hline
\multicolumn{3}{l}{\emph{Ambiguity residue}} \\
\quad Decided: \texttt{unsat} (frontiers overlap) & 13 & 65\% \\
\quad Undecided: \texttt{unknown} & 7 & 35\% \\
\hline
\end{tabular}
\caption{\sys{} vs two baselines---majority vote over equivalence clusters (MV) and
  Universal Self-Consistency (USC)---on 64 properties, against an expert
  reference. \emph{Unsafe} candidates are too weak and admit a violation;
  \emph{conservative} candidates are safe but may reject permitted configurations.
  On the $20$ non-zero-entropy cases MV is unsafe on $5$ and USC on $2$; \sys{} is safe
  on all $20$ ($18$ match the reference, $2$ are conservative). A flat filter that asks the same oracle once per
  candidate, without the graph, is unsafe on $5$ and rejects every candidate on $6$.}
\label{tab:eval}
\end{table}

For the human-adjudicated experiment, we considered 64 established NL
properties for 20 AWS resource types.\footnote{The resources
are: EC2 [Instance, Subnet, VPC Endpoint, VPC Peering Connection], ECS
[Service, Task Definition], IAM [Instance Profile, OIDC Provider,
  Policy, Role, Service Linked Role], Lambda [Event Source Mapping,
  Function, Layer Version], RDS [Cluster Instance, DB Cluster, DB
  Instance, DB Subnet Group], S3 Bucket, and SQS Queue.}
\paragraph{Ground truth.} For each property we deployed conforming and violating
CloudFormation templates to AWS to confirm what the property admits, and judged
each selection against that behavior.
Table~\ref{tab:eval} shows the results for autoformalizing these 64
properties with \sysnames{} (\sys{}) and compares it to the two baselines.
We note that because of the security-critical nature of the domain, the
desired outcome for autoformalization is to guarantee 100\% safety.
More generally, for any verification task, less than 100\%
safety is undesirable.

In $68.8\%$ of the cases all candidates are equivalent (zero entropy) and the three
selectors agree.
On the remaining $20$, \sys{} selects the expert reference in $18$ cases and a safe but
more conservative candidate in $2$, so it is safe in all $20$. MV is safe in $15$
and unsafe in $5$; USC is safe in $18$ and unsafe in $2$.
\sys{}'s selection is sufficient by construction, conditional on the oracle; neither
baseline carries such a guarantee, and both admit a violation on some case.
\sys{} matches the reference in $18/20$ cases, MV in $8/20$, and USC in $10/20$.

To get more insight into \syss{}, we examined the ambiguity residues.
Of the $20$ cases of non-zero entropy, the solver finds the ambiguity residue
unsatisfiable in $13$; in each, the necessary frontier entails the sufficient one.
For the remaining $7$, the solver cannot decide the residue, and \syss{} still select the
safe formula.
One approach to improving \sys{}'s accuracy in cases of ambiguity is to
reduce or eliminate ambiguity by increasing the candidate pool (generate
more candidates), possibly guided by the ambiguity residue itself.
Note that such improvements would increase accuracy; \syss{} remain safe
and sound even in the case of ambiguity.

\paragraph{Ablation: isolating the graph from the oracle.} To test whether \sys{}'s advantage comes from
the \sysnoun{} or only from asking an LLM about entailment, we ran an ablation
experiment, which we call the \emph{flat filter}: it removes the graph and keeps
everything else. We ask the same oracle and model, once per candidate, whether that
candidate alone entails the statement in context, and return the weakest candidate
that passes. With a perfect oracle the flat filter returns \sys{}'s sufficient frontier
(Lemma~\ref{lem:completeness}), so the difference measures how the oracle errs on
one global question per candidate instead of one local question per edge.
The experiments show that the flat filter is unsafe in $5/20$ cases, whereas \sys{} is
unsafe in none, and that in a further $6/20$ it rejects every candidate even
though the pool contains the reference. In full, the flat filter matches the
reference in $6/20$, is conservative but safe in $3/20$, unsafe in $5/20$, and
returns nothing in $6/20$, against \sys{}'s $18/20$, $2/20$, $0/20$, and no
rejections. The flat filter's answers also contradict the entailment order in
$3/20$ statements: a weaker candidate passes while a stronger candidate that
implies it fails. The ablation shows that the graph, not access to an entailment oracle, is the
key to the guarantee: the same oracle and model, asked one global question per
candidate, are both unsafe and frequently unable to answer.

\paragraph{Scaling the study.}
To test \syss{} at a larger scale, we created an NL property set of $192$
statements, of which $152$ are non-zero-entropy, by including mutants
of the original $64$ (Appendix~\ref{sec:eval-judge}).
This study differs from the first in two ways that make its baseline numbers
incomparable with those above. Candidates come from one prompt asking for ten
distinct formalizations, which raises the non-zero-entropy rate ($152$ of the $176$ statements with valid candidates, $86.4\%$, against
$20$ of $64$, $31.2\%$) and handicaps any selector that counts frequency. The statements also
carry no briefing, so the oracle and USC see only the one-line statement.
We judge the safety of each selection in two ways. The first is a solver check, where it
asks whether the selection lies at or above the oracle's sufficient frontier
$F_\downarrow$ in the \sys{}. \sys{} passes this check by construction,
so the check measures only how often a baseline selects a candidate weaker than the oracle
certified. MV does so on $23.7\%$ of statements and USC on $22.4\%$. Measured against \sys{}'s own frontier, which treats its local decisions as correct, and counting abstentions as safe, the
flat filter lies at or above that frontier on $125/152$ ($82.2\%$), against $76.3\%$ for MV and $77.6\%$ for USC.
What removing the graph costs is structural: the flat filter rejects every candidate in
$12$ statements and contradicts the entailment order in $8$, and \sys{} does neither (Table~\ref{tab:eval-judge}).
Because this check shares the oracle, it cannot catch oracle errors. The second is
a manual audit; an LLM reviews \sys{}'s selections, and we inspect every case it flags by
hand. Of the $13$ flagged cases, $10$ pools contain no sufficient candidate, and $3$ are
oracle misses. The audit used the same model as the oracle,
covered only \sys{}'s selections, and left unflagged selections uninspected, so $3$ is
a lower bound on \sys{}'s misses.
If we count the $10$ pools as abstentions, \sys{} is safe on at least $149/152$ ($98.0\%$).

\paragraph{Oracle self-consistency.}
Because safety is conditional on the entailment oracle being sound
(Theorem~\ref{thm:realizability}), we instrument every edge check with a $5$-sample
quorum in the $20$-case study and confirm that the oracle is
highly self-consistent---$86\%$ of edge checks are unanimous and shrinking the quorum
from $5$ to $3$ flips at most $4\%$ of verdicts, so the quorum size is not a
sensitive parameter (full data in Appendix~\ref{sec:oracle-consistency}).
Note that this quantifies the oracle's \emph{self-consistency}, not its
\emph{soundness}---whether the checks it accepts in fact hold---and since one model
fills every role it is weak evidence even of that. We measure soundness next.

\paragraph{Oracle soundness.}
Theorem~\ref{thm:realizability} conditions safety on the oracle being sound, so the
number that matters is how often the oracle \emph{accepts} a check that fails, that
is, answers yes to an entailment that is false. A
wrongly \emph{rejected} check costs only tightness, so the two error directions want
separate numbers and only the acceptance rate bounds how often safety fails.
We measure that direction on the $20$ cases of this section: we take each case's
expert-annotated candidate as the formal statement $s$, formalize the context in
part, and have cvc5 decide every check the searches accepted.
Of the $30$ accepted checks, $22$ hold, $1$ is wrong, and $7$ are undecided because
the solver lacks facts that remain informal. The oracle is therefore
sound on $22$ of the $23$ checks the solver decides.
The one wrong check did not produce \sys{}'s selection, so the selections in
Table~\ref{tab:eval} are unaffected.
The measurement assumes the annotated candidate faithfully formalizes the statement,
an assumption no formal argument can check
(Section~\ref{sec:realizability}). The counts are also small, and the necessity side
rests on $8$ checks.
Appendix~\ref{sec:oracle-soundness} gives the protocol and the per-case analysis.

%% file: fig-ig-payment-app.tex
\begin{figure}[ht]
\centering
\resizebox{\fitcol}{!}{%
\begin{tikzpicture}[
  node distance=1.6cm and 1.2cm,
  vertex/.style={rectangle, rounded corners, draw, thick, font=\small, inner sep=5pt},
  edge/.style={-{Stealth[length=6pt]}, thick},
  edgelabel/.style={font=\scriptsize, align=center, text width=2.8cm, fill=white, inner sep=1pt}
]
  \node[vertex] (A) {$\candTJ$};
  \node[vertex, right=4.4cm of A] (B) {$\candTI$};
  \node[vertex, below right=1.6cm and 0cm of A] (C) {$\candT$};
  \node[vertex, below=of C] (D) {$\candCr$};
  \node[vertex, below=of D] (E) {$\candCo$};

  \draw[edge] (A) -- node[edgelabel, left] {drop JWT} (C);
  \draw[edge] (B) -- node[edgelabel, right] {drop IP allowlist} (C);
  \draw[edge] (C) -- node[edgelabel, right] {drop mTLS} (D);
  \draw[edge] (D) -- node[edgelabel, right] {make conditional} (E);
\end{tikzpicture}}
\caption{\sysname{} for the payment service authentication example. Edges point from stronger (more restrictive) to weaker (more permissive) candidates. Edge labels show what each weakening step drops. Roots $\candTJ$ and $\candTI$ are incomparable; sink $\candCo$ is the weakest. For readability only covering edges are drawn; the \sys{} also contains the transitive edges (e.g., $\candTJ \to \candCr$, $\candT \to \candCo$).}
\label{fig:ig-example-app}
\end{figure}

%% file: fig-ig-generated.tex
\begin{figure}[ht]
\centering
\resizebox{\fitcol}{!}{%
\begin{tikzpicture}[
  vertex/.style={circle, draw, thick, minimum size=2em, inner sep=1pt},
  edge/.style={-{Stealth[length=6pt]}, thick},
  nlab/.style={font=\scriptsize, align=left, text width=3.0cm},
  nlabl/.style={font=\scriptsize, align=right, text width=3.0cm}
]
  \node[vertex] (V1) at (-2.6,0)   {$V_1$};
  \node[vertex] (V2) at (2.2,0)    {$V_2$};
  \node[vertex] (V3) at (2.2,-1.7) {$V_3$};
  \node[vertex] (V4) at (-0.6,-3.4){$V_4$};
  \node[vertex] (V5) at (3.4,-3.4) {$V_5$};
  \node[vertex] (V6) at (1.0,-5.1) {$V_6$};

  \draw[edge] (V1) -- (V4);
  \draw[edge] (V2) -- (V3);
  \draw[edge] (V3) -- (V4);
  \draw[edge] (V3) -- (V5);
  \draw[edge] (V4) -- (V6);
  \draw[edge] (V5) -- (V6);

  \node[nlabl, left=0.35cm of V1] {$I\land C$ \\ (allowlist + credential)};
  \node[nlab, right=0.35cm of V2] {$M\land J$ \quad(= curated $\candTJ$)};
  \node[nlab, right=0.35cm of V3] {$M$ \quad(mTLS only, $=\candT$)};
  \node[nlabl, left=0.35cm of V4] {$C$ \\ (any credential, $=\candCr$)};
  \node[nlab, right=0.35cm of V5] {arrive $\Rightarrow M$};
  \node[nlab, right=0.35cm of V6] {arrive $\Rightarrow C$ \\ (conditional, $=\candCo$)};
\end{tikzpicture}}
\caption{\sysname{} for the machine-generated pool obtained from
per-architecture strength ladders (each architecture laddered within its own
mechanism set). Edges point from stronger to weaker (covering edges only). The
two roots $V_1 = I\land C$ (internal) and $V_2 = M\land J$ (PCI-DSS) live in
disjoint mechanism spaces and are incomparable. Using the domain's reading that
an mTLS certificate and a payment-scope JWT are each instances of a valid
credential ($M \models C$, $J \models C$), the right branch
$V_2 \to V_3 \to V_4 \to V_6$ reproduces the curated chain
$\candTJ \to \candT \to \candCr \to \candCo$ of Figure~\ref{fig:ig-example}, while $V_1$ supplies a
genuine second root---discovered rather than designed.}
\label{fig:ig-generated}
\end{figure}
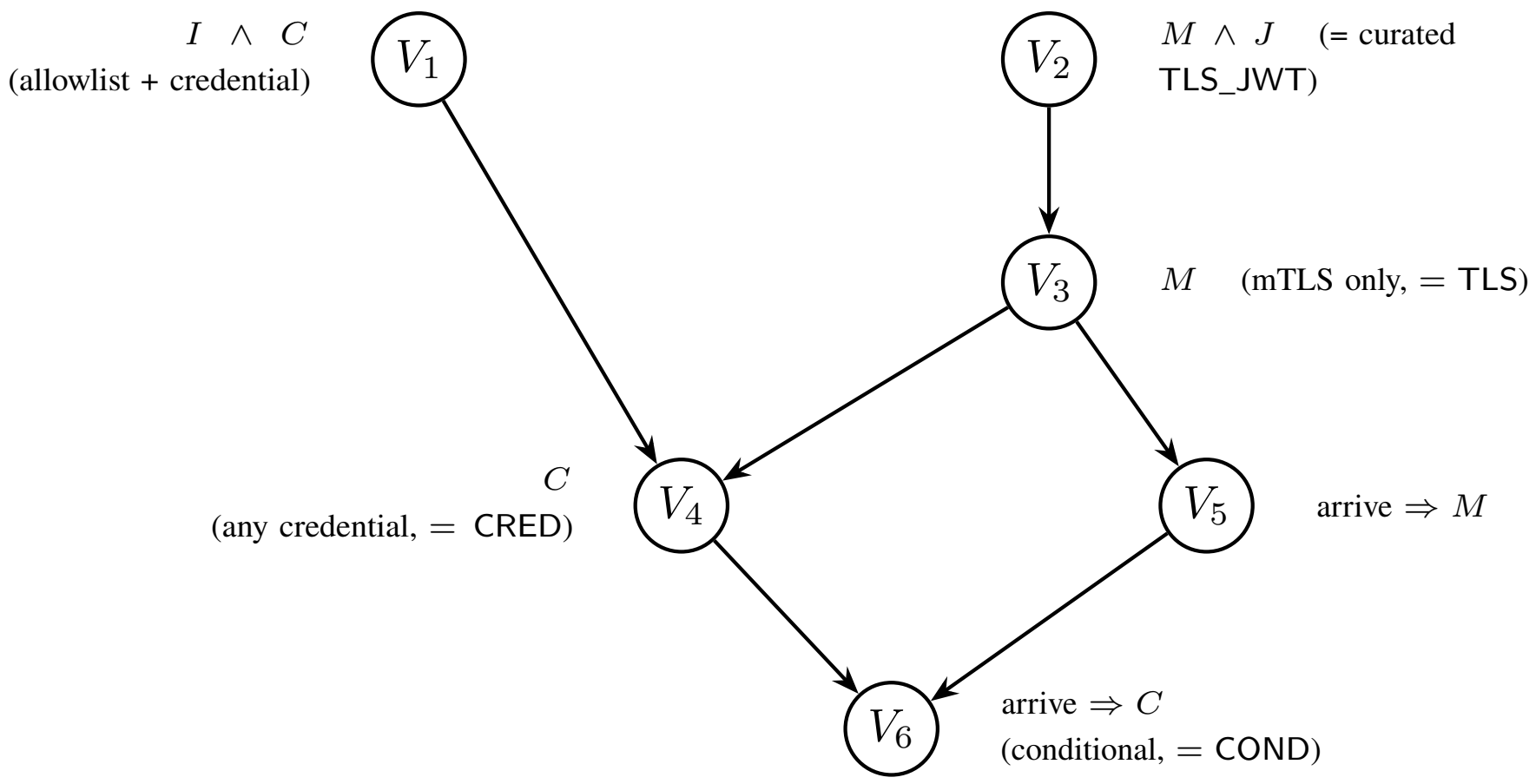